%% file: acl_latex.tex
\pdfoutput=1

\documentclass[11pt]{article}

\usepackage[table,xcdraw]{xcolor}
\usepackage[final]{acl}
\usepackage{times}
\usepackage{latexsym}
\usepackage{CJKutf8}

\usepackage{microtype}
\usepackage{tabularx}
\usepackage{amsmath}
\usepackage{amssymb}
\usepackage{dsfont}
\usepackage{xcolor}
\usepackage{arydshln}
\usepackage{tikz}
\usetikzlibrary{arrows, decorations.text, shapes.geometric, positioning, decorations.pathreplacing, calligraphy}
\usepackage{pgfplots}
\usepackage{pgfmath}
\usepackage{pgffor}
\pgfplotsset{compat=1.17}
\usepackage{subcaption}
\usepackage{mathtools}
\usepackage{multirow}
\usepackage{booktabs}
\usepackage{pifont}
\usepackage{ifthen}
\usepackage{enumitem}
\usepackage{enumitem}
\usepackage{amssymb}
\usepackage{amsfonts}
\usepackage{amsmath}
\usepackage{bbm}
\usepackage{arydshln}
\usepackage{makecell}
\usepackage[T1]{fontenc}
\usepackage{multirow}
\usepackage{booktabs}
\usepackage{bbding}
\usepackage{pifont}
\usepackage[utf8]{inputenc}
\usepackage[capitalize]{cleveref}
\usepackage{microtype}
\usepackage{inconsolata}
\usepackage{graphicx}
\usepackage{float} 
\usepackage{stfloats}
\usepackage{adjustbox}
\usepackage[normalem]{ulem}
\useunder{\uline}{\ul}{}
\usepackage{color}
\usepackage{dsfont}
\definecolor{bleudefrance}{rgb}{0.19, 0.55, 0.91}
\definecolor{cadmiumgreen}{rgb}{0.0, 0.42, 0.24}
\definecolor{chromeyellow}{rgb}{1.0, 0.65, 0.0}
\definecolor{darkpastelpurple}{rgb}{0.59, 0.44, 0.84}
\definecolor{electriclavender}{rgb}{0.96, 0.73, 1.0}
\definecolor{fluorescentorange}{rgb}{1.0, 0.75, 0.0}
\definecolor{lightseagreen}{rgb}{0.13, 0.7, 0.67}
\definecolor{mauvelous}{rgb}{0.94, 0.6, 0.67}
\definecolor{mayablue}{rgb}{0.45, 0.76, 0.98}
\definecolor{nadeshikopink}{rgb}{0.96, 0.68, 0.78}
\definecolor{portlandorange}{rgb}{1.0, 0.35, 0.21}
\definecolor{persianorange}{rgb}{0.85, 0.56, 0.35}
\definecolor{peru}{rgb}{0.8, 0.52, 0.25}
\usepackage{colortbl}
\usepackage{bbm}

\newcommand{\ie}{\emph{i.e., }}

\newcommand{\cf}{\emph{cf. }}

\title{HellaSwag-Pro: A Large-Scale Bilingual Benchmark for Evaluating the Robustness of LLMs in Commonsense Reasoning}
\author{Xiaoyuan Li\textsuperscript{1},
        Moxin Li\textsuperscript{3}\thanks{Corresponding authors. Work done when Xiaoyuan Li and Keqin Bao were intern at Alibaba Group.},
        Rui Men\textsuperscript{2}\footnotemark[1] ,
        Yichang Zhang\textsuperscript{2},
        Keqin Bao\textsuperscript{1},
        Wenjie Wang\textsuperscript{1},\\
        \textbf{Fuli Feng\textsuperscript{1}},
        \textbf{Dayiheng Liu\textsuperscript{2}},
        \textbf{Junyang Lin\textsuperscript{2}}\\
        University of Science and Technology of China\textsuperscript{1} \\
        Alibaba Group\textsuperscript{2} \\
        National University of Singapore\textsuperscript{3} \\
        }
\begin{document}
\maketitle
\input{latex/sections/1_abstract}
\input{latex/sections/2_introduction_v2}

\input{latex/sections/3_chinese_hellaswag_1215}

\input{latex/sections/4_hellaswagpro_1215}

\input{latex/sections/5_experiments_1215}

\input{latex/sections/6_related_work}
\input{latex/sections/7_conclusion}
\input{latex/sections/limitations}
\bibliography{acl_latex}
\input{latex/sections/appendix}
\end{document}

%% file: latex/sections/1_abstract.tex
\begin{abstract}
Large language models (LLMs) have shown remarkable capabilities in commonsense reasoning; however, some variations in questions can trigger incorrect responses.
\textit{Do these models truly understand commonsense knowledge, or just memorize expression patterns?}
To investigate this question, we present the first extensive robustness evaluation of LLMs in commonsense reasoning.
We introduce HellaSwag-Pro, a large-scale bilingual benchmark consisting of 11,200 cases, by designing and compiling seven types of question variants. 
To construct this benchmark, we propose a two-stage method to develop Chinese HellaSwag, a finely annotated dataset comprising 12,000 instances across 56 categories. 
We conduct extensive experiments on 41 representative LLMs, revealing that these LLMs are far from robust in commonsense reasoning. 
Furthermore, this robustness varies depending on the language in which the LLM is tested.
This work establishes a high-quality evaluation benchmark, with extensive experiments offering valuable insights to the community in commonsense reasoning for LLMs.


\begin{table*}[!h]
\centering
\setlength{\abovecaptionskip}{0.05cm}
\setlength{\belowcaptionskip}{0cm}
\scalebox{0.72}{
\begin{tabular}{lll}
\hline
\multicolumn{1}{l}{Variant Type} & Context& Choices\\ \hline
\multicolumn{1}{l}{\multirow{4}{*}{Initial data}} & \multirow{4}{*}{\begin{tabular}[c]{@{}l@{}}A lady walks to a barbell. She bends down \\ and grabs the pole. The lady\end{tabular}} & A. stands and lifts the weight over her head. \\ 
\multicolumn{1}{l}{}&& B. swings and lands in her arms.\\ 
\multicolumn{1}{l}{}&& C. pulls the barbell forward. \\ 
\multicolumn{1}{l}{}&& D. pulls a rope attached to the barbell.\\ \hline
\multicolumn{1}{l}{\multirow{4}{*}{\textcolor{bleudefrance}{Problem restatement}}} & \multirow{4}{*}{\begin{tabular}[c]{@{}l@{}}\textcolor{bleudefrance}{A woman approaches a weightlifting bar.} \\ \textcolor{bleudefrance}{She lowers her body and grasps the metal} \\ \textcolor{bleudefrance}{rod. The woman}\end{tabular}} &\textcolor{bleudefrance}{A. rises and hoists the barbell above her head.} \\ 
\multicolumn{1}{l}{}&&B. swings and lands in her arms.\\ 
\multicolumn{1}{l}{}&&C. pulls the barbell forward. \\ 
\multicolumn{1}{l}{}&&D. pulls a rope attached to the barbell.\\ \hline
\multicolumn{1}{l}{\multirow{4}{*}{\textcolor{cadmiumgreen}{Reverse conversion}}} & \multirow{4}{*}{\begin{tabular}[c]{@{}l@{}}The lady stands and lifts the weight over \\ her head. \textcolor{cadmiumgreen}{Which could be the most} \\ \textcolor{cadmiumgreen}{possible context for this action?}\end{tabular}}&\textcolor{cadmiumgreen}{A. A lady walks to a barbell. She bends down and grabs the pole.}\\
\multicolumn{1}{l}{}&&\textcolor{cadmiumgreen}{B. A lady positions herself at the squat rack. She lowers her body } \\ &&\textcolor{cadmiumgreen}{ \quad before rising steadily.} \\ 
\multicolumn{1}{l}{}&&\textcolor{cadmiumgreen}{C. A lady approaches the kettlebell set. She swings the weight} \\ && \textcolor{cadmiumgreen}{\quad forcefully between her legs.
}\\ 
\multicolumn{1}{l}{}&&\textcolor{cadmiumgreen}{D. A lady stands beside the bench press station. She lies down} \\&& \textcolor{cadmiumgreen}{\quad and lifts the barbell from her chest. }\\ \hline
\multirow{4}{*}{\textcolor{chromeyellow}{Causal inference}} & \multirow{4}{*}{\begin{tabular}[c]{@{}l@{}}A lady walks to a barbell. She bends down \\ and grabs the pole. The lady stands and \\ lifts the weight over her head. \textcolor{chromeyellow}{Which could} \\ \textcolor{chromeyellow}{be the most possible reason for this action?}\end{tabular}} &\textcolor{chromeyellow}{A. She is performing a weightlifting exercise.}\\ 
&&\textcolor{chromeyellow}{B. She is using the barbell as a decoration for an event.}\\ 
&&\textcolor{chromeyellow}{C. She is moving the barbell to a different location in the gym.}\\ 
&&\textcolor{chromeyellow}{D. She is cleaning the barbell after a workout session.} \\ \hline
\multirow{4}{*}{\textcolor{darkpastelpurple}{Sentence ordering}}& \multirow{4}{*}{\begin{tabular}[c]{@{}l@{}}1. She bends down and grabs the pole. \\ 2. A lady walks to a barbell. \\ 3. The lady stands and lifts the weight over \\ her head. \textcolor{darkpastelpurple}{Which is the correct order?}\end{tabular}}&\textcolor{darkpastelpurple}{A. 2-1-3}\\ 
&&\textcolor{darkpastelpurple}{B. 3-1-2}\\ 
&&\textcolor{darkpastelpurple}{C. 2-3-1}\\ 
&&\textcolor{darkpastelpurple}{D. 1-3-2}\\ \hline
\multirow{4}{*}{\textcolor{mauvelous}{Scenario refinement}}& \multirow{4}{*}{\begin{tabular}[c]{@{}l@{}}A lady walks to a barbell. She bends down \\ and grabs the pole. The lady \textcolor{mauvelous}{hesitates for} \\ \textcolor{mauvelous}{a moment, then changes her mind. Instead} \\ \textcolor{mauvelous}{of lifting the barbell, she}\end{tabular}}&A. swings and lands in her arms.\\ 
&&B. stands and lifts the weight over her head. \\ 
&&C. pulls the barbell forward.\\ 
&&D. pulls a rope attached to the barbell.\\ \hline
\multirow{4}{*}{\textcolor{lightseagreen}{Negative transformation}} & \multirow{4}{*}{\begin{tabular}[c]{@{}l@{}}A woman approaches a weightlifting bar. \\ She lowers her body and grasps the metal \\ rod. The lady \textcolor{lightseagreen}{will not}\end{tabular}} &A. swings and lands in her arms.\\ 
&&B. stands and lifts the weight over her head.\\ 
&&\textcolor{lightseagreen}{C. bend her knees and lift the barbell.} \\ 
&&\textcolor{lightseagreen}{D. adjust her grip and lift the weight.} \\ \hline
\multirow{5}{*}{\textcolor{peru}{Critical testing}} & \multirow{5}{*}{\begin{tabular}[c]{@{}l@{}}A lady walks to a barbell. She bends down\\ and grabs the pole. The lady \textcolor{peru}{suddenly} \\ \textcolor{peru}{realizes she forgot her weightlifting gloves}\\ \textcolor{peru}{and decides to postpone her workout.} \\ \textcolor{peru}{The lady}\end{tabular}} &A. stands and lifts the weight over her head. \\ 
&&B. swings and lands in her arms.\\ 
&&C. pulls the barbell forward. \\ 
&&D. pulls a rope attached to the barbell.\\ 
&&\textcolor{peru}{E. None of the above four options are suitable.} \\ \hline
\end{tabular}
}
\caption{Examples of the seven variants we adopt for an initial question, with the correct answer unchanged as (A). Modifications are highlighted in different colors for clarity.}
\label{variant type}
\vspace{-10pt}
\end{table*}

\end{abstract}

%% file: latex/sections/2_introduction_v2.tex
\section{Introduction}

Commonsense reasoning is a crucial part of intelligence, involving contextual understanding, implicit knowledge, and logical deduction (\citealp{liu2004conceptnet, cambria2011isanette, davis2015commonsense}). 
Recent studies have focused on enhancing these capabilities in LLMs, achieving impressive performance (\citealp{yang2024qwen2, openai2024gpt, team2024gemini}). However, even slight changes to questions can lead to incorrect responses from the same models.
For instance, in binary commonsense questions, human naturally recognizes both correct and incorrect options through a single inference process, while LLMs, though able to identify the correct answer, struggle to reason about why the alternative is wrong ~\cite{balepur2024s}. 
Therefore, we ask the question:
\textit{Does this high-level performance stem from a genuine understanding of commonsense knowledge, or is it simply a result of memorizing specific expression patterns in pre-training data?}

\begin{figure}[t]   
\centering
\setlength{\abovecaptionskip}{-0.10cm}
\setlength{\belowcaptionskip}{0cm}
\includegraphics[width=\linewidth,scale=1.00]{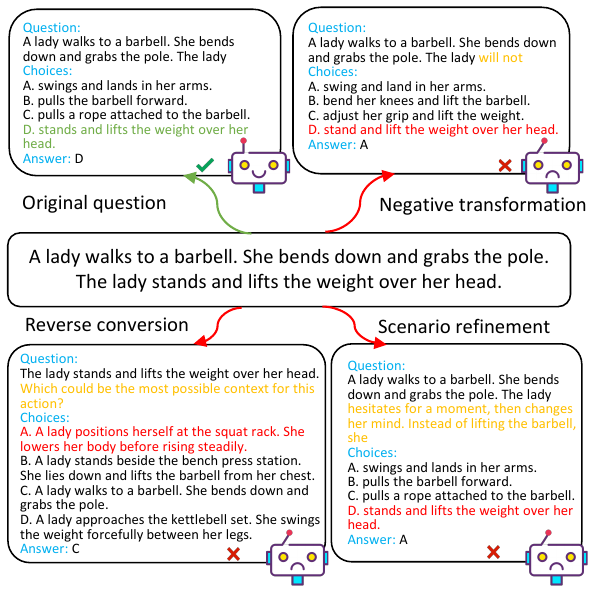}
\caption{Comparison of GPT-4o's responses to an original question and its several meaning-preserving variants. GPT-4o successfully handles the original question but struggles with its variants on the same knowledge.}
\label{example}
\vspace{-10pt}
\end{figure}

To answer this question, an effective approach is to systematically evaluate the robustness of LLMs in answering commonsense reasoning questions. 
As illustrated in Figure~\ref{example}, we find that GPT-4o correctly answers an original question but fails on its variants, \ie questions about the same commonsense knowledge but in different reasoning forms, such as reverse conversion. 
This indicates that GPT-4o has not fully grasped the commonsense knowledge behind the question; a genuine understanding of commonsense knowledge should be able to generalize to these question variants. 

However, existing benchmarks do not yet support a thorough evaluation of LLM robustness in commonsense reasoning.
Most work evaluates LLMs on general benchmarks~\cite{zellers2019hellaswag, talmor2019commonsenseqa, mihaylov2018can}, or in specific domains of commonsense knowledge~\cite{zhou2019going, qin2021timedial, bisk2020piqa}. Although some efforts have considered the robustness of commonsense reasoning, they either focus on whether models can learn genuine question-answer correlations under initial questions~\cite{jia2017adversarial,branco2021shortcutted}, or examine only one type of simplistic question variant such as question paraphrasing~\cite{zhou2021rica,ismayilzada-etal-2023-crow,balepur2024s}, lacking investigation into robustness across diverse and complex variants. 



To address this gap, we present the first extensive evaluation on the robustness of commonsense reasoning for LLMs, starting with dataset construction. 
Firstly, recognizing that existing benchmarks are predominantly in English, which limits the assessment of non-English LLMs (\citealp{davis2023benchmarks}), we develop a Chinese commonsense reasoning dataset based on the widely-used HellaSwag benchmark~\cite{zellers2019hellaswag}, containing 12,000 questions. Specifically, we design 56 fine-grained categories, and propose a two-stage data annotation method including initial dataset generation and difficult sample replacement. 
Secondly, we design and compile seven variants from existing studies (\cf Table~\ref{variant type}), which can be characterized under Bloom Cognitive Model (\cf Appendix ~\ref{bloom_model}).
We then create the variants for the Chinese and English versions of HellaSwag, obtaining HellaSwag-Pro, a high-quality human-verified dataset with 11,200 variants from 1,600 original questions. 

Using HellaSwag-Pro, we conduct a comprehensive evaluation on the robustness of 41 closed-source and open-source LLMs with nine different prompt strategies. 
We derive several key findings:
\textbf{(1)} 
All LLMs are far from robust in commonsense reasoning tasks, as evidenced by their poor performance on question variants and the significant gap compared to human performance. 
Nevertheless, GPT-4o achieves the best robustness among all the evaluated LLMs. 
\textbf{(2)} Among all types of variants, negative transformation is the most challenging, with an average accuracy of only 9.01\%, while problem restatement poses minimal difficulty. 
\textbf{(3)} LLMs achieve the best robustness in the language on which they were adequately trained. 
\textbf{(4)} Incorporating chain-of-thought (CoT) reasoning and using few-shot demonstrations can strengthen their robustness. 


Our contributions are three-fold. 
\textbf{(1)} We present the first extensive evaluation on the robustness of commonsense reasoning for LLMs by designing and compiling seven types of variants. 
\textbf{(2)} We have developed a bilingual, large-scale, human-annotated benchmark for evaluating LLM robustness in commonsense reasoning, which will be publicly released upon acceptance. 
\textbf{(3)} We conduct in-depth experiments on 41 representative LLMs with diverse prompts, yielding critical insights. 


%% file: latex/sections/3_chinese_hellaswag_1215.tex
\section{Chinese HellaSwag} \label{sec:chinese_hellaswag}
\begin{figure}[t]   
\centering
\setlength{\abovecaptionskip}{0.1cm}
\setlength{\belowcaptionskip}{0cm}
\includegraphics[width=\linewidth,scale=0.8]{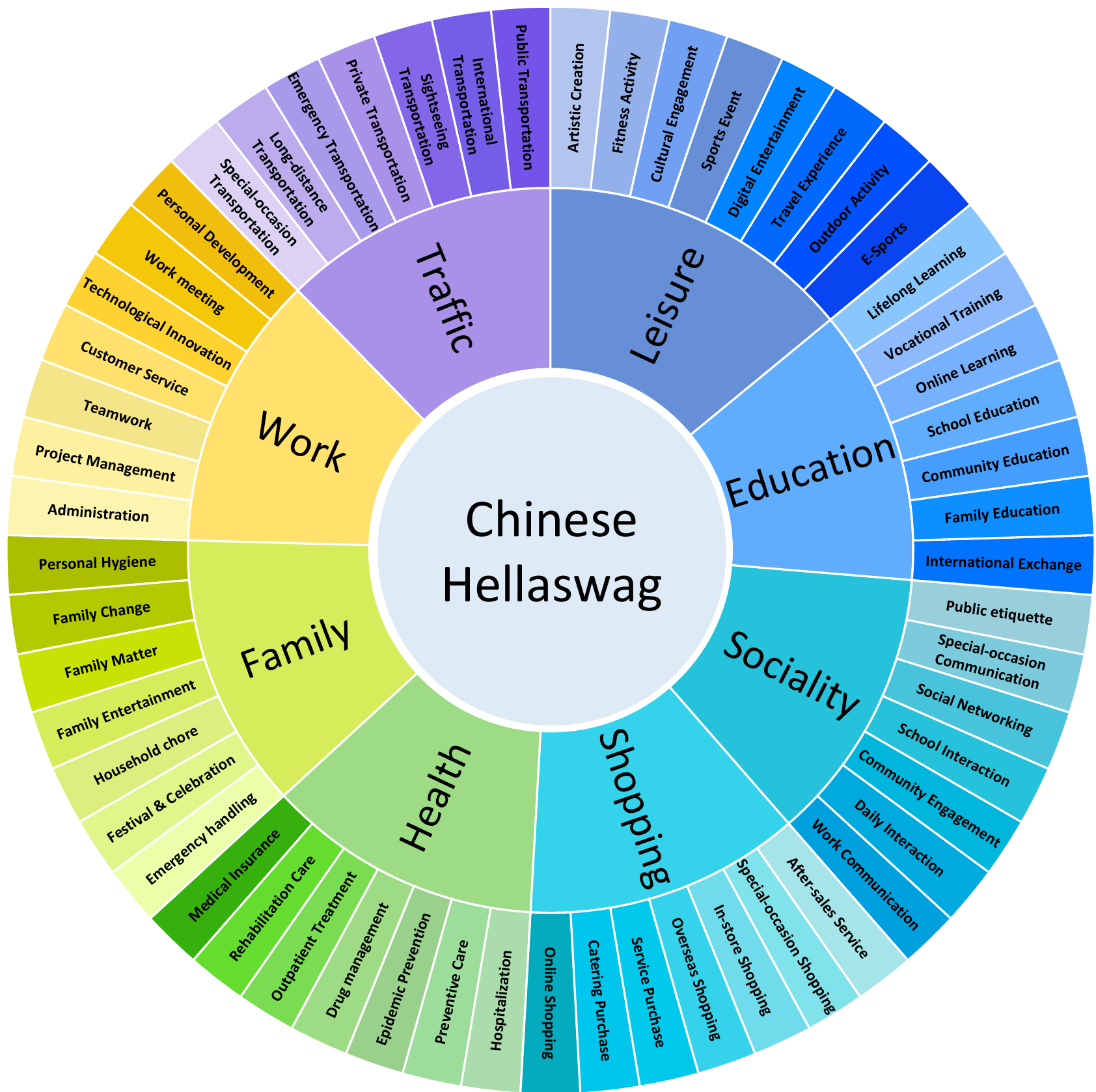}
\caption{Overview of Chinese HellaSwag categories. There are seven broad categories in total, each with eight detailed subcategories.}
\label{cn_hellaswag}
\vspace{-8pt}
\end{figure}

Given the limitation that most existing benchmarks for commonsense reasoning are in English, we begin by building a Chinese benchmark for commonsense reasoning that captures unique aspects of Chinese cultural context.
Firstly, we structure the dataset following the format of HellaSwag \cite{zellers2019hellaswag}, a widely recognized English commonsense reasoning benchmark, which consists of multiple-choice questions with four answer options.
Secondly, to minimize manual effort, we incorporate Qwen-Max \cite{yang2024qwen2}, a state-of-the-art Chinese LLM, into the dataset construction process.
Finally, to enhance the diversity of the dataset, we develop a hierarchical taxonomy of commonsense knowledge, as shown in Figure~\ref{cn_hellaswag}. Our taxonomy consists of seven broad categories summarized from existing literature \cite{zellers2019hellaswag, koupaee2018wikihow, caba2015activitynet}, each containing eight subcategories. We aim to construct our dataset based on the taxonomy, where we inject the categorical information into the instruction for LLM generation. 

\begin{figure}[t] 
\centering
\setlength{\abovecaptionskip}{0.05cm}
\setlength{\belowcaptionskip}{0cm}
\includegraphics[width=\linewidth,scale=0.95]{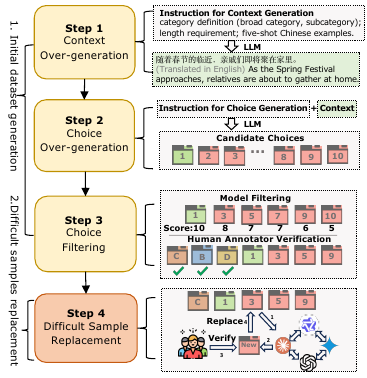}
\caption{The two-stage data construction pipeline for Chinese HellaSwag. See an example in Table \ref{case}.}
\label{overview}
\end{figure}

We propose a two-stage data construction pipeline, \textbf{\textit{initial dataset generation}} and \textbf{\textit{difficult sample replacement}}, as shown in Figure \ref{overview}. 

\paragraph{Initial Dataset Generation} 
In this stage, we employ an over-generate-then-filter (\citealp{yuan2023distilling}) approach, \ie generating excessive question-answer pairs and filtering for high quality ones, to obtain the initial dataset. The generation of the initial dataset consists of three steps. 

\begin{itemize}[leftmargin=*]
\setlength{\itemsep}{0pt}
    \item \textbf{Step 1: Context over-generation.}
    We employ the LLM to create a Chinese context of the question via in-context learning \cite{brown2020language}, incorporating category information, length requirement, and carefully crafted five-shot Chinese examples similar to HellaSwag. 
    For the length requirement, we assign three tiers: short (under 20 characters), medium (20-40 characters), and long (over 40 characters). 
    We then filter the generated contexts based on character count and Jaccard similarity, eliminating samples that do not meet the length requirement or are too similar to other samples. 
    \item \textbf{Step 2: Choice over-generation. } For each context, we instruct the LLM to over-generate ten potential choices, forming a question.   
    
    \item \textbf{Step 3: Choice filtering.} 
    We instruct the LLM to evaluate each question on a ten-point scale and select six choices: one correct answer (10 points) and five high-scoring incorrect choices. 
    Then, human annotators select four choices, ensuring a single correct answer and three challenging incorrect choices, and check the category labels for the question. 
    After the LLM scoring, we obtain 12,960 samples, which human annotators further refine to 12,287. To maintain category balance, we ultimately select 12,000 samples, allocating 1,500 to each broad category. 

\end{itemize}

\paragraph{Difficult Sample Replacement}
After initial dataset generation, we notice that some incorrect choices are rather simple for LLMs to identify, making the Chinese HellaSwag much easier than its English counterpart. 
Following the adversarial filtering \cite{zellers2018swag}, we use a human-in-the-loop adversarial filtering method (\textbf{Step 4}) to further enhance the dataset's difficulty.
This process involves using a generator LLM to rewrite existing incorrect choices into more challenging ones, and then evaluating the generated choices on multiple discriminator LLMs. 
If the generated choice successfully misleads the discriminator LLMs, we replace the original choice with the newly generated one. 
Finally, human annotators filter out the generated choices that are too difficult for humans to identify (see detail in Appendix~\ref{cn_hellaswag_annotation}). We iterative perform this process until the Chinese HellaSwag achieves accuracy comparable to the English HellaSwag, resulting in replacing 2451 samples. 
The dataset statistics of the Chinese HellaSwag can be found in Table~\ref{tab:statistics}. The complete evaluation of it is given in Appendix ~\ref{cn_eval}. 

\begin{table}[t]
    \centering
\setlength{\abovecaptionskip}{0.05cm}
\setlength{\belowcaptionskip}{0cm}
\resizebox{0.8\linewidth}{!}{
\begin{tabular}{l|llll}
\toprule
Length Type  & Long  & Medium & Short & Total  \\
\midrule
\# Questions & 4,179 & 4,033  & 3,788 & 12,000 \\
\bottomrule
\end{tabular} }
    \caption{Statistics for Chinese HellaSwag.}
    \label{tab:statistics}
\vspace{-10pt}
\end{table}

%% file: latex/sections/4_hellaswagpro_1215.tex
\section{HellaSwag-Pro}

Based on the English-Chinese HellaSwag datasets, we construct HellaSwag-Pro, the benchmark for extensive robustness evaluation of commonsense reasoning. 
We begin by designing the seven-type question variants for robustness evaluation, then detail our data generation process. 


\subsection{Variant Types}
We aim to evaluate the robustness of commonsense reasoning on question variants of changed reasoning forms for the same commonsense knowledge. 
The rationality is that the diverse reasoning forms disables the reliance on superficial patterns, ensuring that correct answers from LLMs demonstrate a robust understanding of the underlying commonsense knowledge. 
Building on existing research \cite{guo2024exploring, ma2024kor, balepur2024s} and our own designs, we maintain seven types of variants, as detailed below. 

\begin{itemize}[leftmargin=*]
    \setlength{\itemsep}{0pt}
    \item \textbf{Problem restatement} aims to test the impact of textual description variations on model understanding. We rephrase the context and correct choice while keeping the incorrect choices unchanged, thereby increasing the difficulty of identifying the correct answer. 

    \item \textbf{Reverse conversion} evaluates the capability for reverse reasoning, \ie inferring the context from the outcome, which has been shown to be challenging for LLMs \cite{guo2024exploring}. We utilize the original correct choice as the context, the original context as the correct choice, and generate three additional incorrect choices. 

    \item \textbf{Causal inference} evaluates the understanding of the causality of the event.  We merge the context and the correct choice and ask for the reason. We generate one correct reason and produce three additional incorrect reasons as the choices.

    \item \textbf{Sentence ordering} focuses on the understanding of inter-sentence relationships, such as progression or contrast.     
    We concatenate the context and correct choice into a complete paragraph, then shuffle the order of the sentences. The correct choice refers to the original sentence ordering. 

    \item \textbf{Scenario refinement} investigates the ability to infer counterfactual situations \cite{ma2024kor}. We select a relatively plausible choice from the original incorrect choices, then minimally modify the context to make this choice correct, where the original correct choice becomes incorrect. 

    \item \textbf{Negation transformation} examines the robustness to negation, a known challenge for LLMs \cite{balepur2024s}. This involves altering the context by introducing negations, such as changing "the man will" to "the man will not." In this transformation, the least plausible choice in the original question becomes the correct answer for the variant, while the original correct answer is retained, and two additional plausible options are generated as distractors. 
    
    \item \textbf{Critical testing} evaluates the model's ability to abstain from answering when the context lacks sufficient information to determine a correct answer. We remove key details from the context to make all original choices invalid. 
    We keep the context minimally modified to increase difficulty. 
    A new choice, \emph{``None of the above four options are suitable''},  is introduced as the correct choice. 
\end{itemize}

\subsection{Data Generation}

To construct these variants, we also employ Qwen-Max due to its comparatively strong language ability in reforming the questions. 
We design in-context examples and instructions with transformation rules to guide Qwen-Max to generate the question variants (\cf Appendix~\ref{appe:hellaswag_pro_annotation}). 
However, we observe that Qwen-Max is not consistently reliable, exhibiting issues such as: 
(1) generating variants inconsistent with the definitions,  
(2) producing multiple correct choices or overly simple incorrect choices, 
and (3) generating invalid contexts, particularly in \emph{scenario refinement}. 

To tackle these issues, we leverage manual quality control over the generated data. 
For \emph{reverse conversion} and \emph{causal inference}, we adopt an over-generate-then-filter approach (\cf Section~\ref{sec:chinese_hellaswag}) to control the correctness and the quality of the generated choices. 
Finally, we conduct comprehensive manual verification of all variants generated to ensure data quality. We initially generate 24,260 variants, and eventually filter down to 11,200 high-quality variants from 1,600 original questions. 

%% file: latex/sections/5_experiments_1215.tex
\section{Experiment}
In this section, we conduct extensive experiments to evaluate the performance of various LLMs on our HellaSwag-Pro benchmark. Our study is guided by three key research questions:
\textbf{RQ1}: How do different LLMs perform across all variants?
\textbf{RQ2}: What is the relative difficulty of different variants?
\textbf{RQ3}: Which prompting strategies yield the best robustness in LLMs?

\subsection{Experimental Setup} 
\paragraph{Model Selection and Implementation Details}
We select 41 representative closed-source and open-source LLMs. 
For English LLMs, we use GPT-4o \cite{DBLP:journals/corr/abs-2303-08774}, Claude-3.5-Sonnet \cite{anthropic2024claude}, Gemini-1.5-Pro \cite{DBLP:journals/corr/abs-2312-11805}, Mistral series \cite{DBLP:journals/corr/abs-2310-06825}, Llama3 series \cite{DBLP:journals/corr/abs-2407-21783} and Gemma2 series \cite{riviere2024gemma}. 
For Chinese LLMs, we use Qwen-Max \cite{qwen}, Qwen2.5 series \cite{yang2024qwen2}, InternLM2.5 series \cite{2023internlm}, Yi1.5 series \cite{DBLP:journals/corr/abs-2403-04652}, Baichuan2 series \cite{DBLP:journals/corr/abs-2309-10305} and DeepSeek series \cite{DBLP:journals/corr/abs-2401-02954}. 

We integrate both the Chinese HellaSwag and HellaSwag-Pro into the \texttt{lm-evaluation-harness} platform \cite{eval-harness}. For the open-source models, we use the default settings of the platform: do\_sample is set to false and the temperature is set to the default value of the \texttt{hugging-face} library as 1.0. For the closed-source models, we set the temperature to 0.7. In addition, we set the maximum output length to 1024.

\paragraph{Prompting Strategy} \label{sec:prompting_strategy}
We design nine prompting strategies to evaluate the LLMs across different languages and number of demonstrations. 
\textbf {(1) Direct}: LLM takes the original dataset question directly as input\footnote{
For open-source models, the \textbf{Direct} approach follows the official HellaSwag implementation, computing the log-likelihood for each option and selecting the one with the highest value. We report the normalized accuracy to account for the option length. Other prompting strategies use a generation setup and report accuracy based on exact match.}.
\textbf{(2) CN-CoT}: LLM is instructed to perform CoT in Chinese, regardless of the language of the dataset.
\textbf{(3) EN-CoT}: LLM is instructed to perform CoT in English. 
\textbf{(4) CN-XLT}: LLM is instructed to first translate the English question into Chinese, then reason in Chinese.
\textbf{(5) EN-XLT}: LLM is instructed to first translate the Chinese question into English, then reason in English.
The last four strategies include both zero-shot and three-shot variants.

\begin{table*}[ht]
\centering
\setlength{\abovecaptionskip}{0.05cm}
\setlength{\belowcaptionskip}{0cm}
\setlength{\tabcolsep}{2pt}
\scalebox{0.67}{
\begin{tabular}{ccccccccccccc}
\hline
\multicolumn{1}{c|}{{ }}& \multicolumn{4}{c|}{Chinese}& \multicolumn{4}{c|}{English}& \multicolumn{4}{c}{AVG}\\ \cline{2-13} 
\multicolumn{1}{c|}{\multirow{-2}{*}{{ Model}}} & { OA(\%)$\uparrow$}& { ARA(\%)$\uparrow$} & {RLA(\%)$\downarrow$}& \multicolumn{1}{l|}{{CRA(\%)$\uparrow$}} & { OA(\%)$\uparrow$}& { ARA(\%)$\uparrow$} & { RLA(\%)$\downarrow$}& \multicolumn{1}{l|}{{CRA(\%)$\uparrow$}} & {OA(\%)$\uparrow$}& { ARA(\%)$\uparrow$} & {RLA(\%)$\downarrow$}& { CRA(\%)$\uparrow$} \\ \hline
\multicolumn{1}{c|}{{ Human}} & 96.41& 97.79& -1.38 & \multicolumn{1}{l|}{92.03}& 95.56& 96.04& -0.48 & \multicolumn{1}{l|}{90.02}& 95.99 & 96.92 & -0.93& 91.03 \\ 
\multicolumn{1}{c|}{{ Random}} & 25.00& 25.00& 0.00 & \multicolumn{1}{l|}{0.0015}& 25.00& 25.00& 0.00 & \multicolumn{1}{l|}{0.0015}& 25.00& 25.00& 0.00 & 0.0015 \\ \hline
\multicolumn{13}{c}{\textbf{\textit{Closed-source LLMs}}}\\ 
\multicolumn{1}{c|}{{ Qwen-Max}}& { 93.50} & { 84.82} & { 8.68}& \multicolumn{1}{l|}{{ 78.91}} & { 87.60} & { 62.61} & { 24.99} & \multicolumn{1}{l|}{{ 59.65}} & { 90.55} & { 73.72} & { 16.83} & { 69.28} \\ \hline
\multicolumn{13}{c}{\textbf{\textit{Open-source LLMs}}} \\ 
\multicolumn{1}{c|}{{ Qwen2.5-0.5B}}& { 60.75} & { 45.18} & { \textbf{15.57}} & \multicolumn{1}{l|}{{ 28.70}} & { 49.50} & { 38.21} & { \textbf{11.29}} & \multicolumn{1}{l|}{{ 20.57}} & { 55.13} & { 41.70} & { \textbf{13.43}} & { 24.64} \\
\multicolumn{1}{c|}{{ Qwen2.5-1.5B}}& { 63.25} & { 46.16} & { 17.09} & \multicolumn{1}{l|}{{ 29.89}} & { 56.88} & { 39.57} & { 17.30} & \multicolumn{1}{l|}{{ 23.48}} & { 60.06} & { 42.87} & { 17.20} & { 26.69} \\
\multicolumn{1}{c|}{{ Qwen2.5-3B}}& { 67.50} & { 48.75} & { 18.75} & \multicolumn{1}{l|}{{ 33.79}} & { 61.75} & { 39.98} & { 21.77} & \multicolumn{1}{l|}{{ 25.75}} & { 64.63} & { 44.37} & { 20.26} & { 29.77} \\
\multicolumn{1}{c|}{{ Qwen2.5-7B}}& { 67.63} & { 50.59} & { 17.04} & \multicolumn{1}{l|}{{ 35.62}} & { 65.63} & { 43.93} & { 21.70} & \multicolumn{1}{l|}{{ 30.77}} & { 66.63} & { 47.26} & { 19.37} & { 33.20} \\
\multicolumn{1}{c|}{{ Qwen2.5-14B}} & { 69.00} & { 51.41} & { 17.59} & \multicolumn{1}{l|}{{ 35.84}} & { 68.50} & { 45.20} & { 23.30} & \multicolumn{1}{l|}{{ 32.12}} & { 68.75} & { 48.30} & { 20.45} & { 33.98} \\
\multicolumn{1}{c|}{{ Qwen2.5-32B}} & { 69.75} & { 53.11} & { 16.64} & \multicolumn{1}{l|}{{ 37.54}} & { 70.00} & { 46.10} & { 23.90} & \multicolumn{1}{l|}{{ 32.68}} & { 69.88} & { 49.61} & { 20.27} & { 35.11} \\
\multicolumn{1}{c|}{{ Qwen2.5-72B}} & { 70.87} & { \textbf{54.75}} & { 16.12} & \multicolumn{1}{l|}{{ \textbf{39.64}}} & {{ 72.00}} & { \textbf{47.75}} & { 24.25} & \multicolumn{1}{l|}{{\textbf{ 35.12}}} & { \textbf{71.44}} & { \textbf{51.25}} & { 20.19} & { \textbf{37.38}} \\ \hdashline[0.5pt/5pt]
\multicolumn{1}{c|}{{ Baichuan2-7B}}& { 67.00} & { 46.16} & { 20.84} & \multicolumn{1}{l|}{{ 31.50}} & { 60.62} & { 39.04} & { 21.58} & \multicolumn{1}{l|}{{ 25.21}} & { 63.81} & { 42.60} & { 21.21} & { 28.36} \\
\multicolumn{1}{c|}{{ Baichua2-13B}}& { 69.13} & { 46.98} & { 22.15} & \multicolumn{1}{l|}{{ 33.45}} & { 64.62} & { 38.82} & { 25.80} & \multicolumn{1}{l|}{{ 26.07}} & { 66.88} & { 42.90} & { 23.97} & { 29.76} \\ \hdashline[0.5pt/5pt]
\multicolumn{1}{c|}{{ DeepSeek-7B}} & { 68.13} & { 47.96} & { 20.17} & \multicolumn{1}{l|}{{ 33.30}} & { 63.38} & { 40.39} & { 22.99} & \multicolumn{1}{l|}{{ 26.70}} & { 65.76} & { 44.18} & { 21.58} & { 30.00} \\
\multicolumn{1}{c|}{{ DeepSeek-67B}}& { \textbf{71.50}} & { 49.21} & { 22.29} & \multicolumn{1}{l|}{{ 35.89}} & { 71.37} & { 40.63} & { 30.75} & \multicolumn{1}{l|}{{ 29.71}} & \textbf{ 71.44} & { 44.92} & { 26.52} & { 32.80} \\ \hdashline[0.5pt/5pt]
\multicolumn{1}{c|}{{ InternLM2.5-1.8B}}& { 61.62} & { 42.07} & { 19.55} & \multicolumn{1}{l|}{{ 26.99}} & { 55.37} & { 38.46} & { 16.91} & \multicolumn{1}{l|}{{ 22.61}} & { 58.50} & { 40.27} & { 18.23} & { 24.80} \\
\multicolumn{1}{c|}{{ InternLM2.5-7B}}& { 67.25} & { 49.77} & { 17.48} & \multicolumn{1}{l|}{{ 34.57}} & { 69.50} & { 40.89} & { 28.61} & \multicolumn{1}{l|}{{ 29.75}} & { 68.38} & { 45.33} & { 23.04} & { 32.16} \\
\multicolumn{1}{c|}{{ InternLM2.5-20B}} & { 67.37} & { 48.08} & { 19.29} & \multicolumn{1}{l|}{{ 33.21}} & \textbf{ 73.62} & { 41.11} & { 32.51} & \multicolumn{1}{l|}{{ 31.23}} & { 70.50} & { 44.60} & { 25.90} & { 32.22} \\ \hdashline[0.5pt/5pt]
\multicolumn{1}{c|}{{ Yi1.5-6B}} & { 67.00} & { 49.59} & { 17.41} & \multicolumn{1}{l|}{{ 34.27}} & { 64.38} & { 39.37} & { 25.01} & \multicolumn{1}{l|}{{ 26.62}} & { 65.69} & { 44.48} & { 21.21} & { 30.45} \\
\multicolumn{1}{c|}{{ Yi1.5-9B}} & { 68.50} & { 50.18} & { 18.32} & \multicolumn{1}{l|}{{ 35.55}} & { 66.37} & { 39.58} & { 26.79} & \multicolumn{1}{l|}{{ 27.48}} & { 67.44} & { 44.88} & { 22.56} & { 31.52} \\
\multicolumn{1}{c|}{{ Yi1.5-34B}}& { 71.00} & { 52.23} & { 18.77} & \multicolumn{1}{l|}{{ 38.09}} & { 71.00} & { 40.75} & { 30.25} & \multicolumn{1}{l|}{{ 29.91}} & { 71.00} & { 46.49} & { 24.51} & { 34.00} \\ \hline
\end{tabular}
}
\caption{Results of existing \textbf{Chinese} LLMs on HellaSwag-Pro using \textbf{Direct} prompt. ``AVG'' indicates the average performance on Chinese and English parts of the dataset. 
The best results in each model category are \textbf{bolded}. }
\label{tab:chinese main experiment.}
\vspace{-10pt}
\end{table*}

\begin{table*}[ht]
\centering
\setlength{\abovecaptionskip}{0.05cm}
\setlength{\belowcaptionskip}{0cm}
\setlength{\tabcolsep}{2pt}
\scalebox{0.67}{
\begin{tabular}{ccccccccccccc}
\hline
\multicolumn{1}{c|}{{ }}& \multicolumn{4}{c|}{Chinese}& \multicolumn{4}{c|}{English}& \multicolumn{4}{c}{AVG}\\ \cline{2-13} 
\multicolumn{1}{c|}{\multirow{-2}{*}{{ Model}}} & { OA(\%)$\uparrow$}& { ARA(\%)$\uparrow$} & {RLA(\%)$\downarrow$}& \multicolumn{1}{l|}{{CRA(\%)$\uparrow$}} & { OA(\%)$\uparrow$}& { ARA(\%)$\uparrow$} & { RLA(\%)$\downarrow$}& \multicolumn{1}{l|}{{CRA(\%)$\uparrow$}} & {OA(\%)$\uparrow$}& { ARA(\%)$\uparrow$} & {RLA(\%)$\downarrow$}& { CRA(\%)$\uparrow$} \\ \hline
\multicolumn{13}{c}{\textbf{\textit{Closed-source LLMs}}}\\ 
\multicolumn{1}{c|}{{ GPT-4o}}& { 91.37} & { \textbf{81.97}} & { \textbf{9.40}}& \multicolumn{1}{l|}{{ \textbf{75.55}}} & { \textbf{88.63}} & { \textbf{70.17}} & { \textbf{18.46}} & \multicolumn{1}{l|}{{ \textbf{63.06}}} & { 90.00} & { \textbf{76.07}} & { \textbf{13.93}} & { \textbf{69.31}} \\
\multicolumn{1}{c|}{{ Claude-3.5}}& { \textbf{95.37}} & { 80.15} & { 15.22} & \multicolumn{1}{l|}{{ 75.04}} & { 85.11} & { 66.02} & { 19.08} & \multicolumn{1}{l|}{{ 57.20}} & {\textbf{ 90.24}} & { 73.09} & { 17.15} & { 66.12} \\
\multicolumn{1}{c|}{{ Gemini-1.5-Pro}}& { 90.62} & { 78.36} & { 12.26} & \multicolumn{1}{l|}{{ 70.48}} & { 87.75} & { 60.74} & { 27.01} & \multicolumn{1}{l|}{{ 58.27}} & { 89.19} & { 69.55} & { 19.63} & { 64.38} \\
\hline
\multicolumn{13}{c}{\textbf{\textit{Open-source LLMs}}} \\ 
\multicolumn{1}{c|}{{ Llama3-8B}} & { 59.13} & { 46.62} & { 12.51} & \multicolumn{1}{l|}{{ 28.23}} & { 66.25} & { 40.21} & { 26.04} & \multicolumn{1}{l|}{{ 27.34}} & { 62.69} & { 43.42} & { 19.27} & { 27.79} \\
\multicolumn{1}{c|}{{ Llama3-70B}}& { 65.75} & { 48.63} & { 17.12} & \multicolumn{1}{l|}{{ 32.70}} & { \textbf{72.50}} & { 41.27} & { 31.23} & \multicolumn{1}{l|}{{\textbf{ 30.63}}} & {\textbf{ 69.13}} & { 44.95} & { 24.18} & { 31.67} \\ \hdashline[0.5pt/5pt]
\multicolumn{1}{c|}{{ Mistral-7B-v0.1}} & { 57.75} & { 46.25} & { \textbf{11.50}} & \multicolumn{1}{l|}{{ 27.57}} & { 67.50} & { \textbf{41.52}} & { 25.98} & \multicolumn{1}{l|}{{ 28.93}} & { 62.63} & { 43.88} & { 18.74} & { 28.25} \\
\multicolumn{1}{c|}{{ Mixtral-8x7B-v0.1}} & { 63.62} & { 46.80} & { 16.82} & \multicolumn{1}{l|}{{ 30.82}} & { 69.75} & { 41.21} & { 28.54} & \multicolumn{1}{l|}{{ 29.39}} & { 66.69} & { 44.01} & { 22.68} & { 30.11} \\
\multicolumn{1}{c|}{{ Mixtral-8x22B-v0.1}}& { 66.00} & {\textbf{ 50.73}} & { 15.27} & \multicolumn{1}{l|}{{ \textbf{34.32}}} & { 72.12} & { 41.25} & { 30.87} & \multicolumn{1}{l|}{{ 30.61}} & { 69.06} & { \textbf{45.99}} & { 23.07} & { \textbf{32.47}} \\ \hdashline[0.5pt/5pt]
\multicolumn{1}{c|}{{ Gemma2-2B}}& { 61.88} & { 45.38} & { 16.51} & \multicolumn{1}{l|}{{ 29.02}} & { 59.62} & { 39.13} & { \textbf{20.50}} & \multicolumn{1}{l|}{{ 24.88}} & { 60.75} & { 42.25} & {\textbf{ 18.50}} & { 26.95} \\
\multicolumn{1}{c|}{{ Gemma2-9B}}& { \textbf{69.13}} & { 46.75} & { 22.38} & \multicolumn{1}{l|}{{ 33.29}} & { 64.88} & { 39.80} & { 25.08} & \multicolumn{1}{l|}{{ 26.91}} & { 67.01} & { 43.28} & { 23.73} & { 30.10} \\
\multicolumn{1}{c|}{{ Gemma2-27B}} & { 63.38} & { 48.52} & { 14.86} & \multicolumn{1}{l|}{{ 31.96}} & { 71.88} & { 40.91} & { 30.97} & \multicolumn{1}{l|}{{ 30.25}} & { 67.63} & { 44.71} & { 22.92} & { 31.11} \\ \hline
\end{tabular}
}
\caption{Results of existing \textbf{English} LLMs on HellaSwag-Pro using \textbf{Direct} prompt (Same settings as Table \ref{tab:chinese main experiment.}).  }
\label{tab:english main experiment.}
\vspace{-10pt}
\end{table*}
\paragraph{Evaluation Metric}
We consider four evaluation metrics to measure the performance and robustness of LLMs. 
Denote the original dataset $\mathcal{D} = \{(x, y)\}$, where $x$ and $y$ represent the question and the correct label, respectively. 
Denote the dataset of all seven-type variants $\mathcal{D}_r = \{(x', y')\}$, where each $(x', y')$ corresponds to an original $(x, y)$ in $\mathcal{D}$.  

\noindent %
\textbf{Original Accuracy (OA)} measures the accuracy on original questions.
\begin{equation}
\small
\label{eq1}
OA=\frac{\sum_{(x, y) \in \mathcal{D}} \mathds{1}[\text{LM}(x), y]}{|\mathcal{D}|}.
\end{equation}
\noindent %
\textbf{Average Robust Accuracy (ARA)} measures the average accuracy across all variants.
\begin{equation}
\small
\label{eq2}
ARA=\frac{\sum_{\left(x^{\prime}, y^{\prime}\right) \in \mathcal{D}_{r}} \mathds{1}\left[\text{LM}(x'), y'\right].}{\left|\mathcal{D}_{r}\right|}.
\end{equation}

\noindent %
\textbf{Robust Loss Accuracy (RLA)} refers to the performance gap between all variants and original questions, \ie the difference between OA and ARA. 
\begin{equation}
\small
\label{eq3}
RLA= OA - ARA.
\end{equation}
\noindent %
\textbf{Consistent Robust Accuracy (CRA)} refers to the joint accuracy of LLM correctly answering the variant and its original question, reflecting the LLM's genuine understanding of the knowledge. 
\begin{equation}
\small
\label{eq4}
CRA=\frac{\sum_{(x^{\prime}, y^{\prime}) \in \mathcal{D}_r}\mathds{1}[\text{LM}(x), y] \cdot \mathds{1}[\text{LM}(x^{\prime}), y^{\prime}]}{\left|\mathcal{D}_{r}\right|}.
\end{equation}

\subsection{LLM Performance (RQ1)}
\paragraph{Overall Performance}
The results for \textbf{Direct} prompting on all LLMs are listed in Table~\ref{tab:chinese main experiment.} and Table~\ref{tab:english main experiment.}\footnote{The results of instruct and chat models of Qwen2.5, LLaMA3 and Mixtral\_v0.1 series are shown in Appendix~\ref{detailed}.}. The main observations are as follows.

Firstly, all evaluated LLMs perform well in OA (e.g., in AVG OA, GPT-4o scores 90.00, and Claude-3.5 scores 90.24). 
However, all LLMs show a performance drop on variants, as evidenced by a positive AVG RLA value for all LLMs. 
In contrast, human receive a near-zero RLA value, suggesting that the question variants are not more challenging than the originals for human. 
This disparity further illustrates that current LLMs lack a true understanding of the commonsense knowledge and can easily be affected by the reasoning form. 

Secondly, comparing open-source and closed-source LLMs, closed-source models achieve larger OA, ARA and CRA scores and smaller average RLA scores than open-source LLMs, indicating better robustness in commonsense reasoning. 

Finally, when we compare models within the same series (e.g., Qwen2.5, Llama3), we observe that larger models often achieve higher scores on OA, ARA, and CRA. 
However, their RLA shows no consistent relationship with model size. Across different families, AVG RLA patterns vary - fluctuating with size in Qwen2.5 and Gemma3, while increasing with size in Yi1.5 and Llama3.
This indicates that larger model size does not guarantee better robustness.

\begin{figure}[t]
\centering
\setlength{\abovecaptionskip}{-0.10cm}
\setlength{\belowcaptionskip}{0cm}
\includegraphics[width=1.0\linewidth,scale=1.00]{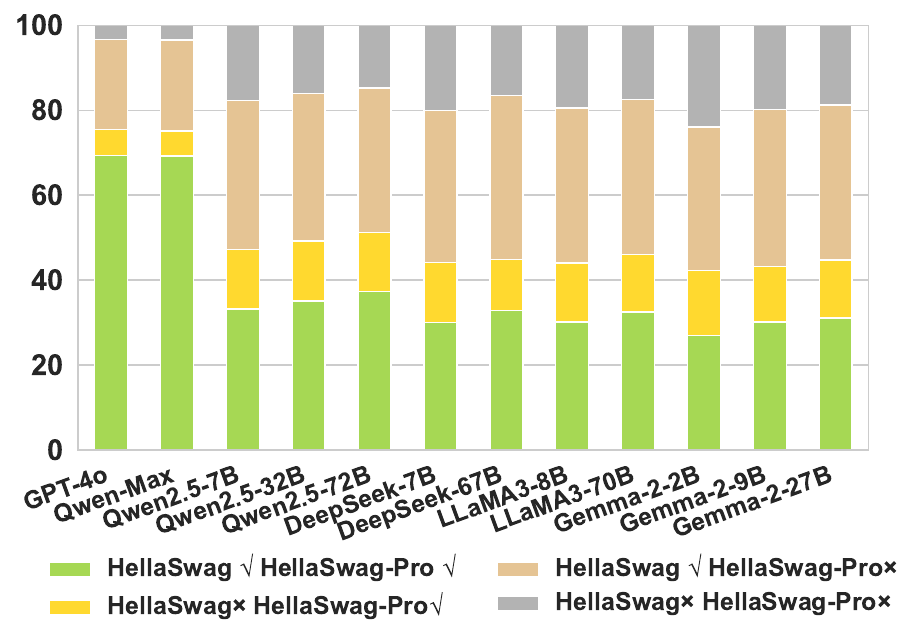}
\caption{Pairwise performance statistics of the original question and its variant. We use ``HellaSwag \ding{51} HellaSwag-Pro \ding{55}'' to denote that the LLM correctly answers the original question but fails on its variant. }
\label{consis}
\vspace{-10pt}
\end{figure}

\paragraph{Analysis on Reasoning Robustness}
To further analyze whether LLMs can maintain reasoning ability from the original question to its variant, Figure~\ref{consis} presents the pairwise performance statistic of the original question and its variant. 
For all LLMs, a significant proportion of variants are answered incorrectly despite LLMs being able to solve the source example. More specifically, closed-source LLMs like GPT-4o and Qwen-Max achieve a 69\% success rate on both HellaSwag and HellaSwag-Pro, with only 3\% failing both. In contrast, open-source LLMs struggle with around 30\% and 20\%, respectively. This shows that closed-source LLMs achieve better alignment between the performance of the original question and its variant, thus better robustness in reasoning ability.


\begin{figure}[t]
\centering
\setlength{\abovecaptionskip}{-0.1cm}
\setlength{\belowcaptionskip}{0cm}
\includegraphics[width=1.0\linewidth,scale=1.0]{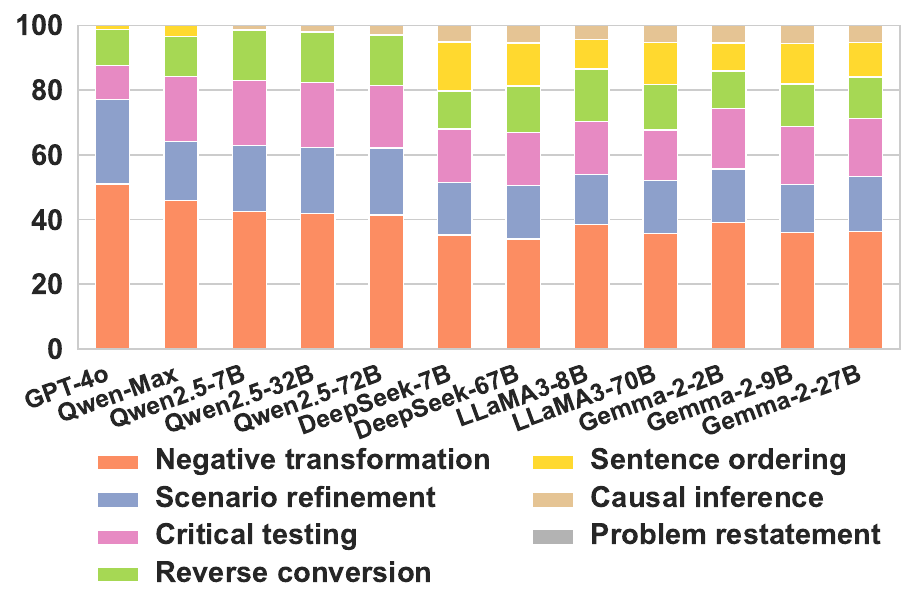}
\caption{Each variant's contribution to the RLA score.}
\label{fig:zhu}
\vspace{-10pt}
\end{figure}


\subsection{Variant Analysis (RQ2)}
To further analyze the robustness on different variants, we assess the contribution of each variant to the RLA score, as shown in Figure~\ref{fig:zhu}. A higher contribution indicates more non-robust in that type. The key observations are as follows:

\emph{Problem restatement}, \emph{causal inference}, and \emph{sentence ordering} are the least challenging. Almost all LLMs perform well on these variants particularly closed-source LLMs and Qwen2.5 series, indicating that LLMs can effectively handle these forms.

\emph{Reverse conversion} and \emph{critical testing} each contribute about 10\% to the RLA score. This indicates that current LLMs struggle to fully generalize to these variants, possibly because these variants do not largely exist in the training data. 

\emph{Negative transformation} and \emph{scenario refinement} are the two most difficult variants, with \emph{negative transformation} being particularly challenging. For almost all LLMs, these two variants account for more than 50\% of the RLA score. 
This might be due to statistical bias in these two types of data during pre-training and the exploitation of shortcuts in the corpus~\cite{chen-etal-2023-say, wu-etal-2024-reasoning}.

\begin{figure*}[t]
\centering
\setlength{\abovecaptionskip}{0.05cm}
\setlength{\belowcaptionskip}{0cm}
\includegraphics[width=1.0\linewidth,scale=1.0]{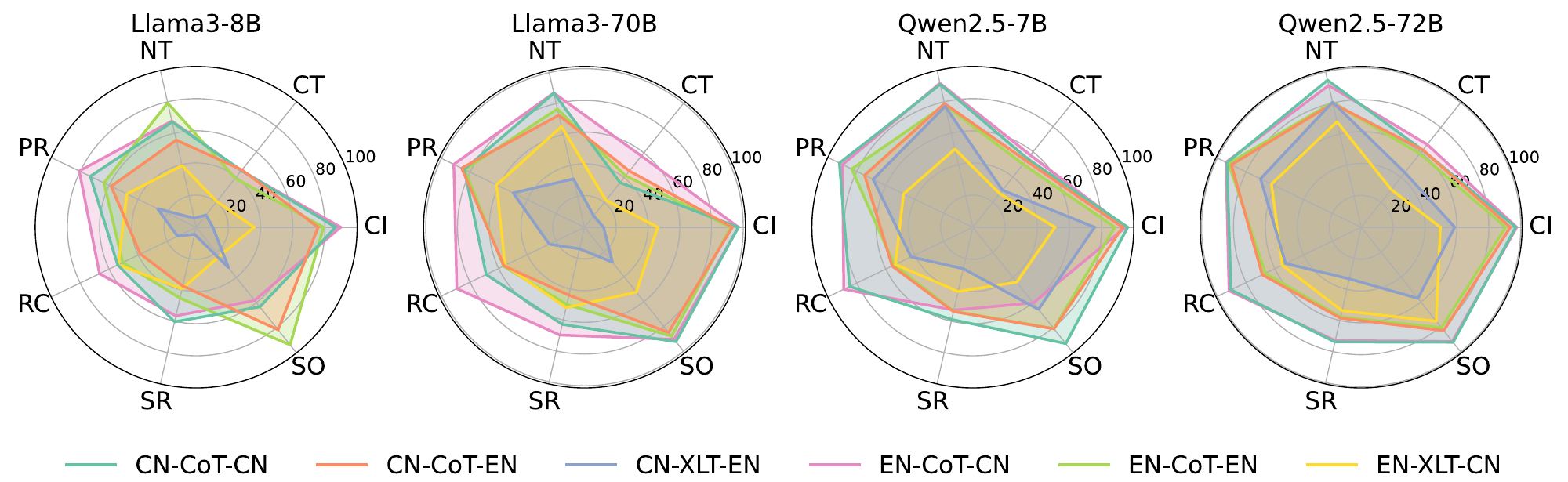}
\caption{Performance on different 3-shot prompts. For the legend, the first two parts are the prompt name, and the third part is the dataset language. NT, CT, CI, SO, SR, RC, PR are the abbreviations for the variant names of Negation Transformation, Critical Testing, Causal Inference, Sentence Ordering, Scenario Refinement, Reverse Conversion and Problem Restatement. }
\label{xing}
\vspace{-10pt}
\end{figure*}

\subsection{Different Prompting Strategies (RQ3)}
To explore the impact of various prompting strategies on our benchmark, we test the performance of all LLMs under different prompting strategies (\cf Section~\ref{sec:prompting_strategy}). 
The results are summarized in Table~\ref{prompt}.

For both Chinese and English datasets, Chinese LLMs perform best under CN-CoT strategy with shots, followed closely by EN-CoT with shots, achieving overall scores of 67.36\% and 67.03\%, respectively. 
Conversely, English LLMs show optimal performance using EN-CoT approach with shots, attaining 67.55\% on the Chinese dataset and 60.36\% on the English one.
This shows that different LLMs favor the prompts in their native language. 
Besides, translating datasets into LLMs' native languages before reasoning does not enhance performance (e.g., 28.69\% for EN LLMs using EN-XLT with shots vs 41.69\% for EN LLMs using Direct).
This phenomenon is further illustrated in Figure~\ref{xing}.
\begin{table}[H]
\centering
\setlength{\abovecaptionskip}{0.05cm}
\setlength{\belowcaptionskip}{0cm}
\setlength{\tabcolsep}{8pt}
\scalebox{0.8}{
\begin{tabular}{cccccc}
\hline
\multicolumn{3}{c|}{Prompt}& \multicolumn{3}{c}{LLM} \\
Strategy & Language & \multicolumn{1}{l|}{\#shot} & CN  & EN  & AVG \\ \hline
\multicolumn{6}{c}{\textbf{\textit{Chinese HellaSwag-Pro}}} \\
Direct& -& \multicolumn{1}{l|}{0}  & 48.95& 41.16& 45.06\\
\hdashline[0.5pt/5pt]
CoT  & CN& \multicolumn{1}{l|}{3}  & \textbf{71.04} & 51.90& 61.47\\ 
CoT  & EN& \multicolumn{1}{l|}{3}  & 70.95& \textbf{67.55} & \textbf{69.25} \\
XLT  & EN& \multicolumn{1}{l|}{3}  & 41.48& 28.69& 35.09\\
\hdashline[0.5pt/5pt]
CoT  & CN& \multicolumn{1}{l|}{0}  & 44.82& 23.89& 34.36\\
CoT  & EN& \multicolumn{1}{l|}{0}  & 45.38& 31.39& 38.39\\
XLT  & EN& \multicolumn{1}{l|}{0}  & 28.57& 12.93& 20.75\\ \hline
\multicolumn{6}{c}{\textbf{\textit{English HellaSwag-Pro}}} \\
Direct& -& \multicolumn{1}{l|}{0}  & 47.46& 40.66& 44.06\\
\hdashline[0.5pt/5pt]
CoT  & CN& \multicolumn{1}{l|}{3}  & \textbf{63.67} & 47.24& 55.46\\
CoT  & EN& \multicolumn{1}{l|}{3}  & 63.12& \textbf{60.36} & \textbf{61.74} \\
XLT  & CN& \multicolumn{1}{l|}{3}  & 48.77& 16.61& 32.69\\
\hdashline[0.5pt/5pt]
CoT  & CN& \multicolumn{1}{l|}{0}  & 34.89& 18.25& 26.57\\
CoT  & EN& \multicolumn{1}{l|}{0}  & 42.41& 31.03& 36.72\\
XLT  & CN& \multicolumn{1}{l|}{0}  & 16.36& 11.22& 13.79\\ \hline
\multicolumn{6}{c}{\textbf{\textit{HellaSwag-Pro}}}\\
Direct& -& \multicolumn{1}{l|}{0}  & 48.21& 40.91& 44.83\\
\hdashline[0.5pt/5pt]
CoT  & CN& \multicolumn{1}{l|}{3}  & \textbf{67.36} & 49.57& 58.46\\
CoT  & EN& \multicolumn{1}{l|}{3}  & 67.03& \textbf{63.95} & \textbf{65.49} \\
XLT  & CN& \multicolumn{1}{l|}{3}  & 59.91& 34.26& 47.08\\
XLT  & EN& \multicolumn{1}{l|}{3}  & 52.30& 44.52& 48.41\\
\hdashline[0.5pt/5pt]
CoT  & CN& \multicolumn{1}{l|}{0}  & 39.86& 21.07& 30.46\\
CoT  & EN& \multicolumn{1}{l|}{0}  & 43.90& 31.21& 37.55\\
XLT  & CN& \multicolumn{1}{l|}{0}  & 30.59& 17.55& 24.07\\
XLT  & EN& \multicolumn{1}{l|}{0}  & 35.49& 21.98& 28.74  \\ \hline
\end{tabular}
}
\caption{Average ARA of all open-source LLMs on different prompting strategies. CN-LLMs contains 17 LLMs, and EN-LLMs contains 7 LLMs. The best results for each dataset are \textbf{bolded}. Detailed results for all evaluated models are provided in the Appendix \ref{detailed}.}
\label{prompt}
\vspace{-10pt}
\end{table}

Our findings differ from previous research \cite{huang-etal-2023-languages,shi2022language}, which suggested that translating non-English tasks into English (XLT) would perform better than using native languages. And these research only focused on English LLMs while overlooking Chinese LLMs. We find that LLMs perform better when reasoning directly in their native language compared to XLT, addressing this gap in previous research.


%% file: latex/sections/6_related_work.tex
\section{Related Work}

\paragraph{Commonsense Reasoning Evaluation} 
There are numerous benchmarks and datasets for commonsense reasoning, most of which are in English. 
Some studies focus on evaluating general commonsense knowledge \cite{zellers2019hellaswag,talmor2019commonsenseqa,OpenBookQA2018}. 
Others target specific aspects of commonsense reasoning\cite{zhou2019going,bisk2020piqa,sap2019socialiqa,lin2020birds,clark2018think,khot2020qasc,li2024evaluating}.
There are some Chinese datasets for commonsense reasoning \cite{sun2024benchmarking,shi2024corecode}. 
For instance, CHARM \cite{sun2024benchmarking} distinguishes between global commonsense and Chinese-specific commonsense but includes only a limited number of everyday commonsense cases. 
However, evaluations aimed at assessing the robustness of commonsense reasoning are still understudied. 

\paragraph{Datasets on Different Reasoning Forms}
There are several datasets relevant to our variant design. For reverse reasoning, ART \cite{DBLP:conf/iclr/BhagavatulaBMSH20}, $\delta$-NLI \cite{DBLP:conf/emnlp/RudingerSHBFBSC20}, and CLUTRR \cite{DBLP:conf/emnlp/SinhaSDPH19} explore different reasoning directions. FCR \cite{DBLP:journals/corr/abs-2204-07408} and NatQuest \cite{ceraolo2024analyzinghumanquestioningbehavior} evaluate causal reasoning, while TimeTravel \cite{DBLP:conf/emnlp/QinBHBCC19} focuses on counterfactual scenario refinement. Additionally, PoE \cite{balepur2024s} assesses reasoning involving negation. 
However, not all these datasets focus on commonsense reasoning, nor are they structured by original questions and their variants. Furthermore, they typically target limited reasoning types. Lastly, our dataset is large-scale and covers diverse commonsense knowledge. 

\paragraph{Robustness and Consistency in LLMs} 
Early work focuses on adversarial attacks, with developing evaluation methods for reading comprehension systems \cite{jia2017adversarial}, followed by universal adversarial triggers \cite{wallace2019universal}. The field then expands to examine spurious correlations, with revealing how models often exploit superficial patterns rather than engaging in genuine reasoning \cite{branco2021shortcutted,geirhos2020shortcut}. And \citealp{ross2022does} investigates whether self-explanation can mitigate these spurious correlations. Coherence and consistency evaluation advances through classifier assessment methods \cite{storks2021beyond} and analysis of accuracy-consistency trade-offs \cite{johnson2023much}. While these studies primarily address model robustness against adversarial attacks or spurious correlations, our work takes a novel approach by examining robustness in reasoning forms.

%% file: latex/sections/7_conclusion.tex
\section{Conclusion}

We conduct a systematic evaluation of the robustness of LLMs in commonsense reasoning in both Chinese and English. 
To facilitate this evaluation process, we introduce two large-scale, finely-annotated datasets: HellaSwag-Pro and Chinese HellaSwag. 
In addition, we design various prompts to evaluate 41 LLMs, offering several key findings that may advance the field of commonsense reasoning. 
We believe this work will serve as a valuable resource to support further research into the commonsense reasoning of LLMs.

%% file: latex/sections/limitations.tex
\section*{Acknowledgements}

This research was also supported by the advanced computing resources provided by the Supercomputing Center of the USTC.

\section*{Limitations}

The limitations of our work are as follows:

\begin{itemize}[leftmargin=*]
    \setlength{\itemsep}{0pt}
    \item Our work only addresses everyday commonsense reasoning and does not encompass specific types, such as temporal or physical commonsense knowledge. Evaluating the robustness of LLMs on these specific types of commonsense reasoning tasks will be our future work.

    \item HellaSwag-Pro is concentrated on assessing the robustness of LLMs in commonsense reasoning tasks and does not investigate the underlying reasons for observed performance declines.

    \item For the sake of evaluation convenience, our setup utilizes multiple-choice questions. We plan to study the open-ended questions in future work.
\end{itemize}

\section*{Ethics Statement}
This work requires manual annotation. We provide annotators with a salary above the local minimum hourly wage. We have also clearly informed them about the purpose of the data and the necessity to ensure that all the data in Hellaswag-Pro does not contain any social biases, ethical concerns, or privacy issues.

Additionally, we develop a challenging dataset for evaluating the robustness of commonsense reasoning in this work. It's important to emphasize that this dataset is intended solely for evaluation, not for training or fine-tuning purposes. We recognize that improper use of this dataset for model training or fine-tuning could lead to persistent inconsistencies in LLMs' understanding of commonsense knowledge, potentially creating a vicious cycle where more such datasets would be needed to address these issues. Therefore, we explicitly state that the intended use of this dataset is strictly limited to evaluation to prevent the formation of long-standing issues in LLMs. We look forward to promoting healthy development in LLM research through responsible use of these research findings. 

%% file: latex/sections/appendix.tex
\appendix

\section{Bloom Cognitive Model}\label{bloom_model}
Bloom Cognitive Model~\cite{krathwohl1973taxonomy} is an educational theoretical framework that outlines six hierarchical levels of cognitive processes, ranging from lower-order to higher-order thinking skills in the learning process as follows:
\begin{itemize}[leftmargin=*]
\item\textbf{Remember}: The capacity to recall, identify, and reproduce information.

\item\textbf{Understand}: The ability to interpret, summarize, and make sense of information.

\item\textbf{Apply}: The skill to use learned knowledge in new contexts.

\item\textbf{Analyze}: The capability to deconstruct information and examine relationships between components.

\item\textbf{Evaluate}: The proficiency in making informed judgments based on specific criteria, involving critical thinking.

\item\textbf{Create}: The ability to synthesize elements into novel patterns or generate original work.
\end{itemize}

Motivated by this framework, we aim to develop the model that goes beyond merely memorizing surface patterns and demonstrates higher-order capabilities. To test whether the model truly understands commonsense knowledge, we create seven variants of each question. It is our view that if the model genuinely understands commonsense knowledge, it should be able to correctly respond to the same knowledge expressed in different reasoning forms. Here's how our seven variants map onto these cognitive levels:
\begin{itemize}[leftmargin=*]
\item\textbf{Understanding} is demonstrated through \textit{Problem Restatement} and \textit{Causal Inference}.

\item\textbf{Application} skills are tested via \textit{Reverse Conversion}, \textit{Scenario Refinement}, and \textit{Negative Transformation}.

\item\textbf{Analysis} capabilities are assessed through \textit{Sentence Ordering}.

\item\textbf{Evaluation} competency is measured by \textit{Critical Testing}.
\end{itemize}

\section{Human Annotation}\label{cn_hellaswag_annotation}

\subsection{Annotator Qualification and Compensation}
We maintained strict control over annotator qualification, data quality, and annotation procedure. Specifically, we recruited 34 professional annotators specializing in NLP tasks totally. All annotators hold at least a bachelor's degree, have passed the College English Test Level 4 of China, and possess extensive annotation experience of NLP tasks. We compensated them at a rate of 23 RMB per hour (significantly higher than the average hourly wage in China), with an average payment of 1.98 RMB per question. We promptly addressed any concerns during the annotation process and allowed sufficient time for each question to prevent unnecessary pressure on annotators.

\subsection{Data Quality and Consistency}
31 out of 34 annotators were involved in data filtering. We enforced the strict annotation guidelines. For Chinese HellaSwag construction, in Stage 1 (initial dataset generation), annotators labeled 12,960 entries in total and filtered down to 12,287 entries. The authors randomly sampled 100 filtered entries and verified them against annotation guidelines, achieving a 98\% compliance rate. In Stage 2 (difficult sample replacement), annotators labeled 5,209 entries in total and filtered down to 2,451 entries. A similar 100-question sample check by authors showed a 96\% compliance rate. For HellaSwag-Pro construction, annotators labeled 24,260 entries, filtering down to 11,200. The authors randomly checked 100 question variants against variant annotation guidelines, achieving a 95\% compliance rate. These measures ensured high quality of our dataset.

\subsection{Human Performance}
To evaluate human performance, we sampled a subset of 400 questions by randomly selecting 25 original questions in both Chinese and English, along with their variants. Three additional crowd workers, who were not involved in the original annotation process, were tested on this subset. We calculated their average accuracy as human performance.

\subsection{Detailed Annotation Guidelines}
We provided rich examples for the annotation tasks to ensure annotators understood the tasks at hand. We maintained close contact with the annotators to clarify any misunderstandings in time. Our annotation tasks were divided into four parts:

\subsubsection{Chinese HellaSwag Annotation for Stage One}
Annotators were given the context, six choices filtered by the model, label, broad type, and detailed type. They scored based on three dimensions: the possibility to select 4 out of 6 choices, and whether they conform to the two category definitions. The annotation requirements for annotators were as follows:
\begin{itemize}[leftmargin=*]
\item\textbf{Possibility to select 4 out of 6}: Using the model's scoring of the 6 choices as a reference, determine if it's possible to select 4 choices, with only one correct answer and the other three being as confusing as possible (i.e., conforming to commonsense but not suitable for the context, or judged by how much modification is needed to make them correct - the less modification needed, the more confusing). Ensure the uniqueness of the answer and avoid controversy. Score 1 if possible, and note the corresponding option numbers, with the first being the correct option and the next three being incorrect options. If not possible, score 0 and select the appropriate reason: A. No correct option or B. Unable to select 3 incorrect options, e.g., more than 4 correct options.

\item\textbf{Broad type}: Score it conforms to the definition, otherwise 0.

\item\textbf{Detailed type}: Score 1 if it conforms to the definition, otherwise 0.
\end{itemize}

The following are the definitions for broad and detailed types. 

\begin{itemize}[leftmargin=*]
\item \textbf{Family}

Household chores: Labor activities to maintain a clean and tidy home environment, including but not limited to cleaning, laundry, and preparing traditional Chinese cuisine.

Personal hygiene: Daily personal cleaning habits such as bathing, brushing teeth, and maintaining good living habits to ensure physical health.

Family entertainment: Leisure activities shared by family members, such as playing family games, pet care, watching TV shows, or reading books together.

Holiday celebrations: Celebrating family members' birthdays, traditional festivals, or special occasions like wedding anniversaries.

Family affairs: Daily life management, emotional communication, and responsibility allocation among family members, including household shopping, financial management, and handling potential disagreements or conflicts.

Family transitions: Changes in family structure or living environment, such as home renovation, moving, marriage, or welcoming a newborn.

Emergency handling: Measures for potential emergencies like fires or natural disasters.

\item \textbf{Education}

School education: Formal education received in school settings, including classroom learning, extracurricular activities, and exam preparation.
Family education: Education provided by parents or other family members, including homework assistance, shared reading, and cultivation of interests and moral qualities.

Online learning: Learning through internet resources, including self-study tools, remote tutoring, and interactive learning platforms.

Community education: Educational activities within the community, such as lectures, interest groups, and practical activities.

Vocational training: Professional training aimed at improving occupational skills, including obtaining professional qualifications and on-the-job continuing education.

Lifelong learning: Continuous learning activities for adults to improve themselves, such as adult education or senior university courses.

International exchange: Consultation for studying abroad, language skill improvement, and other forms of cross-cultural exchange.

\item \textbf{Work}

Work Meetings: Various meetings held in the workplace, including team meetings, departmental reports, and project evaluations.

Project Management: The entire process of managing a project from initiation to completion, including strategy formulation, progress tracking, and problem-solving.

Customer Service: Services provided to meet customer needs, including customer inquiries, complaint handling, sales negotiations, and after-sales support.

Teamwork: Effective collaborative work patterns within a team, including team building, task allocation, conflict resolution, and incentive measures.

Personal Development: The process of individual career growth, covering skill learning, career planning, financial management, and maintaining mental and physical health.

Administrative Management: Daily management activities within a company, including attendance records, performance evaluations, travel expense reimbursements, employee benefits distribution, and company policy communication.

Technological Innovation: Activities driving technological advancement in a company, including new product development, technology application, technical training, and technology exchange.

\item \textbf{Sociality}

Daily Interactions: Everyday social interactions with family, friends, and neighbors.

School Interactions: Communication between students, between teachers and students, and between parents and teachers.

Workplace Interactions: Interactions with colleagues, superiors, or subordinates in the workplace, as well as formal business dinners.

Community Interactions: Participating in community-organized activities or providing volunteer services.

Public Space Interactions: Interactions with others in public spaces such as public transportation, shopping malls, restaurants, and lecture halls.

Online Social Networking: Social activities using online platforms, including social media, online gaming, internet forums, and video live streaming.

Special Occasion Interactions: Interpersonal interactions at weddings, funerals, award ceremonies, and other celebratory events.

\item \textbf{ Shopping}

In-store Shopping: Shopping activities in physical retail stores, such as supermarkets, department stores, and specialty shops.

Online Shopping: Online purchasing behavior through e-commerce platforms, live streaming sales, or social commerce.

Food and Dining Purchases: Buying food products, including dining out, ordering takeout, and home cooking.

Service Purchases: Buying various service products, such as travel services, beauty and fitness, and educational training.

Overseas Shopping: Purchasing foreign goods through cross-border e-commerce or personal shopping agents.

Special Occasion Shopping: Shopping in specific situations, such as promotional events, group buying, auctions, and second-hand transactions.

Returns and After-sales Service: Consumer behavior in seeking refunds, exchanges, and after-sales service when issues arise with products.

\item \textbf{Transportation}

Public Transportation: Using public transit systems, such as buses and subways.

Private Transportation: Using private vehicles, bicycles, etc., for travel.

Long-distance Travel: Travel methods covering longer distances, such as trains, planes, or long-distance buses.

Emergency Travel: Choosing emergency transportation in response to sudden situations, such as travel during severe weather conditions.

Tourist Transportation: Using sightseeing vehicles, boats, or cable cars for tourism purposes.

International Travel: Visa applications, international flight bookings, and entry procedures required for traveling abroad.

Special Occasion Transportation: Transportation services provided for specific situations, such as wedding cars or conference shuttles.

\item \textbf{Health}

Preventive Healthcare: Measures taken to prevent diseases, including health check-ups, vaccinations, and health education.

Outpatient Care: Receiving non-hospitalized treatment at hospitals or clinics, including appointment scheduling, initial diagnosis, follow-up visits, and specialist consultations.

Inpatient Treatment: Hospital admission for treatment, including admission procedures, ward life, surgery arrangements, and discharge preparation.

Rehabilitation Care: Treatment during the recovery period, including rehabilitation training, long-term care, and psychological counseling.

Medication Management: Guidance on medication use and storage methods.

Health Insurance: Purchasing medical insurance products, claim procedures, and health consultation services.

Epidemic Prevention and Control: Measures such as epidemic monitoring, isolation observation, and health code management.

\item \textbf{Leisure}

Outdoor Activities: Recreational activities in natural settings, such as hiking, picnicking, and gardening.

Cultural Experiences: Engaging in cultural activities like visiting museums, watching theatrical performances, or attending film screenings.

Travel Experiences: Domestic or international tourism activities.

Sporting Events: Watching or participating in sports competitions, including esports.

Artistic Pursuits: Engaging in artistic activities such as painting, calligraphy, playing musical instruments, or creating handicrafts.

Digital Entertainment: Leisure activities using digital devices, such as online gaming, live streaming interactions, or virtual reality experiences.

Recreational Fitness: Maintaining physical and mental health through activities like gym workouts or practicing yoga and meditation.

\end{itemize}

\subsubsection{Chinese HellaSwag Annotation for Stage Two}
In order to increase the number of difficult samples, the annotators were given a context and four replaced options regenerated by models to judge whether the label of the question was correct and whether it had a unique correct option. If both are true, the replaced options were retained.

\subsubsection{Hellaswag-Pro Annotation} \label{appe:hellaswag_pro_annotation}
Annotators are provided with the original context, original choices, original label, transformed context, transformed choices, transformed label, and perturbation type for annotation according to different variant definitions. The variant definitions are as follows:

\begin{itemize}[leftmargin=*]
\item\textbf{Problem restatement}:
Restate the original context and the original label corresponding to the original choices in a different way, ensuring the semantics remain unchanged. Other options of the original choices should remain unchanged without restatement. Pay special attention to ensuring that the connection between the context and the choice corresponding to the label is smooth.

\item\textbf{Reverse conversion}:
Combine the original choices corresponding to the original label with the original context into a complete passage. Then, make the last sentence of this passage the context, and transform the remaining sentences into the correct choice. A slight modification is allowed for smoothness. Also, generate five other incorrect options that do not fit the context, modeled on the format and length of the correct option. Place the correct option in the first position and label it as 0. To ensure the context is complete, append "Which is the possible context for this action?" This conversion process aims to infer the potential background through the results. The generated incorrect options should not include supernatural elements and should have a similar word count to the correct option.

\item\textbf{Causal inference}:
Combine the original choices corresponding to the original label with the original context to form a complete passage and turn it into the context. Then, generate the reason for such choices that contain commonsense as the correct option in the choices. The correct choice should be as concise as possible while generating five other evidently incorrect options modeled on the format and length of the correct choice. Put the correct choice in the first position and label it as 0. To ensure the context is complete, append "Which is the possible reason for this action?" This conversion process aims to infer the potential reason through the context and options.

\item\textbf{Negative transformation}:
Modify the original context to end with a negation word as the context, retaining one most unreasonable option and the original choice corresponding to the original label. Then, generate two other reasonable options as choices. Generated options should be similar in length and format to the original options. Place this most unreasonable option as the first element in the choices and label the index of this option in choices as 0. This conversion process aims to transform the original task into a negation prediction, containing one unreasonable option and three other reasonable options.

\item\textbf{Scenario refinement}:
First, select a relatively reasonable option from the incorrect options in the original choices, then modify the original context as the context to allow the selection of this option as the correct choice. The value of choices is equal to the original choices. The label value corresponds to the value of the selected incorrect option. This conversion process aims to refine the context, thereby altering the correct choice.

\item\textbf{Sentence ordering}:

1) Sentence ordering - Short:
First, combine the original choices corresponding to the original label with the original context into a complete sentence. Then, predict the development of subsequent events, continuing to write a few more sentences to form a paragraph. Pay attention to the sequence and completeness of continued sentences, ensuring the uniqueness of the answer. Then, disorder each sentence of this passage and number them. The correct option is the original order of the paragraph, and three other incorrect options are generated based on the correct option by disordering the numbers. Place the correct option in the first position and label it as 0. To ensure the context is complete, append "The correct order is." This conversion process aims to infer the correct order of sentences.

2) Sentence ordering - Long:
Combine the original choices corresponding to the original label with the original context into a complete passage. Then, disorder each sentence of this passage and number them. The correct option is the original order of the paragraph, and three other incorrect options are generated by disordering the numbers. Place the correct option in the first position and label it as 0. To ensure the context is complete, append "The correct order is." This conversion process aims to infer the correct order of sentences.

\item\textbf{Critical testing}:
Modify the original context so that none of the options can be chosen as the context, then add an option of 'None of the above four options are appropriate' to the original choices as choices. The label value corresponds to the index of 'None of the above four options are appropriate'. Note that the modified context should still present a question, ideally with an ending word identical to the original context. This conversion process aims to test the model's critical thinking.
\end{itemize}

\subsubsection{Hellaswag-Pro Human Evaluation}
Annotators were provided with the context and choices from the Hellaswag-Pro and made selections. We then compared the selections made by annotators with the labels to calculate accuracy.

\section{Prompt Strategy}
The prompting strategies we designed, including Direct, CN-CoT, EN-CoT, CN-XLT and EN-XLT, are as shown Figure \ref{prompt1}, \ref{prompt2} and \ref{prompt3}.

\begin{CJK*}{UTF8}{gbsn}
\begin{table*}[t]
 \centering
 \scalebox{0.9}{
\begin{tabularx}{\textwidth}{lXX}
  \toprule
  
  \textbf{Problem:} & Question: 周老师负责学校的英语培训工作，常为学生提供实用的语言学习建议。他建议多读原版英文报纸提高阅读水平。可能的原因是 \newline
Choices: \newline 
A.主张减少课外阅读时间增加听力训练。 \newline
B.阅读原版材料能够增强语感并扩大词汇量。 \newline
C.觉得看漫画书更能激发学生的兴趣。\newline
D.相信通过电影比读书更容易学好语言。 & Question: The mother instructs them on how to brush their teeth while laughing. The boy helps his younger sister brush his teeth. She gets them some water to gargle in their mouths. Which could be the possible reason for this action?
Choices:  \newline
A. She wants to ensure they rinse out any remaining toothpaste and clean their mouth thoroughly. \newline
B. She is conducting an experiment to see if water tastes different after brushing. \newline
C. She is using the water as a distraction so they won't notice her leaving the room. \newline
D. She is training them to be synchronized swimmers by coordinating their gargles. \\
  \cdashline{1-3}[3pt/2.5pt]\noalign{\vskip 0.5ex}
Prompt& Chinese Dataset  & English Dataset   \\ 
 \cdashline{1-3}[3pt/2.5pt]\noalign{\vskip 0.5ex}
Direct& Answer: B & Answer: A\\ \cdashline{1-3}[3pt/2.5pt]\noalign{\vskip 0.5ex}
CN-   CoT & 答案：让我们一步步思考。首先，我们需要分析周老师作为英语培训教师的建议 - 多读原版英文报纸。这建议的目的是为了提高学生的阅读水平。查看选项：A提到减少阅读增加听力，与题目表述矛盾；C提到漫画书和D提到电影都偏离了原版英文报纸这一重点；只有B选项提到阅读原版材料可以增强语感并扩大词汇量，这与读原版英文报纸的效果是一致的。原版英文报纸使用地道的英语表达，包含丰富的词汇和表达方式，长期阅读确实能帮助学习者提高语感，扩大词汇量，提升整体语言水平。答案是B。  & 答案：让我们一步步思考。这是一个关于刷牙后漱口的场景，母亲在指导孩子们刷牙，并让他们用水漱口。从日常生活经验来看，刷牙后漱口的主要目的是清除口腔中残留的牙膏和清洁口腔。漱口这个动作不太可能是为了做实验测试水的味道（选项B），也不是为了分散注意力（选项C），更不可能是在训练同步游泳（选项D）。选项A提到让孩子们漱口是为了冲洗掉剩余的牙膏并彻底清洁口腔，这符合我们日常刷牙后漱口的目的，也是最合理的解释。因此答案是A。  \\  \cdashline{1-3}[3pt/2.5pt]\noalign{\vskip 0.5ex}
EN-CoT& Answer: Let's think step by step. First, we know that Teacher   Zhou is responsible for English training at school and often provides   practical language learning advice to students. Second, he specifically   recommends reading original English newspapers to improve reading skills.   Third, reading original English materials has several benefits: it exposes   students to authentic language usage, helps them understand how native   speakers express ideas, enhances language intuition, and naturally expands   their vocabulary through context. Fourth, looking at the options, option B   directly aligns with these benefits by mentioning both improved language   sense and vocabulary expansion. The other options either suggest different   learning methods (like movies or comics) or propose reducing reading time,   which contradicts the teacher's recommendation. Since Teacher Zhou   specifically advocates for reading original English newspapers, option B is   the most logical explanation for his recommendation. The answer is B.& Answer: Let's think step by step. Gargling with water after brushing teeth is   a common and important part of proper dental hygiene. The mother in this   scenario is teaching her children good oral care habits. When we brush our   teeth, toothpaste helps clean them, but some toothpaste residue remains in   our mouth. This residue, if not rinsed out, can leave a lingering taste and   might contain fluoride that shouldn't be swallowed in large amounts. By   giving the children water to gargle, the mother is helping them complete   their tooth-brushing routine correctly. The gargling action helps remove any   loose debris, leftover toothpaste, and bacteria from the mouth, leading to   better oral hygiene. Looking at the other options, they are either illogical   or unrelated to dental care - testing water taste, using it as a distraction,   or swimming practice are not reasonable explanations for gargling after   brushing teeth. The most sensible and practical reason for the mother's   action is to ensure her children properly rinse their mouths after brushing.   The answer is A. \\

  \bottomrule
   
\end{tabularx}
 }

 \caption{Examples of Direct, CN-CoT and EN-CoT Prompt.}
 \label{prompt1}
\end{table*}
\end{CJK*}
\begin{CJK*}{UTF8}{gbsn}
\begin{table*}[t]
 \centering
 \scalebox{0.9}{
\begin{tabularx}{\textwidth}{lXX}
  \toprule
  
  \textbf{Problem:} & Question: 周老师负责学校的英语培训工作，常为学生提供实用的语言学习建议。他建议多读原版英文报纸提高阅读水平。可能的原因是 \newline
Choices: \newline 
A.主张减少课外阅读时间增加听力训练。 \newline
B.阅读原版材料能够增强语感并扩大词汇量。 \newline
C.觉得看漫画书更能激发学生的兴趣。\newline
D.相信通过电影比读书更容易学好语言。 & Question: The mother instructs them on how to brush their teeth while laughing. The boy helps his younger sister brush his teeth. She gets them some water to gargle in their mouths. Which could be the possible reason for this action?
Choices:  \newline
A. She wants to ensure they rinse out any remaining toothpaste and clean their mouth thoroughly. \newline
B. She is conducting an experiment to see if water tastes different after brushing. \newline
C. She is using the water as a distraction so they won't notice her leaving the room. \newline
D. She is training them to be synchronized swimmers by coordinating their gargles. \\
  \cdashline{1-3}[3pt/2.5pt]\noalign{\vskip 0.5ex}
Prompt& Chinese Dataset  & English Dataset   \\ 
 \cdashline{1-3}[3pt/2.5pt]\noalign{\vskip 0.5ex}
CN-   XLT & - & 我希望你扮演一个英文常识推理专家。\newline 你应该用中文复述问题和选项。\newline 你应该一步一步地回答问题，选出正确的答案。\newline 问题：母亲一边笑一边指导他们如何刷牙。男孩帮助他的小妹妹刷牙。她给他们一些水让他们漱口。对此行为，可能的原因是什么？\newline 选项：\newline A.   她想确保他们漱掉任何剩余的牙膏并清洁嘴巴。\newline B.   她正在进行实验，以查看刷牙后水的味道是否不同。\newline C.   她用水作为分散注意力的方法，以便他们不会注意到她离开房间。\newline D. 她通过协调漱口来训练他们成为同步游泳运动员。\newline 分步解答：\newline 1.   母亲给他们水漱口，通常是为了卫生目的。\newline 2.   漱口可以帮助去除残留的牙膏，确保口腔清洁。\newline 3.   选项A是一个常见且合理的原因，符合情境。\newline 4. 选项B、C和D不符合通常情况下的刷牙步骤或目的。\newline 所以答案是A。   \\

  \bottomrule
   
\end{tabularx}
 }

 \caption{Examples of CN-XLT Prompt.}
 \label{prompt2}
\end{table*}
\end{CJK*}

\begin{CJK*}{UTF8}{gbsn}
\begin{table*}[t]
 \centering
 \scalebox{0.85}{
\begin{tabularx}{\textwidth}{lXX}
  \toprule
  
  \textbf{Problem:} & Question: 周老师负责学校的英语培训工作，常为学生提供实用的语言学习建议。他建议多读原版英文报纸提高阅读水平。可能的原因是 \newline
Choices: \newline 
A.主张减少课外阅读时间增加听力训练。 \newline
B.阅读原版材料能够增强语感并扩大词汇量。 \newline
C.觉得看漫画书更能激发学生的兴趣。\newline
D.相信通过电影比读书更容易学好语言。 & Question: The mother instructs them on how to brush their teeth while laughing. The boy helps his younger sister brush his teeth. She gets them some water to gargle in their mouths. Which could be the possible reason for this action?
Choices:  \newline
A. She wants to ensure they rinse out any remaining toothpaste and clean their mouth thoroughly. \newline
B. She is conducting an experiment to see if water tastes different after brushing. \newline
C. She is using the water as a distraction so they won't notice her leaving the room. \newline
D. She is training them to be synchronized swimmers by coordinating their gargles. \\
  \cdashline{1-3}[3pt/2.5pt]\noalign{\vskip 0.5ex}
Prompt& Chinese Dataset  & English Dataset   \\ 
 \cdashline{1-3}[3pt/2.5pt]\noalign{\vskip 0.5ex}

EN-XLT& I want you to act as a commonsense reasoning expert for   Chinese.\newline You should retell the question and choices in English.\newline You should   do the answer step by step to choose the right answer.\newline You should tell me   the answer in this format 'So the answer is'.\newline Question: Teacher Zhou is in   charge of the English training at school and often provides students with   practical language learning advice. He suggests reading original English   newspapers to improve reading skills. The possible reason is:\newline Choices:\newline A. Advocates   reducing extracurricular reading time to increase listening practice.\newline B.   Reading original materials can enhance language sense and expand   vocabulary.\newline C. Believes that reading comic books can better stimulate   students' interest.\newline D. Believes it is easier to learn a language well   through movies than through reading.\newline \newline Step-by-step answer:\newline 1. The   statement suggests reading original English newspapers as a way to improve   reading skills.\newline 2. Option B suggests that reading original materials can enhance   language sense and expand vocabulary, which aligns with the goal of improving   reading skills.\newline 3. Option A talks about reducing reading time, which   contradicts the suggestion to read newspapers.\newline 4. Option C talks about comic   books, which is not related to reading original newspapers.\newline 5. Option D   suggests movies are more effective, which again, does not align with the   focus on reading newspapers.\newline So the answer is B. & - \\

  \bottomrule
   
\end{tabularx}
 }

 \caption{Examples of EN-XLT Prompt.}
 \label{prompt3}
\end{table*}
\end{CJK*}

\section{Case Study}
Figure \ref{case} shows an example of the Chinese hellaswag generation process, from which we can see that our wrong options are becoming more and more challenging.

\begin{CJK*}{UTF8}{gbsn}
\begin{table*}[]
 \centering
 \scalebox{0.79}{
\begin{tabular}{l|l}
\hline
Step& Gererated or Filtered Content\\ \hline
\begin{tabular}[c]{@{}l@{}}\textbf{Step   1}\\Over-generate\\context with broad\\type and detailed\\type with Qwen- Max\end{tabular}& \begin{tabular}[c]{@{}l@{}}Context：丽丽报名参加了日本京都的一趟文化之旅，深度体验了传统艺伎表演。她\\ Broad Type: 休闲娱乐\\ Detailed Type: 旅游体验\\ Context:   Lili signed up for a cultural tour in Kyoto, Japan, and experienced a  traditional \\ geisha performance. She\\ Broad Type: Leisure\\ Detailed Type: Travel Experience\end{tabular}\\ \hline
\begin{tabular}[c]{@{}l@{}}\textbf{Step   2}\\Over-generate\\candidate choices\\with Qwen-Max\end{tabular} & \begin{tabular}[c]{@{}l@{}}Choices:\\1. 深深地被茶道的精致仪式所吸引。(Score:10)\\2. 抱怨为什么所有的寺庙都没有钟声。(Score:3)\\3. 抱怨旅馆里的榻榻米太过柔软。(Score:3)\\4. 要求在京都找到寿司制作的机器人。(Score:2)\\5. 质疑为什么艺伎穿的是拖鞋而不是高跟鞋。(Score:5)\\6. 坚持要在神社祈祷时使用英文。(Score:2)\\7. 对亲自动手参与和服穿着过程表现出浓厚的兴趣。(Score:9)\\8. 询问为什么没人听说过她喜欢的动漫角色。(Score:4)\\9. 希望能在旅途中看到更多现代化的建筑。(Score:3)\\10. 感到遗憾没能在旅途中参与一场本地的相扑比赛。(Score:4)\\ Choices:\\1. was deeply attracted by the exquisite ritual of the tea ceremony.   (Score:10)\\2. complained why there were no bells in all temples. (Score:3)\\3. complained that the tatami in the hotel was too soft. (Score:3)\\4. requested to find a sushi-making robot in Kyoto. (Score:2)\\5. questioned why geishas wore slippers instead of high heels.   (Score:5)\\6. insisted on using English when praying at the shrine. (Score:2)\\7. showed great interest in participating in the process of wearing a   kimono. (Score:9)\\8. asked why no one has heard of her favorite anime character.   (Score:4)\\9. hoped to see more modern buildings during the trip. (Score:3)\\10. regreted not being able to attend a local sumo match during the trip.   (Score:4)\end{tabular} \\ \hline
\begin{tabular}[c]{@{}l@{}}\textbf{Step   3-1}\\Find filtered choices\\with Qwen-Max\end{tabular}& \begin{tabular}[c]{@{}l@{}}Choices:\\1. 深深地被茶道的精致仪式所吸引。(Score:10)\\5. 质疑为什么艺伎穿的是拖鞋而不是高跟鞋。(Score:5)\\8. 询问为什么没人听说过她喜欢的动漫角色。(Score:4)\\10. 感到遗憾没能在旅途中参与一场本地的相扑比赛。(Score:4)\\2. 抱怨为什么所有的寺庙都没有钟声。(Score:3)\\9. 希望能在旅途中看到更多现代化的建筑。(Score:3) \\ Choices:\\1. was deeply attracted by the exquisite ritual of the tea ceremony.   (Score:10)\\5. questioned why geishas wore slippers instead of high heels. (Score:5)\\8. asked why no one has heard of her favorite anime character.   (Score:4)\\10. regreted not being able to attend a local sumo match during the trip.   (Score:4)\\2. complained why there were no bells in all temples. (Score:3)\\9. hoped to see more modern buildings during the trip. (Score:3)\end{tabular} \\ \hline
\begin{tabular}[c]{@{}l@{}}\textbf{Step   3-2}\\Find filtered choices\\with human annotators\end{tabular}  & \begin{tabular}[c]{@{}l@{}}Choices:\\1. 深深地被茶道的精致仪式所吸引。(Score:10)\\5. 质疑为什么艺伎穿的是拖鞋而不是高跟鞋。(Score:5)\\8. 询问为什么没人听说过她喜欢的动漫角色。(Score:4)\\9. 希望能在旅途中看到更多现代化的建筑。(Score:3)\\ Choices:\\1. was deeply attracted by the exquisite ritual of the tea ceremony.   (Score:10)\\5. questioned why geishas wore slippers instead of high heels.   (Score:5)\\8. asked why no one has heard of her favorite anime character.   (Score:4)\\9. hoped to see more modern buildings during the trip. (Score:3)\end{tabular}  \\ \hline
\begin{tabular}[c]{@{}l@{}}\textbf{Step   4}\\Replace easily-\\identifiable false\\choices with\\adversarial ones\\through human-in-\\the-loop alternating\\adversarial filtering\end{tabular} & \begin{tabular}[c]{@{}l@{}}Choices:\\1.深深地被茶道的精致仪式所吸引。\\2.学习了传统的日式剑道和弓道技巧。\\3.欣赏了京都著名的樱花季和红叶景观。\\4.品尝了正宗的关西风味章鱼烧和大阪烧。\\ Choices:\\1. was deeply attracted by the exquisite ritual of the tea ceremony. \\2. learned traditional Japanese kendo and archery techniques.\\3. enjoyed Kyoto's famous cherry blossom season and red leaves.\\4. tasted authentic Kansai-style takoyaki and okonomiyaki.\end{tabular}   \\ \hline
\end{tabular}
}
 \caption{An example of Chinese HellaSwag Gereration. Step 3-1 filters the top 5 wrong options with scores below 9 to prevent multiple correct options, and Step 3-2 select the most confusing wrong options by human annotators.}
 \label{case}
\end{table*}
\end{CJK*}

\section{Chinese HellaSwag Evaluation}\label{cn_eval}
We also evaluate the overall results of Chinese Hellaswag using both open-source and closed-source models, analyzing them from the perspectives of broad categories and length categories. As shown in Table \ref{cn_type}, within all categories of Chinese Hellaswag, \textit{Traffic} is the most challenging, with an average accuracy of only 58.56\%, while the \textit{Education} category is the easiest, achieving an average accuracy of 77.64\%. Additionally, as the context length increases, the difficulty of the problems generally decreases, with average accuracy of long types at 72\%, medium types at 70\%, and short types at 64\% as shown in Table \ref{cn_length}. Overall, the closed-source models outperform open-source models. Among the closed-source models, Claude-3.5 performs the best, reaching an accuracy of 94\%, whereas among open-source models, Qwen2.5-72B shows the highest performance, achieving 71\%.

\begin{table*}[h]
\centering
 \scalebox{0.9}{

\begin{tabular}{l|llllllll|l}
\hline
Model                 & Education & Health & Famliy & Leisure & Shopping & Sociality & Traffic & Work & AVG  \\ \hline
Baichuan2-7B-Base     & 0.76      & 0.71   & 0.66   & 0.69    & 0.62     & 0.68   & 0.55    & 0.70 & 0.67 \\
Baichuan2-13B-Base    & 0.78      & 0.70   & 0.68   & 0.70    & 0.64     & 0.69   & 0.57    & 0.71 & 0.68 \\
Meta-Llama-3-8B       & 0.74      & 0.59   & 0.55   & 0.57    & 0.54     & 0.56   & 0.46    & 0.61 & 0.58 \\
Meta-Llama-3-70B      & 0.76      & 0.65   & 0.63   & 0.66    & 0.63     & 0.67   & 0.54    & 0.65 & 0.65 \\
Mistral-7B-v0.1       & 0.70      & 0.59   & 0.52   & 0.56    & 0.57     & 0.57   & 0.50    & 0.61 & 0.58 \\
Qwen2.5-0.5B          & 0.72      & 0.66   & 0.53   & 0.60    & 0.53     & 0.58   & 0.47    & 0.66 & 0.59 \\
Qwen2.5-1.5B          & 0.75      & 0.66   & 0.60   & 0.62    & 0.59     & 0.64   & 0.51    & 0.67 & 0.63 \\
Qwen2.5-3B            & 0.75      & 0.67   & 0.63   & 0.66    & 0.61     & 0.66   & 0.55    & 0.68 & 0.65 \\
Qwen2.5-7B            & 0.76      & 0.68   & 0.66   & 0.68    & 0.63     & 0.69   & 0.58    & 0.70 & 0.67 \\
Qwen2.5-14B           & 0.78      & 0.68   & 0.68   & 0.69    & 0.65     & 0.69   & 0.58    & 0.71 & 0.68 \\
Qwen2.5-32B           & 0.77      & 0.69   & 0.68   & 0.69    & 0.66     & 0.69   & 0.58    & 0.69 & 0.68 \\
Qwen2.5-72B           & 0.78      & 0.70   & 0.70   & 0.72    & 0.69     & 0.73   & 0.60    & 0.73 & 0.71 \\
Yi-1.5-6B             & 0.78      & 0.69   & 0.66   & 0.68    & 0.63     & 0.69   & 0.56    & 0.72 & 0.68 \\
Yi-1.5-9B             & 0.78      & 0.70   & 0.67   & 0.70    & 0.64     & 0.69   & 0.57    & 0.72 & 0.68 \\
deepseek-llm-7b-base  & 0.79      & 0.70   & 0.67   & 0.69    & 0.64     & 0.69   & 0.57    & 0.73 & 0.68 \\
deepseek-llm-67b-base & 0.80      & 0.72   & 0.70   & 0.72    & 0.67     & 0.70   & 0.58    & 0.74 & 0.70 \\
gemma-2-2b            & 0.73      & 0.62   & 0.57   & 0.60    & 0.60     & 0.60   & 0.50    & 0.66 & 0.61 \\
gemma-2-9b            & 0.78      & 0.68   & 0.64   & 0.67    & 0.65     & 0.69   & 0.55    & 0.74 & 0.67 \\
gemma-2-27b           & 0.72      & 0.66   & 0.64   & 0.62    & 0.62     & 0.58   & 0.50    & 0.67 & 0.63 \\
internlm2\_5-1\_8b    & 0.73      & 0.64   & 0.58   & 0.64    & 0.54     & 0.60   & 0.49    & 0.65 & 0.61 \\
internlm2\_5-7b       & 0.76      & 0.68   & 0.67   & 0.70    & 0.63     & 0.67   & 0.60    & 0.69 & 0.67 \\
internlm2\_5-20b      & 0.76      & 0.67   & 0.68   & 0.70    & 0.64     & 0.68   & 0.59    & 0.69 & 0.68 \\
GPT-4o                & 0.91      & 0.92   & 0.88   & 0.92    & 0.90     & 0.90   & 0.86    & 0.91 & 0.90 \\
Claude-3-5            & 0.94      & 0.96   & 0.94   & 0.94    & 0.95     & 0.95   & 0.91    & 0.96 & 0.94 \\
Gemini-1.5-pro        & 0.88      & 0.91   & 0.88   & 0.90    & 0.90     & 0.91   & 0.85    & 0.91 & 0.89 \\
Qwen-Max              & 0.91      & 0.95   & 0.91   & 0.92    & 0.93     & 0.94   & 0.88    & 0.94 & 0.92 \\ \hline
AVG                   & 0.78      & 0.71   & 0.68   & 0.70    & 0.67     & 0.70   & 0.60    & 0.73 & 0.69 \\ \hline
\end{tabular}

}
  \caption{Model Performance on Chinese HellaSwag based on broad category under Direct Prompt.}
 \label{cn_type}
\end{table*}

\begin{table*}[]
\centering
 \scalebox{1.1}{

\begin{tabular}{l|lll|l}
\hline
\multicolumn{1}{l|}{Model} & Long & Medium & Short & AVG  \\ \hline
Baichuan2-7B-Base           & 0.70 & 0.70   & 0.62  & 0.67 \\
Baichuan2-13B-Base          & 0.72 & 0.71   & 0.62  & 0.68 \\
Meta-Llama-3-8B             & 0.64 & 0.59   & 0.51  & 0.58 \\
Meta-Llama-3-70B            & 0.70 & 0.67   & 0.58  & 0.65 \\
Mistral-7B-v0.1             & 0.63 & 0.58   & 0.52  & 0.58 \\
Qwen2.5-0.5B                & 0.63 & 0.61   & 0.54  & 0.59 \\
Qwen2.5-1.5B                & 0.67 & 0.64   & 0.58  & 0.63 \\
Qwen2.5-3B                  & 0.68 & 0.67   & 0.60  & 0.65 \\
Qwen2.5-7B                  & 0.71 & 0.69   & 0.62  & 0.67 \\
Qwen2.5-14B                 & 0.72 & 0.70   & 0.63  & 0.68 \\
Qwen2.5-32B                 & 0.72 & 0.69   & 0.62  & 0.68 \\
Qwen2.5-72B                 & 0.74 & 0.73   & 0.65  & 0.71 \\
Yi-1.5-6B                   & 0.72 & 0.70   & 0.62  & 0.68 \\
Yi-1.5-9B                   & 0.73 & 0.70   & 0.62  & 0.68 \\
deepseek-llm-7b-base        & 0.73 & 0.71   & 0.61  & 0.68 \\
deepseek-llm-67b-base       & 0.76 & 0.72   & 0.62  & 0.70 \\
gemma-2-2b                  & 0.65 & 0.61   & 0.56  & 0.61 \\
gemma-2-9b                  & 0.72 & 0.69   & 0.61  & 0.67 \\
gemma-2-27b                 & 0.68 & 0.64   & 0.56  & 0.63 \\
internlm2\_5-1\_8b          & 0.65 & 0.63   & 0.55  & 0.61 \\
internlm2\_5-7b             & 0.72 & 0.69   & 0.61  & 0.67 \\
internlm2\_5-20b            & 0.73 & 0.69   & 0.61  & 0.68 \\
GPT-4o                      & 0.87 & 0.90   & 0.93  & 0.90 \\
Claude-3-5                  & 0.92 & 0.95   & 0.97  & 0.94 \\
Gemini-1.5-pro              & 0.86 & 0.91   & 0.92  & 0.89 \\
Qwen-Max                    & 0.89 & 0.93   & 0.95  & 0.92 \\ \hline
AVG                         & 0.73 & 0.71   & 0.65  & 0.69 \\ \hline
\end{tabular}

}
  \caption{Model Performance on Chinese HellaSwag based on length category under Direct Prompt.}
 \label{cn_length}
\end{table*}

\section{Experiment Detailed Result}\label{detailed}
Figures \ref{qwen1} to \ref{internlm} show the detailed results of the open-source models on the 9 prompt strategies and 7 variants we design.

\begin{figure}[h]
\setlength{\abovecaptionskip}{0.1cm}
\setlength{\belowcaptionskip}{0cm}
\includegraphics[width=0.85\linewidth,scale=0.50]{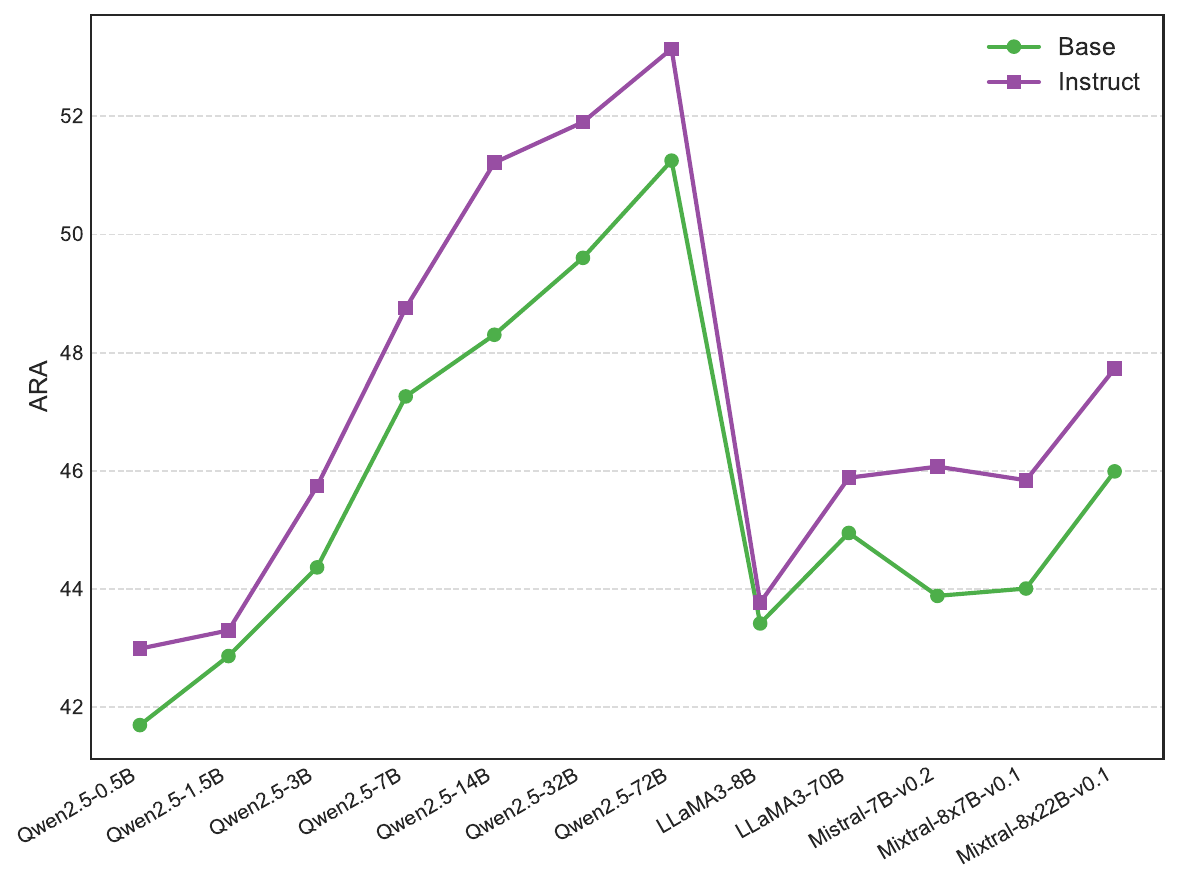}
\caption{Comparasion between base and their instruct models.}
\label{instruct}
\vspace{-15pt}
\end{figure}

In addition, we also use direct prompt to evaluate the performance of the Qwen, LLaMA, and Mixtral series of intruct models on our robustness task, as shown in Table \ref{instruct-tab} and Figure \ref{instruct}. We can observe that the intruct models have improved performance in robustness task compared to the base models.

\begin{table*}[]
\centering
 \scalebox{0.9}{
\begin{tabular}{l|llllllll}
\hline
\textbf{Model\_\_Prompt\_Language}& CI   & CT   & NT   & PR   & RC   & SR   &SO  & \textbf{ARA} \\ \hline
Qwen2.5-0.5B\_\_direct\_cn& 0.66 & 0.36 & 0.06 & 0.64 & 0.50 & 0.36 & 0.58 & 0.45   \\
Qwen2.5-1.5B\_\_direct\_cn& 0.70 & 0.35 & 0.07 & 0.65 & 0.55 & 0.38 & 0.54 & 0.46   \\
Qwen2.5-3B\_\_direct\_cn& 0.66 & 0.37 & 0.06 & 0.66 & 0.57 & 0.42 & 0.68 & 0.49   \\
Qwen2.5-0.5B\_\_direct\_cn& 0.52 & 0.34 & 0.07 & 0.67 & 0.36 & 0.35 & 0.37 & 0.38   \\
Qwen2.5-1.5B\_\_direct\_cn& 0.56 & 0.38 & 0.06 & 0.75 & 0.32 & 0.35 & 0.36 & 0.40   \\
Qwen2.5-3B\_\_direct\_cn& 0.59 & 0.40 & 0.05 & 0.78 & 0.29 & 0.34 & 0.35 & 0.40   \\ \hline
   
Qwen2.5-0.5B\_\_few\_shot\_en\_cot\_cn  & 0.75 & 0.48 & 0.14 & 0.62 & 0.43 & 0.38 & 0.23 & 0.43   \\
Qwen2.5-0.5B\_\_few\_shot\_en\_cot\_en  & 0.80 & 0.29 & 0.42 & 0.47 & 0.41 & 0.32 & 0.74 & 0.49   \\
Qwen2.5-0.5B\_\_few\_shot\_en\_xlt\_cn  & 0.40 & 0.11 & 0.12 & 0.35 & 0.34 & 0.18 & 0.09 & 0.23   \\
Qwen2.5-0.5B\_\_few\_shot\_cn\_cot\_cn  & 0.73 & 0.51 & 0.17 & 0.62 & 0.36 & 0.29 & 0.30 & 0.42   \\
Qwen2.5-0.5B\_\_few\_shot\_cn\_cot\_en  & 0.81 & 0.37 & 0.64 & 0.35 & 0.43 & 0.26 & 0.89 & 0.54   \\
Qwen2.5-0.5B\_\_few\_shot\_cn\_xlt\_en  & 0.73 & 0.28 & 0.15 & 0.29 & 0.36 & 0.18 & 0.88 & 0.41   \\ \hline
   
Qwen2.5-1.5B\_\_few\_shot\_en\_cot\_cn  & 0.91 & 0.40 & 0.75 & 0.82 & 0.79 & 0.43 & 0.40 & 0.64   \\
Qwen2.5-1.5B\_\_few\_shot\_en\_cot\_en  & 0.82 & 0.26 & 0.50 & 0.66 & 0.42 & 0.47 & 0.82 & 0.56   \\
Qwen2.5-1.5B\_\_few\_shot\_en\_xlt\_cn  & 0.33 & 0.18 & 0.41 & 0.50 & 0.45 & 0.22 & 0.23 & 0.33   \\
Qwen2.5-1.5B\_\_few\_shot\_cn\_cot\_cn  & 0.89 & 0.48 & 0.82 & 0.84 & 0.74 & 0.42 & 0.62 & 0.68   \\
Qwen2.5-1.5B\_\_few\_shot\_cn\_cot\_en  & 0.85 & 0.41 & 0.37 & 0.63 & 0.39 & 0.41 & 0.70 & 0.54   \\
Qwen2.5-1.5B\_\_few\_shot\_cn\_xlt\_en  & 0.45 & 0.17 & 0.15 & 0.54 & 0.26 & 0.23 & 0.64 & 0.35   \\ \hline
   
Qwen2.5-3B\_\_few\_shot\_en\_cot\_cn& 0.94 & 0.50 & 0.83 & 0.89 & 0.86 & 0.47 & 0.70 & 0.74   \\
Qwen2.5-3B\_\_few\_shot\_en\_cot\_en& 0.89 & 0.39 & 0.52 & 0.72 & 0.48 & 0.49 & 0.68 & 0.59   \\
Qwen2.5-3B\_\_few\_shot\_en\_xlt\_cn& 0.41 & 0.27 & 0.45 & 0.51 & 0.51 & 0.32 & 0.10 & 0.37   \\
Qwen2.5-3B\_\_few\_shot\_cn\_cot\_cn& 0.92 & 0.49 & 0.90 & 0.89 & 0.81 & 0.46 & 0.80 & 0.75   \\
Qwen2.5-3B\_\_few\_shot\_cn\_cot\_en& 0.89 & 0.35 & 0.57 & 0.70 & 0.48 & 0.44 & 0.73 & 0.59   \\
Qwen2.5-3B\_\_few\_shot\_cn\_xlt\_en& 0.72 & 0.19 & 0.64 & 0.58 & 0.40 & 0.28 & 0.51 & 0.47   \\ \hline
   
Qwen2.5-0.5B\_\_zero\_shot\_en\_cot\_cn & 0.54 & 0.18 & 0.07 & 0.54 & 0.34 & 0.23 & 0.18 & 0.30   \\
Qwen2.5-0.5B\_\_zero\_shot\_en\_cot\_en & 0.53 & 0.17 & 0.29 & 0.36 & 0.30 & 0.29 & 0.27 & 0.32   \\
Qwen2.5-0.5B\_\_zero\_shot\_en\_xlt\_cn & 0.14 & 0.11 & 0.02 & 0.09 & 0.07 & 0.08 & 0.00 & 0.07   \\
Qwen2.5-0.5B\_\_zero\_shot\_cn\_cot\_cn & 0.59 & 0.33 & 0.10 & 0.49 & 0.26 & 0.00 & 0.06 & 0.26   \\
Qwen2.5-0.5B\_\_zero\_shot\_cn\_cot\_en & 0.42 & 0.18 & 0.28 & 0.27 & 0.24 & 0.20 & 0.43 & 0.29   \\
Qwen2.5-0.5B\_\_zero\_shot\_cn\_xlt\_en & 0.01 & 0.01 & 0.02 & 0.10 & 0.01 & 0.06 & 0.01 & 0.03   \\ \hline
   
Qwen2.5-1.5B\_\_zero\_shot\_en\_cot\_cn & 0.95 & 0.57 & 0.23 & 0.86 & 0.74 & 0.49 & 0.39 & 0.61   \\
Qwen2.5-1.5B\_\_zero\_shot\_en\_cot\_en & 0.69 & 0.32 & 0.54 & 0.47 & 0.48 & 0.40 & 0.41 & 0.47   \\
Qwen2.5-1.5B\_\_zero\_shot\_en\_xlt\_cn & 0.03 & 0.03 & 0.01 & 0.01 & 0.00 & 0.04 & 0.01 & 0.02   \\
Qwen2.5-1.5B\_\_zero\_shot\_cn\_cot\_cn & 0.70 & 0.45 & 0.35 & 0.72 & 0.55 & 0.00 & 0.38 & 0.45   \\
Qwen2.5-1.5B\_\_zero\_shot\_cn\_cot\_en & 0.48 & 0.26 & 0.05 & 0.53 & 0.28 & 0.36 & 0.40 & 0.34   \\
Qwen2.5-1.5B\_\_zero\_shot\_cn\_xlt\_en & 0.02 & 0.02 & 0.03 & 0.04 & 0.00 & 0.03 & 0.00 & 0.02   \\ \hline
   
Qwen2.5-3B\_\_zero\_shot\_en\_cot\_cn   & 0.92 & 0.44 & 0.50 & 0.88 & 0.78 & 0.46 & 0.73 & 0.67   \\
Qwen2.5-3B\_\_zero\_shot\_en\_cot\_en   & 0.81 & 0.27 & 0.33 & 0.66 & 0.43 & 0.45 & 0.56 & 0.50   \\
Qwen2.5-3B\_\_zero\_shot\_en\_xlt\_cn   & 0.52 & 0.51 & 0.48 & 0.54 & 0.50 & 0.43 & 0.53 & 0.50   \\
Qwen2.5-3B\_\_zero\_shot\_cn\_cot\_cn   & 0.75 & 0.43 & 0.28 & 0.72 & 0.55 & 0.00 & 0.58 & 0.47   \\
Qwen2.5-3B\_\_zero\_shot\_cn\_cot\_en   & 0.51 & 0.20 & 0.07 & 0.53 & 0.37 & 0.32 & 0.39 & 0.34   \\
Qwen2.5-3B\_\_zero\_shot\_cn\_xlt\_en   & 0.82 & 0.25 & 0.03 & 0.73 & 0.37 & 0.47 & 0.13 & 0.40   \\ \hline
\end{tabular}
}
  \caption{Performance of Qwen Series (0.5B-3B). And NT, CT, CI, SO, SR, RC, PR are the abbreviations for the variant names of Negation Transformation, Critical Testing, Causal Inference, Sentence Ordering, Scenario Refinement, Reverse Conversion and Problem Restatement.}
 \label{qwen1}
\end{table*}

\begin{table*}[]
\centering
 \scalebox{0.8}{
\begin{tabular}{l|llllllll}
\hline
\textbf{Model\_\_Prompt\_Language}& CI   & CT   & NT   & PR   & RC   & SR   &SO  & \textbf{ARA} \\ \hline
Qwen2.5-7B\_\_direct\_cn& 0.66 & 0.34 & 0.07 & 0.67 & 0.62 & 0.41 & 0.77 & 0.51   \\
Qwen2.5-14B\_\_direct\_cn   & 0.67 & 0.35 & 0.07 & 0.69 & 0.63 & 0.43 & 0.75 & 0.51   \\
Qwen2.5-32B\_\_direct\_cn   & 0.68 & 0.36 & 0.06 & 0.68 & 0.63 & 0.43 & 0.87 & 0.53   \\
Qwen2.5-72B\_\_direct\_cn   & 0.67 & 0.39 & 0.08 & 0.69 & 0.65 & 0.44 & 0.92 & 0.55   \\
Qwen2.5-7B\_\_direct\_en& 0.64 & 0.41 & 0.05 & 0.82 & 0.27 & 0.33 & 0.56 & 0.44   \\
Qwen2.5-14B\_\_direct\_en   & 0.66 & 0.42 & 0.05 & 0.83 & 0.28 & 0.35 & 0.58 & 0.45   \\
Qwen2.5-32B\_\_direct\_en   & 0.65 & 0.42 & 0.05 & 0.83 & 0.29 & 0.34 & 0.65 & 0.46   \\
Qwen2.5-72B\_\_direct\_en   & 0.67 & 0.43 & 0.05 & 0.86 & 0.30 & 0.34 & 0.71 & 0.48   \\ \hline

Qwen2.5-7B\_\_few\_shot\_en\_cot\_cn  & 0.95 & 0.57 & 0.92 & 0.90 & 0.89 & 0.53 & 0.60 & 0.77   \\
Qwen2.5-7B\_\_few\_shot\_en\_cot\_en  & 0.88 & 0.50 & 0.78 & 0.84 & 0.54 & 0.54 & 0.81 & 0.70   \\
Qwen2.5-7B\_\_few\_shot\_en\_xlt\_cn  & 0.51 & 0.27 & 0.50 & 0.48 & 0.53 & 0.41 & 0.44 & 0.45   \\
Qwen2.5-7B\_\_few\_shot\_cn\_cot\_cn  & 0.96 & 0.55 & 0.91 & 0.92 & 0.85 & 0.59 & 0.93 & 0.81   \\
Qwen2.5-7B\_\_few\_shot\_cn\_cot\_en  & 0.93 & 0.53 & 0.79 & 0.75 & 0.56 & 0.54 & 0.81 & 0.70   \\
Qwen2.5-7B\_\_few\_shot\_cn\_xlt\_en  & 0.76 & 0.30 & 0.77 & 0.69 & 0.43 & 0.27 & 0.65 & 0.55   \\ \hline
 
Qwen2.5-14B\_\_few\_shot\_en\_cot\_cn & 0.97 & 0.58 & 0.93 & 0.93 & 0.88 & 0.66 & 0.94 & 0.84   \\
Qwen2.5-14B\_\_few\_shot\_en\_cot\_en & 0.93 & 0.50 & 0.75 & 0.88 & 0.57 & 0.55 & 0.81 & 0.71   \\
Qwen2.5-14B\_\_few\_shot\_en\_xlt\_cn & 0.63 & 0.40 & 0.65 & 0.69 & 0.58 & 0.49 & 0.69 & 0.59   \\
Qwen2.5-14B\_\_few\_shot\_cn\_cot\_cn & 0.97 & 0.60 & 0.94 & 0.91 & 0.87 & 0.66 & 0.92 & 0.84   \\
Qwen2.5-14B\_\_few\_shot\_cn\_cot\_en & 0.93 & 0.56 & 0.77 & 0.83 & 0.56 & 0.55 & 0.82 & 0.72   \\
Qwen2.5-14B\_\_few\_shot\_cn\_xlt\_en & 0.82 & 0.38 & 0.74 & 0.71 & 0.42 & 0.36 & 0.57 & 0.57   \\ \hline
 
Qwen2.5-32B\_\_few\_shot\_en\_cot\_cn & 0.98 & 0.63 & 0.91 & 0.94 & 0.92 & 0.71 & 0.95 & 0.86   \\
Qwen2.5-32B\_\_few\_shot\_en\_cot\_en & 0.93 & 0.59 & 0.84 & 0.88 & 0.64 & 0.58 & 0.81 & 0.75   \\
Qwen2.5-32B\_\_few\_shot\_en\_xlt\_cn & 0.68 & 0.46 & 0.75 & 0.66 & 0.60 & 0.49 & 0.80 & 0.64   \\
Qwen2.5-32B\_\_few\_shot\_cn\_cot\_cn & 0.98 & 0.61 & 0.94 & 0.93 & 0.90 & 0.68 & 0.95 & 0.85   \\
Qwen2.5-32B\_\_few\_shot\_cn\_cot\_en & 0.94 & 0.66 & 0.84 & 0.90 & 0.62 & 0.59 & 0.83 & 0.77   \\
Qwen2.5-32B\_\_few\_shot\_cn\_xlt\_en & 0.82 & 0.46 & 0.82 & 0.82 & 0.51 & 0.42 & 0.59 & 0.63   \\ \hline
 
Qwen2.5-72B\_\_few\_shot\_en\_cot\_cn & 0.98 & 0.66 & 0.91 & 0.94 & 0.92 & 0.73 & 0.92 & 0.87   \\
Qwen2.5-72B\_\_few\_shot\_en\_cot\_en & 0.91 & 0.59 & 0.80 & 0.92 & 0.67 & 0.58 & 0.81 & 0.75   \\
Qwen2.5-72B\_\_few\_shot\_en\_xlt\_cn & 0.50 & 0.30 & 0.68 & 0.62 & 0.55 & 0.54 & 0.76 & 0.56   \\
Qwen2.5-72B\_\_few\_shot\_cn\_cot\_cn & 0.97 & 0.62 & 0.95 & 0.93 & 0.91 & 0.74 & 0.92 & 0.86   \\
Qwen2.5-72B\_\_few\_shot\_cn\_cot\_en & 0.94 & 0.62 & 0.80 & 0.90 & 0.69 & 0.59 & 0.83 & 0.77   \\
Qwen2.5-72B\_\_few\_shot\_cn\_xlt\_en & 0.59 & 0.44 & 0.81 & 0.70 & 0.53 & 0.33 & 0.57 & 0.57   \\ \hline

Qwen2.5-7B\_\_zero\_shot\_en\_cot\_cn & 0.82 & 0.56 & 0.73 & 0.85 & 0.70 & 0.54 & 0.78 & 0.71   \\
Qwen2.5-7B\_\_zero\_shot\_en\_cot\_en & 0.82 & 0.40 & 0.53 & 0.70 & 0.38 & 0.53 & 0.59 & 0.57   \\
Qwen2.5-7B\_\_zero\_shot\_en\_xlt\_cn & 0.83 & 0.55 & 0.62 & 0.80 & 0.73 & 0.53 & 0.83 & 0.70   \\
Qwen2.5-7B\_\_zero\_shot\_cn\_cot\_cn & 0.74 & 0.42 & 0.30 & 0.76 & 0.65 & 0.00 & 0.50 & 0.48   \\
Qwen2.5-7B\_\_zero\_shot\_cn\_cot\_en & 0.70 & 0.27 & 0.09 & 0.57 & 0.33 & 0.37 & 0.54 & 0.41   \\
Qwen2.5-7B\_\_zero\_shot\_cn\_xlt\_en & 0.02 & 0.00 & 0.00 & 0.00 & 0.01 & 0.00 & 0.01 & 0.01   \\ \hline

Qwen2.5-14B\_\_zero\_shot\_en\_cot\_cn& 0.62 & 0.46 & 0.81 & 0.78 & 0.66 & 0.54 & 0.86 & 0.68   \\
Qwen2.5-14B\_\_zero\_shot\_en\_cot\_en& 0.88 & 0.44 & 0.41 & 0.70 & 0.46 & 0.53 & 0.72 & 0.59   \\
Qwen2.5-14B\_\_zero\_shot\_en\_xlt\_cn& 0.93 & 0.63 & 0.85 & 0.93 & 0.79 & 0.70 & 0.92 & 0.82   \\
Qwen2.5-14B\_\_zero\_shot\_cn\_cot\_cn& 0.79 & 0.56 & 0.74 & 0.81 & 0.73 & 0.51 & 0.82 & 0.71   \\
Qwen2.5-14B\_\_zero\_shot\_cn\_cot\_en& 0.72 & 0.37 & 0.24 & 0.64 & 0.44 & 0.38 & 0.59 & 0.48   \\
Qwen2.5-14B\_\_zero\_shot\_cn\_xlt\_en& 0.01 & 0.00 & 0.17 & 0.09 & 0.22 & 0.02 & 0.01 & 0.07   \\ \hline
 
Qwen2.5-32B\_\_zero\_shot\_en\_cot\_cn& 0.80 & 0.43 & 0.83 & 0.81 & 0.68 & 0.57 & 0.86 & 0.71   \\
Qwen2.5-32B\_\_zero\_shot\_en\_cot\_en& 0.86 & 0.52 & 0.56 & 0.82 & 0.53 & 0.54 & 0.75 & 0.65   \\
Qwen2.5-32B\_\_zero\_shot\_en\_xlt\_cn& 0.78 & 0.58 & 0.83 & 0.75 & 0.49 & 0.52 & 0.49 & 0.63   \\
Qwen2.5-32B\_\_zero\_shot\_cn\_cot\_cn& 0.87 & 0.60 & 0.81 & 0.87 & 0.76 & 0.00 & 0.91 & 0.69   \\
Qwen2.5-32B\_\_zero\_shot\_cn\_cot\_en& 0.82 & 0.47 & 0.30 & 0.79 & 0.51 & 0.48 & 0.61 & 0.57   \\
Qwen2.5-32B\_\_zero\_shot\_cn\_xlt\_en& 0.64 & 0.45 & 0.36 & 0.62 & 0.55 & 0.27 & 0.62 & 0.50   \\ \hline
 
Qwen2.5-72B\_\_zero\_shot\_en\_cot\_cn& 0.84 & 0.48 & 0.82 & 0.84 & 0.72 & 0.60 & 0.73 & 0.72   \\
Qwen2.5-72B\_\_zero\_shot\_en\_cot\_en& 0.78 & 0.50 & 0.44 & 0.79 & 0.49 & 0.51 & 0.75 & 0.61   \\
Qwen2.5-72B\_\_zero\_shot\_en\_xlt\_cn& 0.06 & 0.06 & 0.18 & 0.21 & 0.19 & 0.05 & 0.17 & 0.13   \\
Qwen2.5-72B\_\_zero\_shot\_cn\_cot\_cn& 0.79 & 0.59 & 0.73 & 0.86 & 0.73 & 0.56 & 0.82 & 0.73   \\
Qwen2.5-72B\_\_zero\_shot\_cn\_cot\_en& 0.70 & 0.31 & 0.15 & 0.76 & 0.55 & 0.51 & 0.48 & 0.49   \\
Qwen2.5-72B\_\_zero\_shot\_cn\_xlt\_en& 0.03 & 0.01 & 0.14 & 0.20 & 0.39 & 0.07 & 0.04 & 0.12   \\ \hline
\end{tabular}
}
 \caption{Performance of Qwen Series (7B-72B). And NT, CT, CI, SO, SR, RC, PR are the abbreviations for the variant names of Negation Transformation, Critical Testing, Causal Inference, Sentence Ordering, Scenario Refinement, Reverse Conversion and Problem Restatement.}
 \label{qwen2}
\end{table*}

\begin{table*}[]
\centering
 \scalebox{0.9}{
\begin{tabular}{l|llllllll}
\hline
\textbf{Model\_\_Prompt\_Language}  & CI   & CT   & NT   & PR   & RC   & SR   &SO  & \textbf{ARA} \\ \hline
deepseek-llm-7b-base\_\_direct\_cn & 0.61 & 0.35 & 0.07 & 0.67 & 0.59 & 0.43 & 0.43 & 0.48\\
deepseek-llm-67b-base\_\_direct\_cn& 0.65 & 0.38 & 0.08 & 0.71 & 0.63 & 0.46 & 0.55 & 0.49\\
deepseek-llm-7b-base\_\_direct\_en & 0.53 & 0.41 & 0.05 & 0.81 & 0.33 & 0.34 & 0.37 & 0.40\\
deepseek-llm-67b-base\_\_direct\_en& 0.57 & 0.42 & 0.05 & 0.85 & 0.25 & 0.34 & 0.37 & 0.41\\ \hline

deepseek-llm-7b-base\_\_few\_shot\_en\_cot\_cn   & 0.85 & 0.49 & 0.35 & 0.81 & 0.62 & 0.43 & 0.40 & 0.56\\
deepseek-llm-7b-base\_\_few\_shot\_en\_cot\_en   & 0.85 & 0.20 & 0.45 & 0.53 & 0.26 & 0.40 & 0.90 & 0.51\\
deepseek-llm-7b-base\_\_few\_shot\_en\_xlt\_cn   & 0.27 & 0.06 & 0.10 & 0.33 & 0.42 & 0.28 & 0.28 & 0.25\\
deepseek-llm-7b-base\_\_few\_shot\_cn\_cot\_cn   & 0.88 & 0.55 & 0.66 & 0.74 & 0.63 & 0.49 & 0.56 & 0.64\\
deepseek-llm-7b-base\_\_few\_shot\_cn\_cot\_en   & 0.82 & 0.27 & 0.62 & 0.44 & 0.38 & 0.34 & 0.82 & 0.52\\
deepseek-llm-7b-base\_\_few\_shot\_cn\_xlt\_en   & 0.34 & 0.16 & 0.51 & 0.46 & 0.45 & 0.29 & 0.81 & 0.43\\ \hline

deepseek-llm-67b-base\_\_few\_shot\_en\_cot\_cn  & 0.96 & 0.51 & 0.89 & 0.91 & 0.84 & 0.68 & 0.81 & 0.80\\
deepseek-llm-67b-base\_\_few\_shot\_en\_cot\_en  & 0.92 & 0.39 & 0.88 & 0.85 & 0.53 & 0.49 & 0.88 & 0.71\\
deepseek-llm-67b-base\_\_few\_shot\_en\_xlt\_cn  & 0.42 & 0.10 & 0.70 & 0.63 & 0.51 & 0.50 & 0.36 & 0.46\\
deepseek-llm-67b-base\_\_few\_shot\_cn\_cot\_cn  & 0.97 & 0.61 & 0.91 & 0.89 & 0.82 & 0.71 & 0.90 & 0.83\\
deepseek-llm-67b-base\_\_few\_shot\_cn\_cot\_en  & 0.92 & 0.43 & 0.84 & 0.79 & 0.45 & 0.44 & 0.90 & 0.68\\
deepseek-llm-67b-base\_\_few\_shot\_cn\_xlt\_en  & 0.52 & 0.24 & 0.18 & 0.74 & 0.46 & 0.34 & 0.65 & 0.45\\ \hline

deepseek-llm-7b-base\_\_zero\_shot\_en\_cot\_cn  & 0.18 & 0.03 & 0.03 & 0.15 & 0.05 & 0.05 & 0.12 & 0.09\\
deepseek-llm-7b-base\_\_zero\_shot\_en\_cot\_en  & 0.11 & 0.14 & 0.01 & 0.16 & 0.09 & 0.19 & 0.01 & 0.10\\
deepseek-llm-7b-base\_\_zero\_shot\_en\_xlt\_cn  & 0.03 & 0.01 & 0.01 & 0.08 & 0.02 & 0.02 & 0.01 & 0.02\\
deepseek-llm-7b-base\_\_zero\_shot\_cn\_cot\_cn  & 0.40 & 0.18 & 0.07 & 0.23 & 0.18 & 0.00 & 0.09 & 0.16\\
deepseek-llm-7b-base\_\_zero\_shot\_cn\_cot\_en  & 0.27 & 0.12 & 0.07 & 0.17 & 0.13 & 0.11 & 0.00 & 0.12\\
deepseek-llm-7b-base\_\_zero\_shot\_cn\_xlt\_en  & 0.01 & 0.00 & 0.00 & 0.02 & 0.01 & 0.01 & 0.00 & 0.01\\ \hline

deepseek-llm-67b-base\_\_zero\_shot\_en\_cot\_cn & 0.03 & 0.17 & 0.01 & 0.14 & 0.12 & 0.12 & 0.40 & 0.14\\
deepseek-llm-67b-base\_\_zero\_shot\_en\_cot\_en & 0.08 & 0.12 & 0.01 & 0.11 & 0.02 & 0.08 & 0.33 & 0.11\\
deepseek-llm-67b-base\_\_zero\_shot\_en\_xlt\_cn & 0.64 & 0.34 & 0.02 & 0.29 & 0.29 & 0.30 & 0.03 & 0.27\\
deepseek-llm-67b-base\_\_zero\_shot\_cn\_cot\_cn & 0.36 & 0.14 & 0.03 & 0.36 & 0.18 & 0.20 & 0.34 & 0.23\\
deepseek-llm-67b-base\_\_zero\_shot\_cn\_cot\_en & 0.20 & 0.08 & 0.01 & 0.08 & 0.02 & 0.07 & 0.16 & 0.09\\
deepseek-llm-67b-base\_\_zero\_shot\_cn\_xlt\_en & 0.49 & 0.06 & 0.00 & 0.29 & 0.14 & 0.11 & 0.07 & 0.17\\ \hline
\end{tabular}
}
 \caption{Performance of DeepSeek Series. And NT, CT, CI, SO, SR, RC, PR are the abbreviations for the variant names of Negation Transformation, Critical Testing, Causal Inference, Sentence Ordering, Scenario Refinement, Reverse Conversion and Problem Restatement.}
 \label{deepseek}
\end{table*}

\begin{table*}[]
\centering
 \scalebox{0.9}{

\begin{tabular}{l|llllllll}
\hline
\textbf{Model\_\_Prompt\_Language} & CI   & CT   & NT   & PR   & RC   & SR   &SO  & \textbf{ARA} \\ \hline
Yi-6B\_\_direct\_cn & 0.68 & 0.37 & 0.07 & 0.67 & 0.60 & 0.42 & 0.67 & 0.50\\
Yi-9B\_\_direct\_cn & 0.70 & 0.37 & 0.08 & 0.66 & 0.64 & 0.44 & 0.62 & 0.50\\
Yi-34B\_\_direct\_cn& 0.69 & 0.38 & 0.09 & 0.67 & 0.67 & 0.44 & 0.72 & 0.52\\
Yi-6B\_\_direct\_en & 0.58 & 0.39 & 0.05 & 0.77 & 0.27 & 0.34 & 0.36 & 0.39\\
Yi-9B\_\_direct\_en & 0.56 & 0.41 & 0.06 & 0.80 & 0.26 & 0.35 & 0.35 & 0.40\\
Yi-34B\_\_direct\_en& 0.62 & 0.41 & 0.05 & 0.81 & 0.27 & 0.34 & 0.36 & 0.41\\ \hline
   
Yi-1.5-6B\_\_few\_shot\_en\_cot\_cn   & 0.91 & 0.60 & 0.78 & 0.84 & 0.82 & 0.56 & 0.68 & 0.74\\
Yi-1.5-6B\_\_few\_shot\_en\_cot\_en   & 0.89 & 0.36 & 0.45 & 0.68 & 0.33 & 0.51 & 0.65 & 0.55\\
Yi-1.5-6B\_\_few\_shot\_en\_xlt\_cn   & 0.41 & 0.10 & 0.47 & 0.48 & 0.44 & 0.35 & 0.23 & 0.35\\
Yi-1.5-6B\_\_few\_shot\_cn\_cot\_cn   & 0.94 & 0.58 & 0.85 & 0.86 & 0.76 & 0.57 & 0.72 & 0.75\\
Yi-1.5-6B\_\_few\_shot\_cn\_cot\_en   & 0.83 & 0.46 & 0.60 & 0.57 & 0.33 & 0.40 & 0.86 & 0.58\\
Yi-1.5-6B\_\_few\_shot\_cn\_xlt\_en   & 0.72 & 0.23 & 0.15 & 0.55 & 0.41 & 0.27 & 0.41 & 0.39\\ \hline
   
Yi-1.5-9B\_\_few\_shot\_en\_cot\_cn   & 0.97 & 0.60 & 0.85 & 0.88 & 0.88 & 0.64 & 0.78 & 0.80\\
Yi-1.5-9B\_\_few\_shot\_en\_cot\_en   & 0.93 & 0.45 & 0.78 & 0.77 & 0.55 & 0.50 & 0.77 & 0.68\\
Yi-1.5-9B\_\_few\_shot\_en\_xlt\_cn   & 0.48 & 0.24 & 0.68 & 0.50 & 0.45 & 0.39 & 0.41 & 0.45\\
Yi-1.5-9B\_\_few\_shot\_cn\_cot\_cn   & 0.96 & 0.61 & 0.90 & 0.89 & 0.86 & 0.65 & 0.87 & 0.82\\
Yi-1.5-9B\_\_few\_shot\_cn\_cot\_en   & 0.92 & 0.50 & 0.67 & 0.70 & 0.52 & 0.47 & 0.77 & 0.65\\
Yi-1.5-9B\_\_few\_shot\_cn\_xlt\_en   & 0.84 & 0.25 & 0.53 & 0.69 & 0.46 & 0.29 & 0.45 & 0.50\\ \hline
   
Yi-1.5-34B\_\_few\_shot\_en\_cot\_cn  & 0.96 & 0.54 & 0.91 & 0.91 & 0.88 & 0.70 & 0.93 & 0.83\\
Yi-1.5-34B\_\_few\_shot\_en\_cot\_en  & 0.92 & 0.59 & 0.79 & 0.92 & 0.60 & 0.54 & 0.82 & 0.74\\
Yi-1.5-34B\_\_few\_shot\_en\_xlt\_cn  & 0.27 & 0.12 & 0.58 & 0.36 & 0.44 & 0.40 & 0.43 & 0.37\\
Yi-1.5-34B\_\_few\_shot\_cn\_cot\_cn  & 0.96 & 0.57 & 0.92 & 0.89 & 0.86 & 0.68 & 0.93 & 0.83\\
Yi-1.5-34B\_\_few\_shot\_cn\_cot\_en  & 0.93 & 0.57 & 0.79 & 0.87 & 0.57 & 0.52 & 0.78 & 0.72\\
Yi-1.5-34B\_\_few\_shot\_cn\_xlt\_en  & 0.62 & 0.33 & 0.71 & 0.74 & 0.49 & 0.30 & 0.57 & 0.54\\ \hline
   
Yi-1.5-6B\_\_zero\_shot\_en\_cot\_cn  & 0.77 & 0.52 & 0.12 & 0.67 & 0.64 & 0.34 & 0.48 & 0.51\\
Yi-1.5-6B\_\_zero\_shot\_en\_cot\_en  & 0.83 & 0.43 & 0.50 & 0.59 & 0.55 & 0.54 & 0.21 & 0.52\\
Yi-1.5-6B\_\_zero\_shot\_en\_xlt\_cn  & 0.20 & 0.17 & 0.02 & 0.19 & 0.15 & 0.06 & 0.30 & 0.16\\
Yi-1.5-6B\_\_zero\_shot\_cn\_cot\_cn  & 0.69 & 0.27 & 0.05 & 0.63 & 0.57 & 0.00 & 0.24 & 0.35\\
Yi-1.5-6B\_\_zero\_shot\_cn\_cot\_en  & 0.58 & 0.26 & 0.04 & 0.48 & 0.25 & 0.37 & 0.17 & 0.31\\
Yi-1.5-6B\_\_zero\_shot\_cn\_xlt\_en  & 0.05 & 0.05 & 0.02 & 0.02 & 0.01 & 0.03 & 0.07 & 0.04\\ \hline
   
Yi-1.5-9B\_\_zero\_shot\_en\_cot\_cn  & 0.60 & 0.47 & 0.14 & 0.67 & 0.48 & 0.39 & 0.70 & 0.49\\
Yi-1.5-9B\_\_zero\_shot\_en\_cot\_en  & 0.87 & 0.38 & 0.26 & 0.68 & 0.47 & 0.51 & 0.48 & 0.52\\
Yi-1.5-9B\_\_zero\_shot\_en\_xlt\_cn  & 0.94 & 0.51 & 0.27 & 0.85 & 0.76 & 0.56 & 0.70 & 0.65\\
Yi-1.5-9B\_\_zero\_shot\_cn\_cot\_cn  & 0.79 & 0.36 & 0.12 & 0.72 & 0.70 & 0.50 & 0.14 & 0.47\\
Yi-1.5-9B\_\_zero\_shot\_cn\_cot\_en  & 0.61 & 0.18 & 0.10 & 0.50 & 0.35 & 0.36 & 0.29 & 0.34\\
Yi-1.5-9B\_\_zero\_shot\_cn\_xlt\_en  & 0.91 & 0.64 & 0.03 & 0.84 & 0.45 & 0.58 & 0.59 & 0.58\\ \hline
   
Yi-1.5-34B\_\_zero\_shot\_en\_cot\_cn & 0.28 & 0.37 & 0.18 & 0.24 & 0.28 & 0.17 & 0.58 & 0.30\\
Yi-1.5-34B\_\_zero\_shot\_en\_cot\_en & 0.85 & 0.48 & 0.35 & 0.75 & 0.50 & 0.47 & 0.66 & 0.58\\
Yi-1.5-34B\_\_zero\_shot\_en\_xlt\_cn & 0.09 & 0.03 & 0.08 & 0.08 & 0.07 & 0.05 & 0.09 & 0.07\\
Yi-1.5-34B\_\_zero\_shot\_cn\_cot\_cn & 0.82 & 0.36 & 0.29 & 0.72 & 0.69 & 0.00 & 0.77 & 0.52\\
Yi-1.5-34B\_\_zero\_shot\_cn\_cot\_en & 0.77 & 0.31 & 0.05 & 0.64 & 0.50 & 0.43 & 0.64 & 0.48\\
Yi-1.5-34B\_\_zero\_shot\_cn\_xlt\_en & 0.78 & 0.25 & 0.02 & 0.26 & 0.43 & 0.16 & 0.34 & 0.32\\ \hline
\end{tabular}

}
 \caption{Performance of Yi Series. And NT, CT, CI, SO, SR, RC, PR are the abbreviations for the variant names of Negation Transformation, Critical Testing, Causal Inference, Sentence Ordering, Scenario Refinement, Reverse Conversion and Problem Restatement.}
 \label{baichuan}
\end{table*}

\begin{table*}[]
\centering
 \scalebox{0.9}{
\begin{tabular}{l|llllllll}
\hline
\textbf{Model\_\_Prompt\_Language}  & CI   & CT   & NT   & PR   & RC   & SR   &SO  & \textbf{ARA} \\ \hline
Meta-Llama-3-8B\_\_direct\_cn   & 0.57 & 0.37 & 0.09 & 0.63 & 0.51 & 0.46 & 0.63 & 0.47\\
Meta-Llama-3-70B\_\_direct\_cn  & 0.63 & 0.40 & 0.08 & 0.67 & 0.60 & 0.46 & 0.57 & 0.49\\
Meta-Llama-3-8B\_\_direct\_en   & 0.56 & 0.41 & 0.05 & 0.82 & 0.27 & 0.35 & 0.36 & 0.40\\
Meta-Llama-3-70B\_\_direct\_cn  & 0.57 & 0.43 & 0.04 & 0.86 & 0.29 & 0.34 & 0.36 & 0.41\\
 \hline
Meta-Llama-3-8B\_\_few\_shot\_en\_cot\_cn   & 0.90 & 0.45 & 0.68 & 0.81 & 0.67 & 0.57 & 0.58 & 0.66\\
Meta-Llama-3-8B\_\_few\_shot\_en\_cot\_en   & 0.79 & 0.39 & 0.79 & 0.64 & 0.51 & 0.45 & 0.94 & 0.64\\
Meta-Llama-3-8B\_\_few\_shot\_en\_xlt\_cn   & 0.36 & 0.21 & 0.39 & 0.48 & 0.54 & 0.39 & 0.23 & 0.37\\
Meta-Llama-3-8B\_\_few\_shot\_cn\_cot\_cn   & 0.87 & 0.45 & 0.67 & 0.73 & 0.54 & 0.60 & 0.63 & 0.64\\
Meta-Llama-3-8B\_\_few\_shot\_cn\_cot\_en   & 0.76 & 0.46 & 0.56 & 0.59 & 0.39 & 0.38 & 0.82 & 0.56\\
Meta-Llama-3-8B\_\_few\_shot\_cn\_xlt\_en   & 0.10 & 0.10 & 0.06 & 0.27 & 0.13 & 0.05 & 0.32 & 0.15\\ \hline

Meta-Llama-3-70B\_\_few\_shot\_en\_cot\_cn  & 0.97 & 0.58 & 0.87 & 0.91 & 0.89 & 0.70 & 0.91 & 0.83\\
Meta-Llama-3-70B\_\_few\_shot\_en\_cot\_en  & 0.92 & 0.42 & 0.77 & 0.84 & 0.57 & 0.50 & 0.88 & 0.70\\
Meta-Llama-3-70B\_\_few\_shot\_en\_xlt\_cn  & 0.46 & 0.22 & 0.65 & 0.62 & 0.56 & 0.52 & 0.52 & 0.51\\
Meta-Llama-3-70B\_\_few\_shot\_cn\_cot\_cn  & 0.97 & 0.36 & 0.87 & 0.84 & 0.69 & 0.63 & 0.92 & 0.75\\
Meta-Llama-3-70B\_\_few\_shot\_cn\_cot\_en  & 0.93 & 0.45 & 0.73 & 0.86 & 0.56 & 0.44 & 0.85 & 0.69\\
Meta-Llama-3-70B\_\_few\_shot\_cn\_xlt\_en  & 0.12 & 0.09 & 0.31 & 0.50 & 0.25 & 0.14 & 0.28 & 0.24\\ \hline

Meta-Llama-3-8B\_\_zero\_shot\_en\_cot\_cn  & 0.59 & 0.24 & 0.07 & 0.40 & 0.32 & 0.29 & 0.40 & 0.33\\
Meta-Llama-3-8B\_\_zero\_shot\_en\_cot\_en  & 0.52 & 0.18 & 0.11 & 0.38 & 0.19 & 0.33 & 0.38 & 0.30\\
Meta-Llama-3-8B\_\_zero\_shot\_en\_xlt\_cn  & 0.39 & 0.17 & 0.01 & 0.42 & 0.16 & 0.19 & 0.16 & 0.22\\
Meta-Llama-3-8B\_\_zero\_shot\_cn\_cot\_cn  & 0.53 & 0.21 & 0.09 & 0.40 & 0.34 & 0.50 & 0.12 & 0.31\\
Meta-Llama-3-8B\_\_zero\_shot\_cn\_cot\_en  & 0.50 & 0.16 & 0.04 & 0.30 & 0.22 & 0.24 & 0.12 & 0.23\\
Meta-Llama-3-8B\_\_zero\_shot\_cn\_xlt\_en  & 0.43 & 0.08 & 0.02 & 0.22 & 0.23 & 0.15 & 0.03 & 0.17\\ \hline

Meta-Llama-3-70B\_\_zero\_shot\_en\_cot\_cn & 0.78 & 0.43 & 0.06 & 0.64 & 0.63 & 0.37 & 0.56 & 0.50\\
Meta-Llama-3-70B\_\_zero\_shot\_en\_cot\_en & 0.78 & 0.35 & 0.04 & 0.64 & 0.45 & 0.41 & 0.57 & 0.46\\
Meta-Llama-3-70B\_\_zero\_shot\_en\_xlt\_cn & 0.79 & 0.55 & 0.05 & 0.77 & 0.49 & 0.36 & 0.22 & 0.46\\
Meta-Llama-3-70B\_\_zero\_shot\_cn\_cot\_cn & 0.63 & 0.35 & 0.13 & 0.56 & 0.48 & 0.34 & 0.63 & 0.44\\
Meta-Llama-3-70B\_\_zero\_shot\_cn\_cot\_en & 0.58 & 0.30 & 0.02 & 0.50 & 0.32 & 0.31 & 0.38 & 0.34\\
Meta-Llama-3-70B\_\_zero\_shot\_cn\_xlt\_en & 0.90 & 0.39 & 0.01 & 0.86 & 0.52 & 0.50 & 0.62 & 0.54\\ \hline
\end{tabular}
}
 \caption{Performance of LLaMA Series. And NT, CT, CI, SO, SR, RC, PR are the abbreviations for the variant names of Negation Transformation, Critical Testing, Causal Inference, Sentence Ordering, Scenario Refinement, Reverse Conversion and Problem Restatement.}
 \label{llama}
\end{table*}

\begin{table*}[]
\centering
 \scalebox{0.9}{

\begin{tabular}{l|llllllll}
\hline
\textbf{Model\_\_Prompt\_Language}& CI   & CT   & NT   & PR   & RC   & SR   &SO  & \textbf{ARA} \\ \hline
Mistral-7B-v0.1\_\_direct\_cn& 0.58 & 0.37 & 0.10 & 0.64 & 0.50 & 0.44 & 0.62 & 0.46 \\
Mixtral-8x7B-v0.1\_\_direct\_cn  & 0.59 & 0.36 & 0.09 & 0.65 & 0.54 & 0.46 & 0.58 & 0.47 \\
Mixtral-8x22B-v0.1\_\_direct\_cn & 0.66 & 0.39 & 0.07 & 0.69 & 0.57 & 0.43 & 0.74 & 0.51 \\
Mistral-7B-v0.1\_\_direct\_en& 0.57 & 0.41 & 0.05 & 0.82 & 0.31 & 0.34 & 0.40 & 0.42 \\
Mixtral-8x7B-v0.1\_\_direct\_en  & 0.56 & 0.42 & 0.06 & 0.85 & 0.31 & 0.33 & 0.36 & 0.41 \\
Mixtral-8x22B-v0.1\_\_direct\_en & 0.57 & 0.42 & 0.04 & 0.85 & 0.31 & 0.34 & 0.36 & 0.41 \\ \hline
  
Mistral-7B-v0.1\_\_few\_shot\_en\_cot\_cn & 0.90 & 0.51 & 0.62 & 0.83 & 0.71 & 0.56 & 0.57 & 0.67 \\
Mistral-7B-v0.1\_\_few\_shot\_en\_cot\_en & 0.86 & 0.34 & 0.84 & 0.63 & 0.53 & 0.47 & 0.90 & 0.65 \\
Mistral-7B-v0.1\_\_few\_shot\_en\_xlt\_cn & 0.13 & 0.05 & 0.17 & 0.20 & 0.30 & 0.25 & 0.44 & 0.22 \\
Mistral-7B-v0.1\_\_few\_shot\_cn\_cot\_cn & 0.56 & 0.06 & 0.34 & 0.49 & 0.41 & 0.43 & 0.51 & 0.40 \\
Mistral-7B-v0.1\_\_few\_shot\_cn\_cot\_en & 0.61 & 0.28 & 0.35 & 0.45 & 0.17 & 0.16 & 0.67 & 0.39 \\
Mistral-7B-v0.1\_\_few\_shot\_cn\_xlt\_en & 0.00 & 0.00 & 0.01 & 0.01 & 0.00 & 0.00 & 0.00 & 0.00 \\ \hline
Mixtral-8x7B-v0.1\_\_few\_shot\_en\_cot\_cn   & 0.94 & 0.60 & 0.80 & 0.90 & 0.81 & 0.66 & 0.65 & 0.77 \\
Mixtral-8x7B-v0.1\_\_few\_shot\_en\_cot\_en   & 0.91 & 0.45 & 0.59 & 0.82 & 0.56 & 0.54 & 0.87 & 0.68 \\
Mixtral-8x7B-v0.1\_\_few\_shot\_en\_xlt\_cn   & 0.15 & 0.04 & 0.18 & 0.18 & 0.34 & 0.23 & 0.24 & 0.19 \\
Mixtral-8x7B-v0.1\_\_few\_shot\_cn\_cot\_cn   & 0.75 & 0.13 & 0.35 & 0.47 & 0.52 & 0.42 & 0.76 & 0.48 \\
Mixtral-8x7B-v0.1\_\_few\_shot\_cn\_cot\_en   & 0.61 & 0.33 & 0.38 & 0.56 & 0.28 & 0.13 & 0.64 & 0.42 \\
Mixtral-8x7B-v0.1\_\_few\_shot\_cn\_xlt\_en   & 0.00 & 0.00 & 0.02 & 0.02 & 0.01 & 0.00 & 0.00 & 0.01 \\ \hline
  
Mixtral-8x22B-v0.1\_\_few\_shot\_en\_cot\_cn  & 0.96 & 0.63 & 0.90 & 0.92 & 0.86 & 0.69 & 0.87 & 0.83 \\
Mixtral-8x22B-v0.1\_\_few\_shot\_en\_cot\_en  & 0.92 & 0.59 & 0.69 & 0.88 & 0.63 & 0.57 & 0.78 & 0.72 \\
Mixtral-8x22B-v0.1\_\_few\_shot\_en\_xlt\_cn  & 0.26 & 0.08 & 0.36 & 0.43 & 0.36 & 0.31 & 0.33 & 0.30 \\
Mixtral-8x22B-v0.1\_\_few\_shot\_cn\_cot\_cn  & 0.60 & 0.08 & 0.36 & 0.37 & 0.57 & 0.32 & 0.82 & 0.45 \\
Mixtral-8x22B-v0.1\_\_few\_shot\_cn\_cot\_en  & 0.62 & 0.44 & 0.34 & 0.65 & 0.26 & 0.21 & 0.76 & 0.47 \\
Mixtral-8x22B-v0.1\_\_few\_shot\_cn\_xlt\_en  & 0.01 & 0.00 & 0.01 & 0.01 & 0.01 & 0.00 & 0.01 & 0.01 \\ \hline
Mistral-7B-v0.1\_\_zero\_shot\_en\_cot\_cn& 0.05 & 0.06 & 0.01 & 0.06 & 0.03 & 0.06 & 0.11 & 0.05 \\
Mistral-7B-v0.1\_\_zero\_shot\_en\_cot\_en& 0.27 & 0.31 & 0.02 & 0.19 & 0.08 & 0.19 & 0.11 & 0.17 \\
Mistral-7B-v0.1\_\_zero\_shot\_en\_xlt\_cn& 0.01 & 0.00 & 0.00 & 0.00 & 0.01 & 0.00 & 0.00 & 0.00 \\
Mistral-7B-v0.1\_\_zero\_shot\_cn\_cot\_cn& 0.21 & 0.08 & 0.01 & 0.17 & 0.08 & 0.00 & 0.00 & 0.08 \\
Mistral-7B-v0.1\_\_zero\_shot\_cn\_cot\_en& 0.11 & 0.04 & 0.02 & 0.08 & 0.04 & 0.05 & 0.08 & 0.06 \\
Mistral-7B-v0.1\_\_zero\_shot\_cn\_xlt\_en& 0.00 & 0.00 & 0.00 & 0.00 & 0.00 & 0.00 & 0.00 & 0.00 \\ \hline
Mixtral-8x7B-v0.1\_\_zero\_shot\_en\_cot\_en  & 0.66 & 0.25 & 0.07 & 0.47 & 0.24 & 0.40 & 0.29 & 0.34 \\
Mixtral-8x7B-v0.1\_\_zero\_shot\_en\_xlt\_cn  & 0.00 & 0.01 & 0.00 & 0.01 & 0.00 & 0.00 & 0.00 & 0.00 \\
Mixtral-8x7B-v0.1\_\_zero\_shot\_cn\_cot\_cn  & 0.52 & 0.20 & 0.04 & 0.38 & 0.27 & 0.28 & 0.14 & 0.26 \\
Mixtral-8x7B-v0.1\_\_zero\_shot\_cn\_cot\_en  & 0.48 & 0.17 & 0.01 & 0.31 & 0.15 & 0.23 & 0.09 & 0.21 \\
Mixtral-8x7B-v0.1\_\_zero\_shot\_cn\_xlt\_en  & 0.06 & 0.00 & 0.00 & 0.04 & 0.02 & 0.02 & 0.00 & 0.02 \\ \hline

Mixtral-8x22B-v0.1\_\_zero\_shot\_en\_cot\_cn & 0.89 & 0.57 & 0.05 & 0.83 & 0.60 & 0.51 & 0.57 & 0.57 \\
Mixtral-8x22B-v0.1\_\_zero\_shot\_en\_cot\_en & 0.82 & 0.33 & 0.13 & 0.52 & 0.48 & 0.36 & 0.58 & 0.46 \\
Mixtral-8x22B-v0.1\_\_zero\_shot\_en\_xlt\_cn & 0.00 & 0.00 & 0.00 & 0.00 & 0.00 & 0.00 & 0.00 & 0.00 \\
Mixtral-8x22B-v0.1\_\_zero\_shot\_cn\_cot\_cn & 0.45 & 0.27 & 0.15 & 0.39 & 0.33 & 0.26 & 0.21 & 0.29 \\
Mixtral-8x22B-v0.1\_\_zero\_shot\_cn\_cot\_en & 0.44 & 0.22 & 0.04 & 0.35 & 0.22 & 0.21 & 0.35 & 0.26 \\
Mixtral-8x22B-v0.1\_\_zero\_shot\_cn\_xlt\_en & 0.22 & 0.02 & 0.00 & 0.07 & 0.07 & 0.04 & 0.10 & 0.07 \\ \hline
\end{tabular}

}
 \caption{Performance of Mixtral Series. And NT, CT, CI, SO, SR, RC, PR are the abbreviations for the variant names of Negation Transformation, Critical Testing, Causal Inference, Sentence Ordering, Scenario Refinement, Reverse Conversion and Problem Restatement.}
 \label{mixtral}
\end{table*}

\begin{table*}[]
\centering
 \scalebox{0.9}{

\begin{tabular}{l|llllllll}
\hline
\textbf{Model\_\_Prompt\_Language}     & CI   & CT   & NT   & PR   & RC   & SR   &SO  & \textbf{ARA} \\ \hline
gemma-2-2b\_\_direct\_cn     & 0.58 & 0.33 & 0.08 & 0.64 & 0.55 & 0.41 & 0.59 & 0.45   \\
gemma-2-9b\_\_direct\_cn     & 0.59 & 0.33 & 0.06 & 0.68 & 0.59 & 0.47 & 0.55 & 0.47   \\
gemma-2-27\_\_direct\_cn     & 0.61 & 0.33 & 0.09 & 0.64 & 0.64 & 0.44 & 0.64 & 0.49   \\
gemma-2-2b\_\_direct\_en     & 0.49 & 0.38 & 0.05 & 0.74 & 0.35 & 0.35 & 0.39 & 0.39   \\
gemma-2-9b\_\_direct\_en     & 0.56 & 0.39 & 0.04 & 0.77 & 0.30 & 0.36 & 0.36 & 0.40   \\
gemma-2-27\_\_direct\_en     & 0.57 & 0.42 & 0.05 & 0.85 & 0.28 & 0.34 & 0.36 & 0.41   \\ \hline
   
gemma-2-2b\_\_few\_shot\_en\_cot\_cn   & 0.86 & 0.47 & 0.24 & 0.75 & 0.55 & 0.43 & 0.30 & 0.52   \\
gemma-2-2b\_\_few\_shot\_en\_cot\_en   & 0.69 & 0.25 & 0.48 & 0.59 & 0.37 & 0.40 & 0.80 & 0.51   \\
gemma-2-2b\_\_few\_shot\_en\_xlt\_cn   & 0.28 & 0.11 & 0.22 & 0.40 & 0.43 & 0.27 & 0.15 & 0.26   \\
gemma-2-2b\_\_few\_shot\_cn\_cot\_cn   & 0.70 & 0.45 & 0.31 & 0.71 & 0.50 & 0.33 & 0.46 & 0.49   \\
gemma-2-2b\_\_few\_shot\_cn\_cot\_en   & 0.85 & 0.41 & 0.33 & 0.27 & 0.31 & 0.26 & 0.55 & 0.43   \\
gemma-2-2b\_\_few\_shot\_cn\_xlt\_en   & 0.58 & 0.14 & 0.67 & 0.35 & 0.39 & 0.24 & 0.55 & 0.42   \\ \hline
   
gemma-2-9b\_\_few\_shot\_en\_cot\_cn   & 0.96 & 0.56 & 0.80 & 0.89 & 0.86 & 0.62 & 0.79 & 0.78   \\
gemma-2-9b\_\_few\_shot\_en\_cot\_en   & 0.88 & 0.50 & 0.79 & 0.84 & 0.51 & 0.55 & 0.80 & 0.69   \\
gemma-2-9b\_\_few\_shot\_en\_xlt\_cn   & 0.33 & 0.16 & 0.34 & 0.51 & 0.49 & 0.41 & 0.40 & 0.38   \\
gemma-2-9b\_\_few\_shot\_cn\_cot\_cn   & 0.93 & 0.47 & 0.77 & 0.85 & 0.80 & 0.54 & 0.87 & 0.75   \\
gemma-2-9b\_\_few\_shot\_cn\_cot\_en   & 0.91 & 0.46 & 0.71 & 0.67 & 0.50 & 0.44 & 0.76 & 0.63   \\
gemma-2-9b\_\_few\_shot\_cn\_xlt\_en   & 0.73 & 0.25 & 0.42 & 0.58 & 0.41 & 0.30 & 0.61 & 0.47   \\ \hline
   
gemma-2-27b\_\_few\_shot\_en\_cot\_cn  & 0.43 & 0.33 & 0.33 & 0.43 & 0.42 & 0.32 & 0.14 & 0.34   \\
gemma-2-27b\_\_few\_shot\_en\_cot\_en  & 0.34 & 0.15 & 0.24 & 0.27 & 0.29 & 0.21 & 0.11 & 0.23   \\
gemma-2-27b\_\_few\_shot\_en\_xlt\_cn  & 0.09 & 0.04 & 0.07 & 0.08 & 0.08 & 0.04 & 0.04 & 0.06   \\
gemma-2-27b\_\_few\_shot\_cn\_cot\_cn  & 0.30 & 0.06 & 0.15 & 0.21 & 0.23 & 0.16 & 0.22 & 0.19   \\
gemma-2-27b\_\_few\_shot\_cn\_cot\_en  & 0.34 & 0.10 & 0.18 & 0.23 & 0.21 & 0.13 & 0.21 & 0.20   \\
gemma-2-27b\_\_few\_shot\_cn\_xlt\_en  & 0.14 & 0.01 & 0.03 & 0.02 & 0.03 & 0.02 & 0.01 & 0.04   \\ \hline
   
gemma-2-2b\_\_zero\_shot\_en\_cot\_cn  & 0.31 & 0.19 & 0.08 & 0.15 & 0.19 & 0.09 & 0.15 & 0.17   \\
gemma-2-2b\_\_zero\_shot\_en\_cot\_en  & 0.30 & 0.28 & 0.04 & 0.17 & 0.20 & 0.16 & 0.43 & 0.22   \\
gemma-2-2b\_\_zero\_shot\_en\_xlt\_cn  & 0.43 & 0.17 & 0.11 & 0.25 & 0.25 & 0.15 & 0.07 & 0.20   \\
gemma-2-2b\_\_zero\_shot\_cn\_cot\_cn  & 0.10 & 0.13 & 0.06 & 0.08 & 0.05 & 0.00 & 0.03 & 0.06   \\
gemma-2-2b\_\_zero\_shot\_cn\_cot\_en  & 0.07 & 0.04 & 0.09 & 0.05 & 0.07 & 0.03 & 0.01 & 0.05   \\
gemma-2-2b\_\_zero\_shot\_cn\_xlt\_en  & 0.00 & 0.00 & 0.00 & 0.00 & 0.00 & 0.00 & 0.00 & 0.00   \\ \hline
   
gemma-2-9b\_\_zero\_shot\_en\_cot\_cn  & 0.66 & 0.42 & 0.06 & 0.62 & 0.53 & 0.34 & 0.40 & 0.43   \\
gemma-2-9b\_\_zero\_shot\_en\_cot\_en  & 0.71 & 0.37 & 0.04 & 0.52 & 0.27 & 0.37 & 0.53 & 0.40   \\
gemma-2-9b\_\_zero\_shot\_en\_xlt\_cn  & 0.04 & 0.02 & 0.02 & 0.05 & 0.11 & 0.01 & 0.00 & 0.04   \\
gemma-2-9b\_\_zero\_shot\_cn\_cot\_cn  & 0.51 & 0.36 & 0.09 & 0.54 & 0.45 & 0.29 & 0.39 & 0.37   \\
gemma-2-9b\_\_zero\_shot\_cn\_cot\_en  & 0.64 & 0.17 & 0.03 & 0.36 & 0.23 & 0.23 & 0.12 & 0.25   \\
gemma-2-9b\_\_zero\_shot\_cn\_xlt\_en  & 0.33 & 0.01 & 0.00 & 0.04 & 0.12 & 0.02 & 0.04 & 0.08   \\ \hline
   
gemma-2-27b\_\_zero\_shot\_en\_cot\_cn & 0.37 & 0.18 & 0.08 & 0.31 & 0.29 & 0.24 & 0.15 & 0.23   \\
gemma-2-27b\_\_zero\_shot\_en\_cot\_en & 0.21 & 0.07 & 0.05 & 0.17 & 0.14 & 0.13 & 0.12 & 0.13   \\
gemma-2-27b\_\_zero\_shot\_en\_xlt\_cn & 0.22 & 0.13 & 0.05 & 0.13 & 0.14 & 0.09 & 0.03 & 0.11   \\
gemma-2-27b\_\_zero\_shot\_cn\_cot\_cn & 0.15 & 0.08 & 0.05 & 0.11 & 0.13 & 0.00 & 0.07 & 0.08   \\
gemma-2-27b\_\_zero\_shot\_cn\_cot\_en & 0.15 & 0.04 & 0.04 & 0.05 & 0.06 & 0.04 & 0.05 & 0.06   \\
gemma-2-27b\_\_zero\_shot\_cn\_xlt\_en & 0.03 & 0.01 & 0.02 & 0.01 & 0.03 & 0.01 & 0.01 & 0.02   \\ \hline
\end{tabular}

}
 \caption{Performance of Gemma Series. And NT, CT, CI, SO, SR, RC, PR are the abbreviations for the variant names of Negation Transformation, Critical Testing, Causal Inference, Sentence Ordering, Scenario Refinement, Reverse Conversion and Problem Restatement.}
 \label{gemma}
\end{table*}

\begin{table*}[]
\centering
 \scalebox{0.9}{
\begin{tabular}{l|llllllll}
\hline
\multicolumn{1}{l|}{\textbf{Model\_\_Prompt\_Language}}   & CI   & CT   & NT   & PR   & RC   & SR   &SO  & \textbf{ARA} \\ \hline
Baichuan2-7B-Base\_\_\_direct\_cn & 0.59 & 0.34 & 0.09 & 0.65 & 0.60 & 0.41 & 0.55 & 0.46 \\
Baichuan2-13B-Base\_\_\_direct\_cn& 0.62 & 0.37 & 0.08 & 0.66 & 0.60 & 0.43 & 0.54 & 0.47 \\
Baichuan2-7B-Base\_\_\_direct\_en & 0.51 & 0.38 & 0.05 & 0.76 & 0.31 & 0.34 & 0.38 & 0.39 \\
Baichuan2-13B-Base\_\_\_direct\_en& 0.47 & 0.40 & 0.05 & 0.80 & 0.30 & 0.34 & 0.35 & 0.39 \\ \hline
Baichuan2-7B-Base\_\_few\_shot\_en\_cot\_cn   & 0.82 & 0.49 & 0.32 & 0.72 & 0.62 & 0.44 & 0.55 & 0.56 \\
Baichuan2-7B-Base\_\_few\_shot\_en\_cot\_en   & 0.67 & 0.22 & 0.70 & 0.56 & 0.24 & 0.41 & 0.68 & 0.50 \\
Baichuan2-7B-Base\_\_few\_shot\_en\_xlt\_cn   & 0.15 & 0.01 & 0.23 & 0.25 & 0.39 & 0.23 & 0.30 & 0.22 \\
Baichuan2-7B-Base\_\_few\_shot\_cn\_cot\_cn   & 0.89 & 0.50 & 0.54 & 0.75 & 0.70 & 0.53 & 0.51 & 0.63 \\
Baichuan2-7B-Base\_\_few\_shot\_cn\_cot\_en   & 0.83 & 0.28 & 0.75 & 0.50 & 0.50 & 0.33 & 0.70 & 0.56 \\
Baichuan2-7B-Base\_\_few\_shot\_cn\_xlt\_en   & 0.48 & 0.20 & 0.22 & 0.56 & 0.49 & 0.33 & 0.41 & 0.38 \\ \hline

Baichuan2-13B-Base\_\_few\_shot\_en\_cot\_cn  & 0.94 & 0.56 & 0.71 & 0.84 & 0.78 & 0.60 & 0.58 & 0.72 \\
Baichuan2-13B-Base\_\_few\_shot\_en\_cot\_en  & 0.78 & 0.31 & 0.36 & 0.56 & 0.48 & 0.47 & 0.81 & 0.54 \\
Baichuan2-13B-Base\_\_few\_shot\_en\_xlt\_cn  & 0.22 & 0.13 & 0.39 & 0.53 & 0.47 & 0.30 & 0.41 & 0.35 \\
Baichuan2-13B-Base\_\_few\_shot\_cn\_cot\_cn  & 0.89 & 0.53 & 0.76 & 0.86 & 0.72 & 0.53 & 0.61 & 0.70 \\
Baichuan2-13B-Base\_\_few\_shot\_cn\_cot\_en  & 0.86 & 0.42 & 0.52 & 0.57 & 0.47 & 0.35 & 0.87 & 0.58 \\
Baichuan2-13B-Base\_\_few\_shot\_cn\_xlt\_en  & 0.83 & 0.26 & 0.25 & 0.49 & 0.44 & 0.34 & 0.74 & 0.48 \\ \hline

Baichuan2-7B-Base\_\_zero\_shot\_en\_cot\_cn  & 0.31 & 0.16 & 0.01 & 0.24 & 0.21 & 0.13 & 0.03 & 0.16 \\
Baichuan2-7B-Base\_\_zero\_shot\_en\_cot\_en  & 0.29 & 0.20 & 0.01 & 0.35 & 0.08 & 0.22 & 0.01 & 0.16 \\
Baichuan2-7B-Base\_\_zero\_shot\_en\_xlt\_cn  & 0.19 & 0.21 & 0.02 & 0.17 & 0.10 & 0.06 & 0.01 & 0.11 \\
Baichuan2-7B-Base\_\_zero\_shot\_cn\_cot\_cn  & 0.42 & 0.23 & 0.07 & 0.38 & 0.22 & 0.00 & 0.02 & 0.19 \\
Baichuan2-7B-Base\_\_zero\_shot\_cn\_cot\_en  & 0.22 & 0.11 & 0.11 & 0.29 & 0.13 & 0.16 & 0.03 & 0.15 \\
Baichuan2-7B-Base\_\_zero\_shot\_cn\_xlt\_en  & 0.02 & 0.00 & 0.00 & 0.00 & 0.00 & 0.00 & 0.00 & 0.00 \\ \hline

Baichuan2-13B-Base\_\_zero\_shot\_en\_cot\_cn & 0.41 & 0.21 & 0.02 & 0.25 & 0.54 & 0.11 & 0.21 & 0.25 \\
Baichuan2-13B-Base\_\_zero\_shot\_en\_cot\_en & 0.31 & 0.38 & 0.05 & 0.45 & 0.10 & 0.35 & 0.13 & 0.25 \\
Baichuan2-13B-Base\_\_zero\_shot\_en\_xlt\_cn & 0.23 & 0.19 & 0.01 & 0.28 & 0.20 & 0.05 & 0.10 & 0.15 \\
Baichuan2-13B-Base\_\_zero\_shot\_cn\_cot\_cn & 0.41 & 0.17 & 0.07 & 0.32 & 0.30 & 0.16 & 0.01 & 0.21 \\
Baichuan2-13B-Base\_\_zero\_shot\_cn\_cot\_en & 0.13 & 0.10 & 0.06 & 0.10 & 0.14 & 0.07 & 0.04 & 0.09 \\
Baichuan2-13B-Base\_\_zero\_shot\_cn\_xlt\_en & 0.00 & 0.00 & 0.00 & 0.00 & 0.00 & 0.00 & 0.00 & 0.00 \\ \hline
\end{tabular}

}
 \caption{Performance of Baichuan Series. And NT, CT, CI, SO, SR, RC, PR are the abbreviations for the variant names of Negation Transformation, Critical Testing, Causal Inference, Sentence Ordering, Scenario Refinement, Reverse Conversion and Problem Restatement.}
 \label{baichuan}
\end{table*}

\begin{table*}[]
\centering
\begin{tabular}{l|llllllll}
\hline
\textbf{Model\_\_Prompt\_Language}& CI   & CT   & NT   & PR   & RC   & SR   &SO  & \textbf{ARA} \\ \hline
internlm2\_5-1\_8b\_\_direct\_cn& 0.62 & 0.33 & 0.09 & 0.63 & 0.52 & 0.37 & 0.38 & 0.42   \\
internlm2\_5-7b\_\_direct\_cn & 0.71 & 0.34 & 0.08 & 0.66 & 0.60 & 0.43 & 0.67 & 0.50   \\
internlm2\_5-20b\_\_direct\_cn& 0.69 & 0.36 & 0.08 & 0.70 & 0.66 & 0.42 & 0.45 & 0.48   \\
internlm2\_5-1\_8b\_\_direct\_en& 0.55 & 0.33 & 0.08 & 0.63 & 0.41 & 0.35 & 0.35 & 0.38   \\
internlm2\_5-7b\_\_direct\_en & 0.59 & 0.39 & 0.06 & 0.76 & 0.36 & 0.32 & 0.37 & 0.41   \\
internlm2\_5-20b\_\_direct\_en& 0.62 & 0.41 & 0.05 & 0.78 & 0.35 & 0.31 & 0.37 & 0.41   \\ \hline
   
internlm2\_5-1\_8b\_\_few\_shot\_en\_cot\_cn  & 0.89 & 0.47 & 0.39 & 0.83 & 0.73 & 0.45 & 0.40 & 0.60   \\
internlm2\_5-1\_8b\_\_few\_shot\_en\_cot\_en  & 0.84 & 0.32 & 0.37 & 0.56 & 0.39 & 0.45 & 0.42 & 0.48   \\
internlm2\_5-1\_8b\_\_few\_shot\_en\_xlt\_cn  & 0.44 & 0.27 & 0.47 & 0.48 & 0.53 & 0.35 & 0.40 & 0.42   \\
internlm2\_5-1\_8b\_\_few\_shot\_cn\_cot\_cn  & 0.86 & 0.43 & 0.55 & 0.80 & 0.67 & 0.44 & 0.50 & 0.61   \\
internlm2\_5-1\_8b\_\_few\_shot\_cn\_cot\_en  & 0.80 & 0.44 & 0.20 & 0.49 & 0.41 & 0.45 & 0.67 & 0.49   \\
internlm2\_5-1\_8b\_\_few\_shot\_cn\_xlt\_en  & 0.62 & 0.20 & 0.17 & 0.48 & 0.47 & 0.35 & 0.42 & 0.39   \\ \hline
   
internlm2\_5-7b\_\_few\_shot\_en\_cot\_cn& 0.78 & 0.65 & 0.88 & 0.84 & 0.87 & 0.63 & 0.77 & 0.77   \\
internlm2\_5-7b\_\_few\_shot\_en\_cot\_en& 0.93 & 0.52 & 0.85 & 0.84 & 0.53 & 0.53 & 0.88 & 0.72   \\
internlm2\_5-7b\_\_few\_shot\_en\_xlt\_cn& 0.62 & 0.41 & 0.61 & 0.57 & 0.59 & 0.51 & 0.42 & 0.53   \\
internlm2\_5-7b\_\_few\_shot\_cn\_cot\_cn& 0.95 & 0.52 & 0.90 & 0.86 & 0.85 & 0.61 & 0.84 & 0.79   \\
internlm2\_5-7b\_\_few\_shot\_cn\_cot\_en& 0.91 & 0.60 & 0.75 & 0.77 & 0.47 & 0.54 & 0.82 & 0.69   \\
internlm2\_5-7b\_\_few\_shot\_cn\_xlt\_en& 0.78 & 0.31 & 0.48 & 0.77 & 0.40 & 0.38 & 0.62 & 0.54   \\ \hline
   
internlm2\_5-20b\_\_few\_shot\_en\_cot\_cn& 0.62 & 0.58 & 0.92 & 0.28 & 0.00 & 0.48 & 0.79 & 0.52   \\
internlm2\_5-20b\_\_few\_shot\_en\_cot\_en& 0.91 & 0.63 & 0.85 & 0.88 & 0.53 & 0.55 & 0.82 & 0.74   \\
internlm2\_5-20b\_\_few\_shot\_en\_xlt\_cn& 0.66 & 0.38 & 0.63 & 0.42 & 0.54 & 0.59 & 0.18 & 0.48   \\
internlm2\_5-20b\_\_few\_shot\_cn\_cot\_cn& 0.00 & 0.27 & 0.91 & 0.01 & 0.00 & 0.00 & 0.57 & 0.25   \\
internlm2\_5-20b\_\_few\_shot\_cn\_cot\_en& 0.91 & 0.65 & 0.79 & 0.86 & 0.51 & 0.56 & 0.84 & 0.73   \\
internlm2\_5-20b\_\_few\_shot\_cn\_xlt\_en& 0.74 & 0.46 & 0.90 & 0.83 & 0.53 & 0.44 & 0.69 & 0.65   \\ \hline
   
internlm2\_5-1\_8b\_\_zero\_shot\_en\_cot\_cn & 0.51 & 0.22 & 0.03 & 0.42 & 0.31 & 0.26 & 0.17 & 0.27   \\
internlm2\_5-1\_8b\_\_zero\_shot\_en\_cot\_en & 0.15 & 0.06 & 0.06 & 0.16 & 0.14 & 0.13 & 0.28 & 0.14   \\
internlm2\_5-1\_8b\_\_zero\_shot\_en\_xlt\_cn & 0.06 & 0.02 & 0.00 & 0.02 & 0.05 & 0.03 & 0.00 & 0.03   \\
internlm2\_5-1\_8b\_\_zero\_shot\_cn\_cot\_cn & 0.67 & 0.42 & 0.17 & 0.51 & 0.40 & 0.50 & 0.34 & 0.43   \\
internlm2\_5-1\_8b\_\_zero\_shot\_cn\_cot\_en & 0.45 & 0.32 & 0.20 & 0.32 & 0.32 & 0.28 & 0.47 & 0.34   \\
internlm2\_5-1\_8b\_\_zero\_shot\_cn\_xlt\_en & 0.07 & 0.04 & 0.08 & 0.07 & 0.03 & 0.06 & 0.01 & 0.05   \\ \hline
   
internlm2\_5-7b\_\_zero\_shot\_en\_cot\_cn& 0.78 & 0.65 & 0.88 & 0.84 & 0.87 & 0.63 & 0.77 & 0.77   \\
internlm2\_5-7b\_\_zero\_shot\_en\_cot\_en& 0.93 & 0.52 & 0.85 & 0.84 & 0.53 & 0.53 & 0.88 & 0.72   \\
internlm2\_5-7b\_\_zero\_shot\_en\_xlt\_cn& 0.15 & 0.08 & 0.12 & 0.12 & 0.18 & 0.11 & 0.14 & 0.13   \\
internlm2\_5-7b\_\_zero\_shot\_cn\_cot\_cn& 0.95 & 0.52 & 0.90 & 0.86 & 0.85 & 0.61 & 0.84 & 0.79   \\
internlm2\_5-7b\_\_zero\_shot\_cn\_cot\_en& 0.91 & 0.60 & 0.75 & 0.77 & 0.47 & 0.54 & 0.82 & 0.69   \\
internlm2\_5-7b\_\_zero\_shot\_cn\_xlt\_en& 0.00 & 0.00 & 0.00 & 0.00 & 0.00 & 0.00 & 0.00 & 0.00   \\ \hline
   
internlm2\_5-20b\_\_zero\_shot\_en\_cot\_cn   & 0.39 & 0.36 & 0.15 & 0.48 & 0.55 & 0.31 & 0.22 & 0.35   \\
internlm2\_5-20b\_\_zero\_shot\_en\_cot\_en   & 0.48 & 0.28 & 0.11 & 0.48 & 0.41 & 0.25 & 0.75 & 0.39   \\
internlm2\_5-20b\_\_zero\_shot\_en\_xlt\_cn   & 0.27 & 0.33 & 0.33 & 0.54 & 0.42 & 0.26 & 0.59 & 0.39   \\
internlm2\_5-20b\_\_zero\_shot\_cn\_cot\_cn   & 0.55 & 0.53 & 0.27 & 0.71 & 0.56 & 0.00 & 0.71 & 0.48   \\
internlm2\_5-20b\_\_zero\_shot\_cn\_cot\_en   & 0.54 & 0.47 & 0.03 & 0.69 & 0.43 & 0.42 & 0.30 & 0.41   \\
internlm2\_5-20b\_\_zero\_shot\_cn\_xlt\_en   & 0.80 & 0.18 & 0.02 & 0.64 & 0.43 & 0.32 & 0.89 & 0.47   \\ \hline
\end{tabular}
 \caption{Performance of InternLM Series. And NT, CT, CI, SO, SR, RC, PR are the abbreviations for the variant names of Negation Transformation, Critical Testing, Causal Inference, Sentence Ordering, Scenario Refinement, Reverse Conversion and Problem Restatement.}
 \label{internlm}
\end{table*}

\begin{table*}[]
\centering
 \scalebox{0.9}{
\begin{tabular}{l|llllllll}
\hline
\textbf{Instruct-Model} & CI   & CT   & NT   & PR   & RC   & SR   &SO  & \textbf{ARA} \\ \hline
Qwen2.5-0.5B            & 0.58 & 0.36 & 0.07 & 0.65 & 0.41 & 0.36 & 0.58 & 0.43         \\
Qwen2.5-1.5B            & 0.64 & 0.36 & 0.07 & 0.70 & 0.41 & 0.37 & 0.49 & 0.43         \\
Qwen2.5-3B              & 0.66 & 0.40 & 0.05 & 0.75 & 0.41 & 0.40 & 0.53 & 0.46         \\
Qwen2.5-7B              & 0.74 & 0.39 & 0.06 & 0.76 & 0.44 & 0.39 & 0.64 & 0.49         \\
Qwen2.5-14B             & 0.78 & 0.40 & 0.06 & 0.77 & 0.44 & 0.40 & 0.74 & 0.51         \\
Qwen2.5-32B             & 0.75 & 0.40 & 0.06 & 0.78 & 0.45 & 0.40 & 0.79 & 0.52         \\
Qwen2.5-72B             & 0.78 & 0.41 & 0.06 & 0.79 & 0.48 & 0.41 & 0.80 & 0.53         \\ \hline
Meta-Llama-3-8B            & 0.65 & 0.38 & 0.07 & 0.71 & 0.34 & 0.40 & 0.52 & 0.44         \\
Meta-Llama-3-70B             & 0.68 & 0.40 & 0.06 & 0.73 & 0.37 & 0.41 & 0.57 & 0.46         \\ \hline
Mistral-7B-v0.2              & 0.67 & 0.41 & 0.07 & 0.76 & 0.35 & 0.38 & 0.58 & 0.46         \\
Mixtral-8x7B-v0.1            & 0.65 & 0.41 & 0.07 & 0.76 & 0.41 & 0.40 & 0.52 & 0.46         \\
Mixtral-8x22B-v0.1           & 0.70 & 0.43 & 0.05 & 0.78 & 0.42 & 0.39 & 0.57 & 0.48         \\ \hline
\end{tabular}
}
 \caption{Performance of intruct models under Direct Prompt. And NT, CT, CI, SO, SR, RC, PR are the abbreviations for the variant names of Negation Transformation, Critical Testing, Causal Inference, Sentence Ordering, Scenario Refinement, Reverse Conversion and Problem Restatement.}
 \label{instruct-tab}
\end{table*}

\clearpage

\begin{figure}[H]
\centering
\setlength{\abovecaptionskip}{0.1cm}
\setlength{\belowcaptionskip}{0cm}
\includegraphics[width=\linewidth,scale=1.00]{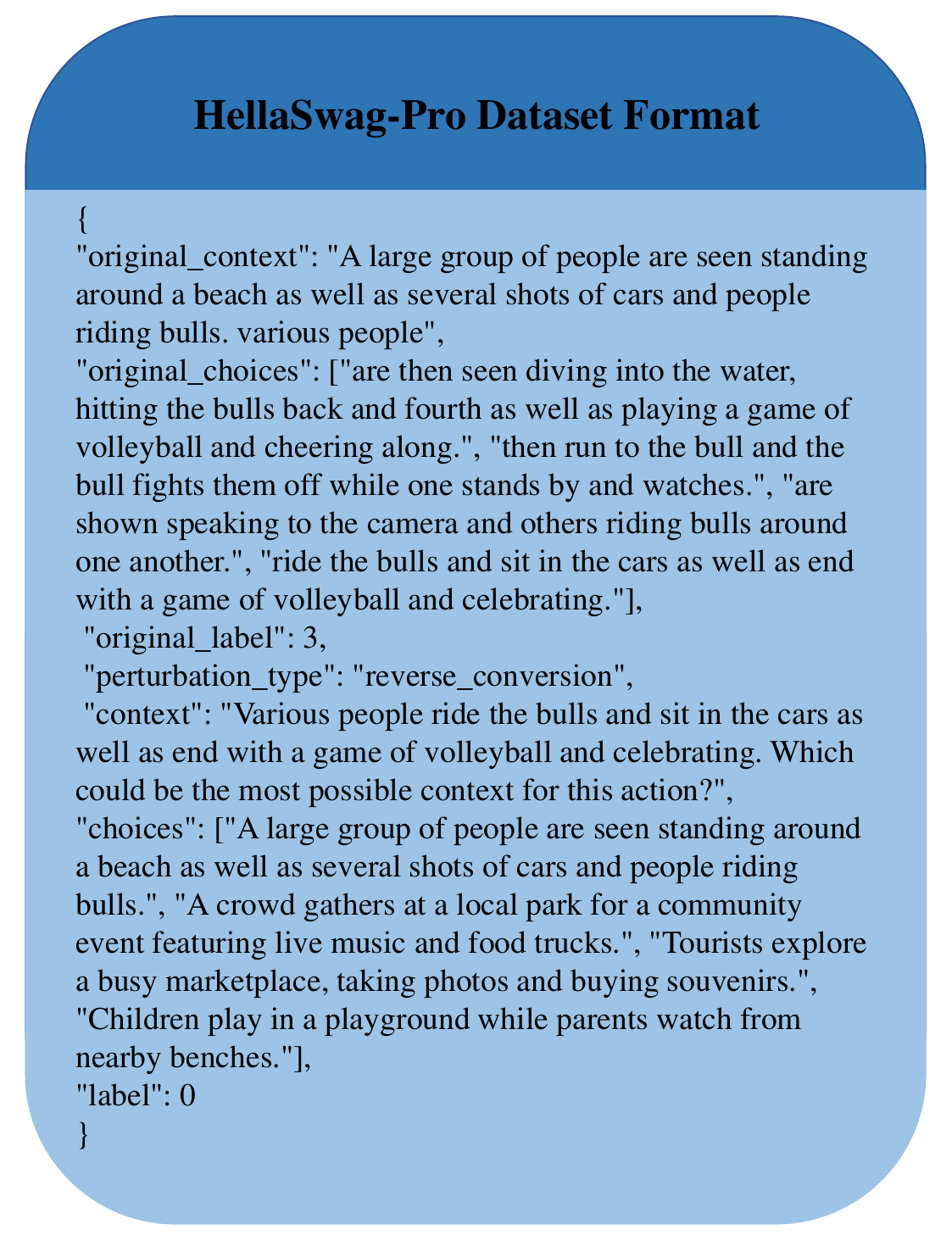}
\caption{An example of HellaSwag-Pro.}
\label{robust_data}
\vspace{-15pt}
\end{figure}

\begin{figure}[H]
\centering
\setlength{\abovecaptionskip}{0.1cm}
\setlength{\belowcaptionskip}{0cm}
\includegraphics[width=\linewidth,scale=1.00]{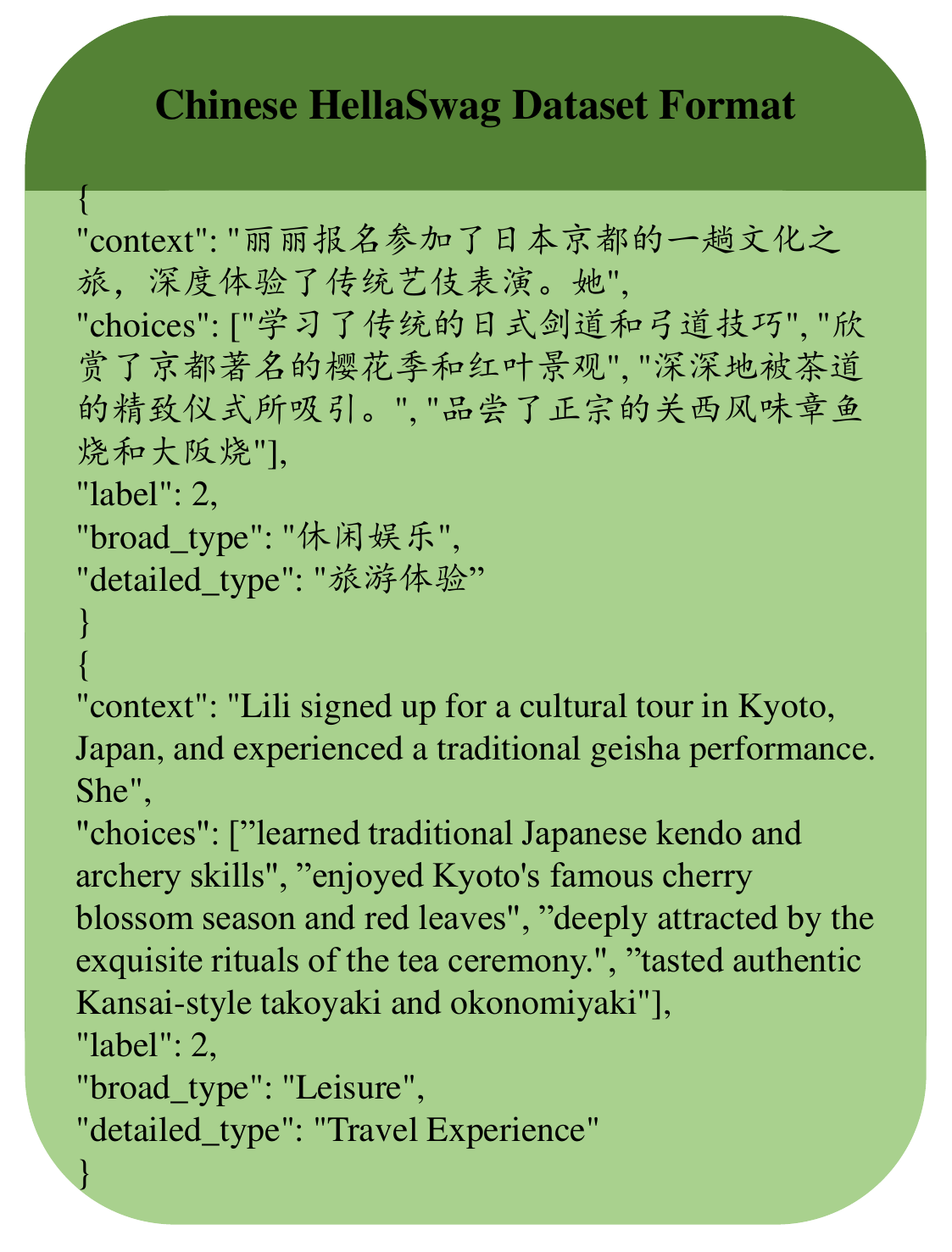}
\caption{An example of Chinese HellaSwag.}
\label{cn_data}
\vspace{-15pt}
\end{figure}

\begin{figure}[H]
\setlength{\abovecaptionskip}{0.1cm}
\setlength{\belowcaptionskip}{0cm}
\includegraphics[width=\linewidth,scale=1.00]{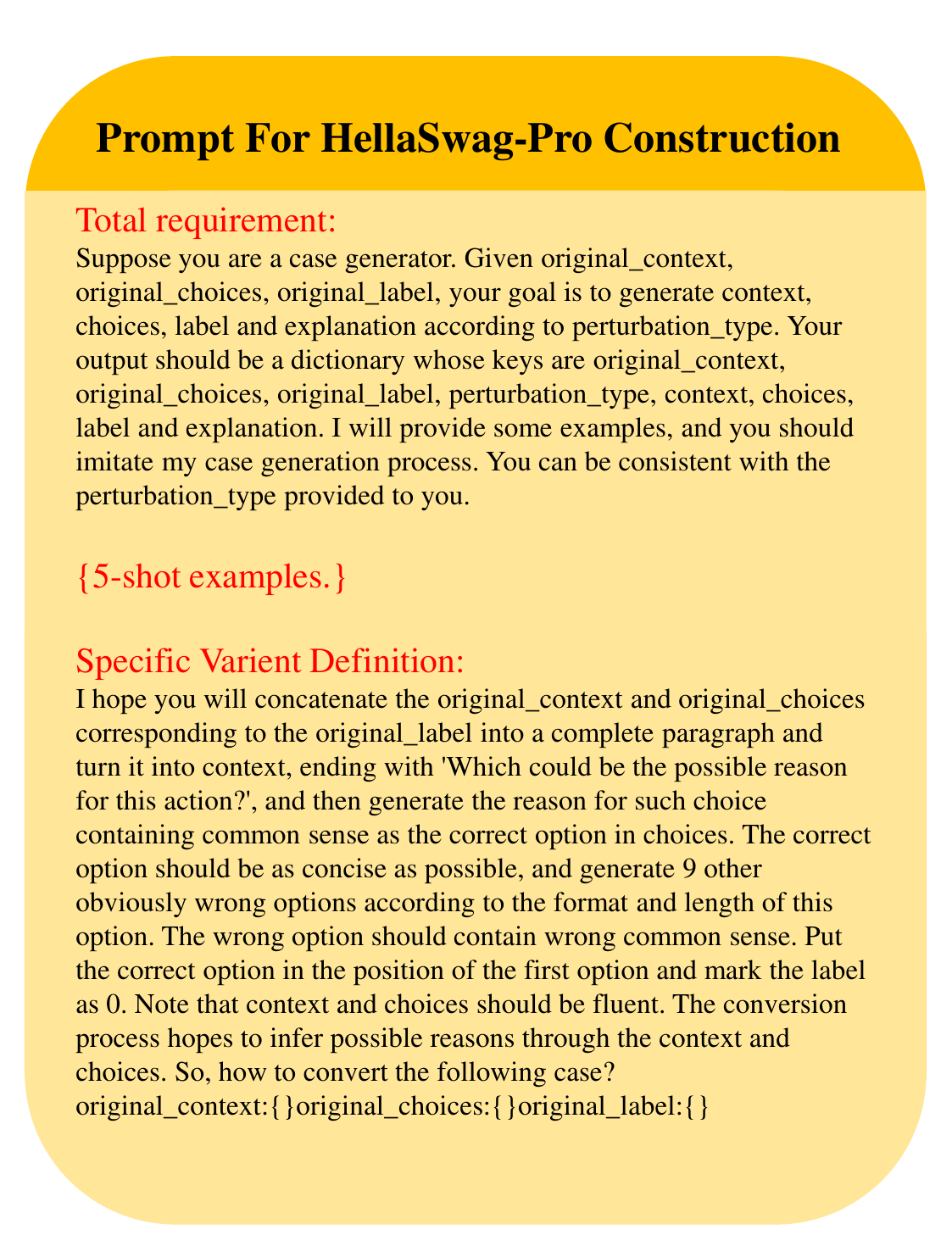}
\caption{Prompt for HellaSwag-Pro construction.}
\label{pro_prompt}
\vspace{-15pt}
\end{figure}

\begin{figure}[H]
\setlength{\abovecaptionskip}{0.1cm}
\setlength{\belowcaptionskip}{0cm}
\includegraphics[width=\linewidth,scale=1.00]{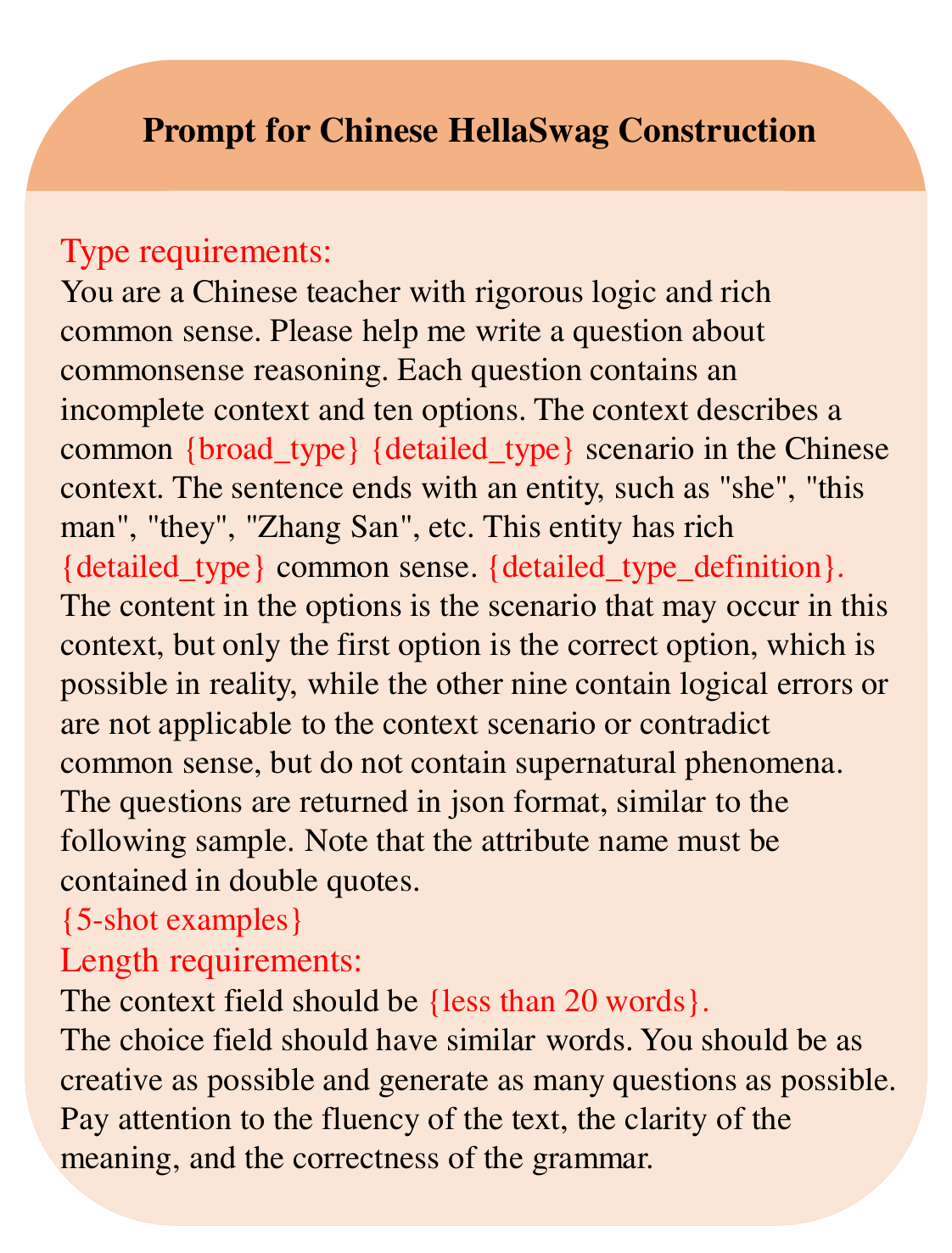}
\caption{Prompt for Chinese HellaSwag construction.}
\label{cn_prompt}
\vspace{-15pt}
\end{figure}

%% file: acl_latex.bbl
\begin{thebibliography}{61}
\expandafter\ifx\csname natexlab\endcsname\relax\def\natexlab#1{#1}\fi

\bibitem[{Anil et~al.(2023)Anil, Borgeaud, Wu, Alayrac, Yu, Soricut, Schalkwyk, Dai, Hauth, Millican, Silver, Petrov, Johnson, Antonoglou, Schrittwieser, Glaese, Chen, Pitler, Lillicrap, Lazaridou, Firat, Molloy, Isard, Barham, Hennigan, Lee, Viola, Reynolds, Xu, Doherty, Collins, Meyer, Rutherford, Moreira, Ayoub, Goel, Tucker, Piqueras, Krikun, Barr, Savinov, Danihelka, Roelofs, White, Andreassen, von Glehn, Yagati, Kazemi, Gonzalez, Khalman, Sygnowski, and et~al.}]{DBLP:journals/corr/abs-2312-11805}
Rohan Anil, Sebastian Borgeaud, Yonghui Wu, Jean{-}Baptiste Alayrac, Jiahui Yu, Radu Soricut, Johan Schalkwyk, Andrew~M. Dai, Anja Hauth, Katie Millican, David Silver, Slav Petrov, Melvin Johnson, Ioannis Antonoglou, Julian Schrittwieser, Amelia Glaese, Jilin Chen, Emily Pitler, Timothy~P. Lillicrap, Angeliki Lazaridou, Orhan Firat, James Molloy, Michael Isard, Paul~Ronald Barham, Tom Hennigan, Benjamin Lee, Fabio Viola, Malcolm Reynolds, Yuanzhong Xu, Ryan Doherty, Eli Collins, Clemens Meyer, Eliza Rutherford, Erica Moreira, Kareem Ayoub, Megha Goel, George Tucker, Enrique Piqueras, Maxim Krikun, Iain Barr, Nikolay Savinov, Ivo Danihelka, Becca Roelofs, Ana{\"{\i}}s White, Anders Andreassen, Tamara von Glehn, Lakshman Yagati, Mehran Kazemi, Lucas Gonzalez, Misha Khalman, Jakub Sygnowski, and et~al. 2023.
\newblock \href {https://doi.org/10.48550/ARXIV.2312.11805} {Gemini: {A} family of highly capable multimodal models}.
\newblock \emph{CoRR}, abs/2312.11805.

\bibitem[{Anthropic(2024)}]{anthropic2024claude}
Anthropic. 2024.
\newblock Introducing the next generation of claude.
\newblock \url{https://www.anthropic.com/news/claude-3-family}.

\bibitem[{Bai et~al.(2023)Bai, Bai, Chu, Cui, Dang, Deng, Fan, Ge, Han, Huang, Hui, Ji, Li, Lin, Lin, Liu, Liu, Lu, Lu, Ma, Men, Ren, Ren, Tan, Tan, Tu, Wang, Wang, Wang, Wu, Xu, Xu, Yang, Yang, Yang, Yang, Yao, Yu, Yuan, Yuan, Zhang, Zhang, Zhang, Zhang, Zhou, Zhou, Zhou, and Zhu}]{qwen}
Jinze Bai, Shuai Bai, Yunfei Chu, Zeyu Cui, Kai Dang, Xiaodong Deng, Yang Fan, Wenbin Ge, Yu~Han, Fei Huang, Binyuan Hui, Luo Ji, Mei Li, Junyang Lin, Runji Lin, Dayiheng Liu, Gao Liu, Chengqiang Lu, Keming Lu, Jianxin Ma, Rui Men, Xingzhang Ren, Xuancheng Ren, Chuanqi Tan, Sinan Tan, Jianhong Tu, Peng Wang, Shijie Wang, Wei Wang, Shengguang Wu, Benfeng Xu, Jin Xu, An~Yang, Hao Yang, Jian Yang, Shusheng Yang, Yang Yao, Bowen Yu, Hongyi Yuan, Zheng Yuan, Jianwei Zhang, Xingxuan Zhang, Yichang Zhang, Zhenru Zhang, Chang Zhou, Jingren Zhou, Xiaohuan Zhou, and Tianhang Zhu. 2023.
\newblock Qwen technical report.
\newblock \emph{arXiv preprint arXiv:2309.16609}.

\bibitem[{Balepur et~al.(2024)Balepur, Palta, and Rudinger}]{balepur2024s}
Nishant Balepur, Shramay Palta, and Rachel Rudinger. 2024.
\newblock It’s not easy being wrong: Large language models struggle with process of elimination reasoning.
\newblock In \emph{Findings of the Association for Computational Linguistics ACL 2024}, pages 10143--10166.

\bibitem[{Bhagavatula et~al.(2020)Bhagavatula, Bras, Malaviya, Sakaguchi, Holtzman, Rashkin, Downey, Yih, and Choi}]{DBLP:conf/iclr/BhagavatulaBMSH20}
Chandra Bhagavatula, Ronan~Le Bras, Chaitanya Malaviya, Keisuke Sakaguchi, Ari Holtzman, Hannah Rashkin, Doug Downey, Wen{-}tau Yih, and Yejin Choi. 2020.
\newblock \href {https://openreview.net/forum?id=Byg1v1HKDB} {Abductive commonsense reasoning}.
\newblock In \emph{8th International Conference on Learning Representations, {ICLR} 2020, Addis Ababa, Ethiopia, April 26-30, 2020}. OpenReview.net.

\bibitem[{Bi et~al.(2024)Bi, Chen, Chen, Chen, Dai, Deng, Ding, Dong, Du, Fu, Gao, Gao, Gao, Ge, Guan, Guo, Guo, Hao, Hao, He, Hu, Huang, Li, Li, Li, Li, Li, Liang, Lin, Liu, Liu, Liu, Liu, Liu, Liu, Lu, Lu, Luo, Ma, Nie, Pei, Piao, Qiu, Qu, Ren, Ren, Ruan, Sha, Shao, Song, Su, Sun, Sun, Tang, Wang, Wang, Wang, Wang, Wang, Wu, Wu, Xie, Xie, Xie, Xiong, Xu, Xu, Xu, Yang, You, Yu, Yu, Zhang, Zhang, Zhang, Zhang, Zhang, Zhang, Zhang, Zhang, Zhao, Zhao, Zhou, Zhou, Zhu, and Zou}]{DBLP:journals/corr/abs-2401-02954}
Xiao Bi, Deli Chen, Guanting Chen, Shanhuang Chen, Damai Dai, Chengqi Deng, Honghui Ding, Kai Dong, Qiushi Du, Zhe Fu, Huazuo Gao, Kaige Gao, Wenjun Gao, Ruiqi Ge, Kang Guan, Daya Guo, Jianzhong Guo, Guangbo Hao, Zhewen Hao, Ying He, Wenjie Hu, Panpan Huang, Erhang Li, Guowei Li, Jiashi Li, Yao Li, Y.~K. Li, Wenfeng Liang, Fangyun Lin, Alex~X. Liu, Bo~Liu, Wen Liu, Xiaodong Liu, Xin Liu, Yiyuan Liu, Haoyu Lu, Shanghao Lu, Fuli Luo, Shirong Ma, Xiaotao Nie, Tian Pei, Yishi Piao, Junjie Qiu, Hui Qu, Tongzheng Ren, Zehui Ren, Chong Ruan, Zhangli Sha, Zhihong Shao, Junxiao Song, Xuecheng Su, Jingxiang Sun, Yaofeng Sun, Minghui Tang, Bingxuan Wang, Peiyi Wang, Shiyu Wang, Yaohui Wang, Yongji Wang, Tong Wu, Y.~Wu, Xin Xie, Zhenda Xie, Ziwei Xie, Yiliang Xiong, Hanwei Xu, R.~X. Xu, Yanhong Xu, Dejian Yang, Yuxiang You, Shuiping Yu, Xingkai Yu, B.~Zhang, Haowei Zhang, Lecong Zhang, Liyue Zhang, Mingchuan Zhang, Minghua Zhang, Wentao Zhang, Yichao Zhang, Chenggang Zhao, Yao Zhao, Shangyan Zhou, Shunfeng Zhou, Qihao
  Zhu, and Yuheng Zou. 2024.
\newblock \href {https://doi.org/10.48550/ARXIV.2401.02954} {Deepseek {LLM:} scaling open-source language models with longtermism}.
\newblock \emph{CoRR}, abs/2401.02954.

\bibitem[{Bisk et~al.(2020)Bisk, Zellers, Gao, Choi et~al.}]{bisk2020piqa}
Yonatan Bisk, Rowan Zellers, Jianfeng Gao, Yejin Choi, et~al. 2020.
\newblock Piqa: Reasoning about physical commonsense in natural language.
\newblock In \emph{Proceedings of the AAAI conference on artificial intelligence}, volume~34, pages 7432--7439.

\bibitem[{Branco et~al.(2021)Branco, Branco, Rodrigues, and Silva}]{branco2021shortcutted}
Ruben Branco, Ant{\'o}nio Branco, Joao Rodrigues, and Joao Silva. 2021.
\newblock Shortcutted commonsense: Data spuriousness in deep learning of commonsense reasoning.
\newblock In \emph{Proceedings of the 2021 Conference on Empirical Methods in Natural Language Processing}, pages 1504--1521.

\bibitem[{Brown et~al.(2020)Brown, Mann, Ryder, Subbiah, Kaplan, Dhariwal, Neelakantan, Shyam, Sastry, Askell et~al.}]{brown2020language}
Tom Brown, Benjamin Mann, Nick Ryder, Melanie Subbiah, Jared~D Kaplan, Prafulla Dhariwal, Arvind Neelakantan, Pranav Shyam, Girish Sastry, Amanda Askell, et~al. 2020.
\newblock Language models are few-shot learners.
\newblock \emph{Advances in neural information processing systems}, 33:1877--1901.

\bibitem[{Caba~Heilbron et~al.(2015)Caba~Heilbron, Escorcia, Ghanem, and Carlos~Niebles}]{caba2015activitynet}
Fabian Caba~Heilbron, Victor Escorcia, Bernard Ghanem, and Juan Carlos~Niebles. 2015.
\newblock Activitynet: A large-scale video benchmark for human activity understanding.
\newblock In \emph{Proceedings of the ieee conference on computer vision and pattern recognition}, pages 961--970.

\bibitem[{Cambria et~al.(2011)Cambria, Song, Wang, and Hussain}]{cambria2011isanette}
Erik Cambria, Yangqiu Song, Haixun Wang, and Amir Hussain. 2011.
\newblock Isanette: A common and common sense knowledge base for opinion mining.
\newblock In \emph{2011 IEEE 11th International Conference on Data Mining Workshops}, pages 315--322. IEEE.

\bibitem[{Ceraolo et~al.(2024)Ceraolo, Kharlapenko, Khan, Reymond, Mihalcea, Schölkopf, Sachan, and Jin}]{ceraolo2024analyzinghumanquestioningbehavior}
Roberto Ceraolo, Dmitrii Kharlapenko, Ahmad Khan, Amélie Reymond, Rada Mihalcea, Bernhard Schölkopf, Mrinmaya Sachan, and Zhijing Jin. 2024.
\newblock \href {http://arxiv.org/abs/2405.20318} {Analyzing human questioning behavior and causal curiosity through natural queries}.

\bibitem[{Chen et~al.(2023)Chen, Shi, Fu, Cheng, Li, and Xiao}]{chen-etal-2023-say}
Jiangjie Chen, Wei Shi, Ziquan Fu, Sijie Cheng, Lei Li, and Yanghua Xiao. 2023.
\newblock \href {https://doi.org/10.18653/v1/2023.acl-long.550} {Say what you mean! large language models speak too positively about negative commonsense knowledge}.
\newblock In \emph{Proceedings of the 61st Annual Meeting of the Association for Computational Linguistics (Volume 1: Long Papers)}, pages 9890--9908, Toronto, Canada. Association for Computational Linguistics.

\bibitem[{Clark et~al.(2018)Clark, Cowhey, Etzioni, Khot, Sabharwal, Schoenick, and Tafjord}]{clark2018think}
Peter Clark, Isaac Cowhey, Oren Etzioni, Tushar Khot, Ashish Sabharwal, Carissa Schoenick, and Oyvind Tafjord. 2018.
\newblock Think you have solved question answering? try arc, the ai2 reasoning challenge.
\newblock \emph{arXiv preprint arXiv:1803.05457}.

\bibitem[{Davis(2023)}]{davis2023benchmarks}
Ernest Davis. 2023.
\newblock Benchmarks for automated commonsense reasoning: A survey.
\newblock \emph{ACM Computing Surveys}, 56(4):1--41.

\bibitem[{Davis and Marcus(2015)}]{davis2015commonsense}
Ernest Davis and Gary Marcus. 2015.
\newblock Commonsense reasoning and commonsense knowledge in artificial intelligence.
\newblock \emph{Communications of the ACM}, 58(9):92--103.

\bibitem[{Dubey et~al.(2024)Dubey, Jauhri, Pandey, Kadian, Al{-}Dahle, Letman, Mathur, Schelten, Yang, Fan, Goyal, Hartshorn, Yang, Mitra, Sravankumar, Korenev, Hinsvark, Rao, Zhang, Rodriguez, Gregerson, Spataru, Rozi{\`{e}}re, Biron, Tang, Chern, Caucheteux, Nayak, Bi, Marra, McConnell, Keller, Touret, Wu, Wong, Ferrer, Nikolaidis, Allonsius, Song, Pintz, Livshits, Esiobu, Choudhary, Mahajan, Garcia{-}Olano, Perino, Hupkes, Lakomkin, AlBadawy, Lobanova, Dinan, Smith, Radenovic, Zhang, Synnaeve, Lee, Anderson, Nail, Mialon, Pang, Cucurell, Nguyen, Korevaar, Xu, Touvron, Zarov, Ibarra, Kloumann, Misra, Evtimov, Copet, Lee, Geffert, Vranes, Park, Mahadeokar, Shah, van~der Linde, Billock, Hong, Lee, Fu, Chi, Huang, Liu, Wang, Yu, Bitton, Spisak, Park, Rocca, Johnstun, Saxe, Jia, Alwala, Upasani, Plawiak, Li, Heafield, Stone, and et~al.}]{DBLP:journals/corr/abs-2407-21783}
Abhimanyu Dubey, Abhinav Jauhri, Abhinav Pandey, Abhishek Kadian, Ahmad Al{-}Dahle, Aiesha Letman, Akhil Mathur, Alan Schelten, Amy Yang, Angela Fan, Anirudh Goyal, Anthony Hartshorn, Aobo Yang, Archi Mitra, Archie Sravankumar, Artem Korenev, Arthur Hinsvark, Arun Rao, Aston Zhang, Aur{\'{e}}lien Rodriguez, Austen Gregerson, Ava Spataru, Baptiste Rozi{\`{e}}re, Bethany Biron, Binh Tang, Bobbie Chern, Charlotte Caucheteux, Chaya Nayak, Chloe Bi, Chris Marra, Chris McConnell, Christian Keller, Christophe Touret, Chunyang Wu, Corinne Wong, Cristian~Canton Ferrer, Cyrus Nikolaidis, Damien Allonsius, Daniel Song, Danielle Pintz, Danny Livshits, David Esiobu, Dhruv Choudhary, Dhruv Mahajan, Diego Garcia{-}Olano, Diego Perino, Dieuwke Hupkes, Egor Lakomkin, Ehab AlBadawy, Elina Lobanova, Emily Dinan, Eric~Michael Smith, Filip Radenovic, Frank Zhang, Gabriel Synnaeve, Gabrielle Lee, Georgia~Lewis Anderson, Graeme Nail, Gr{\'{e}}goire Mialon, Guan Pang, Guillem Cucurell, Hailey Nguyen, Hannah Korevaar, Hu~Xu, Hugo
  Touvron, Iliyan Zarov, Imanol~Arrieta Ibarra, Isabel~M. Kloumann, Ishan Misra, Ivan Evtimov, Jade Copet, Jaewon Lee, Jan Geffert, Jana Vranes, Jason Park, Jay Mahadeokar, Jeet Shah, Jelmer van~der Linde, Jennifer Billock, Jenny Hong, Jenya Lee, Jeremy Fu, Jianfeng Chi, Jianyu Huang, Jiawen Liu, Jie Wang, Jiecao Yu, Joanna Bitton, Joe Spisak, Jongsoo Park, Joseph Rocca, Joshua Johnstun, Joshua Saxe, Junteng Jia, Kalyan~Vasuden Alwala, Kartikeya Upasani, Kate Plawiak, Ke~Li, Kenneth Heafield, Kevin Stone, and et~al. 2024.
\newblock \href {https://doi.org/10.48550/ARXIV.2407.21783} {The llama 3 herd of models}.
\newblock \emph{CoRR}, abs/2407.21783.

\bibitem[{Gao et~al.(2024)Gao, Tow, Abbasi, Biderman, Black, DiPofi, Foster, Golding, Hsu, Le~Noac'h, Li, McDonell, Muennighoff, Ociepa, Phang, Reynolds, Schoelkopf, Skowron, Sutawika, Tang, Thite, Wang, Wang, and Zou}]{eval-harness}
Leo Gao, Jonathan Tow, Baber Abbasi, Stella Biderman, Sid Black, Anthony DiPofi, Charles Foster, Laurence Golding, Jeffrey Hsu, Alain Le~Noac'h, Haonan Li, Kyle McDonell, Niklas Muennighoff, Chris Ociepa, Jason Phang, Laria Reynolds, Hailey Schoelkopf, Aviya Skowron, Lintang Sutawika, Eric Tang, Anish Thite, Ben Wang, Kevin Wang, and Andy Zou. 2024.
\newblock \href {https://doi.org/10.5281/zenodo.12608602} {A framework for few-shot language model evaluation}.

\bibitem[{Geirhos et~al.(2020)Geirhos, Jacobsen, Michaelis, Zemel, Brendel, Bethge, and Wichmann}]{geirhos2020shortcut}
Robert Geirhos, J{\"o}rn-Henrik Jacobsen, Claudio Michaelis, Richard Zemel, Wieland Brendel, Matthias Bethge, and Felix~A Wichmann. 2020.
\newblock Shortcut learning in deep neural networks.
\newblock \emph{Nature Machine Intelligence}, 2(11):665--673.

\bibitem[{Guo et~al.(2024)Guo, You, Li, Bowen, and Zhang}]{guo2024exploring}
Pei Guo, Wangjie You, Juntao Li, Yan Bowen, and Min Zhang. 2024.
\newblock Exploring reversal mathematical reasoning ability for large language models.
\newblock In \emph{Findings of the Association for Computational Linguistics ACL 2024}, pages 13671--13685.

\bibitem[{Huang et~al.(2023)Huang, Tang, Zhang, Zhao, Song, Xia, and Wei}]{huang-etal-2023-languages}
Haoyang Huang, Tianyi Tang, Dongdong Zhang, Xin Zhao, Ting Song, Yan Xia, and Furu Wei. 2023.
\newblock \href {https://doi.org/10.18653/v1/2023.findings-emnlp.826} {Not all languages are created equal in {LLM}s: Improving multilingual capability by cross-lingual-thought prompting}.
\newblock In \emph{Findings of the Association for Computational Linguistics: EMNLP 2023}, pages 12365--12394, Singapore. Association for Computational Linguistics.

\bibitem[{Ismayilzada et~al.(2023)Ismayilzada, Paul, Montariol, Geva, and Bosselut}]{ismayilzada-etal-2023-crow}
Mete Ismayilzada, Debjit Paul, Syrielle Montariol, Mor Geva, and Antoine Bosselut. 2023.
\newblock \href {https://aclanthology.org/2023.emnlp-main.607} {{CR}o{W}: Benchmarking commonsense reasoning in real-world tasks}.
\newblock In \emph{Proceedings of the 2023 Conference on Empirical Methods in Natural Language Processing}, pages 9785--9821, Singapore. Association for Computational Linguistics.

\bibitem[{Jia and Liang(2017)}]{jia2017adversarial}
Robin Jia and Percy Liang. 2017.
\newblock Adversarial examples for evaluating reading comprehension systems.
\newblock In \emph{Proceedings of the 2017 Conference on Empirical Methods in Natural Language Processing}, pages 2021--2031.

\bibitem[{Jiang et~al.(2023)Jiang, Sablayrolles, Mensch, Bamford, Chaplot, de~Las~Casas, Bressand, Lengyel, Lample, Saulnier, Lavaud, Lachaux, Stock, Scao, Lavril, Wang, Lacroix, and Sayed}]{DBLP:journals/corr/abs-2310-06825}
Albert~Q. Jiang, Alexandre Sablayrolles, Arthur Mensch, Chris Bamford, Devendra~Singh Chaplot, Diego de~Las~Casas, Florian Bressand, Gianna Lengyel, Guillaume Lample, Lucile Saulnier, L{\'{e}}lio~Renard Lavaud, Marie{-}Anne Lachaux, Pierre Stock, Teven~Le Scao, Thibaut Lavril, Thomas Wang, Timoth{\'{e}}e Lacroix, and William~El Sayed. 2023.
\newblock \href {https://doi.org/10.48550/ARXIV.2310.06825} {Mistral 7b}.
\newblock \emph{CoRR}, abs/2310.06825.

\bibitem[{Johnson and Marasovic(2023)}]{johnson2023much}
Jacob~K Johnson and Ana Marasovic. 2023.
\newblock How much consistency is your accuracy worth?
\newblock \emph{EMNLP 2023}, page 250.

\bibitem[{Khot et~al.(2020)Khot, Clark, Guerquin, Jansen, and Sabharwal}]{khot2020qasc}
Tushar Khot, Peter Clark, Michal Guerquin, Peter Jansen, and Ashish Sabharwal. 2020.
\newblock Qasc: A dataset for question answering via sentence composition.
\newblock In \emph{Proceedings of the AAAI Conference on Artificial Intelligence}, volume~34, pages 8082--8090.

\bibitem[{Koupaee and Wang(2018)}]{koupaee2018wikihow}
Mahnaz Koupaee and William~Yang Wang. 2018.
\newblock Wikihow: A large scale text summarization dataset.
\newblock \emph{arXiv preprint arXiv:1810.09305}.

\bibitem[{Krathwohl(1973)}]{krathwohl1973taxonomy}
David~R Krathwohl. 1973.
\newblock Taxonomy of educational objectives.
\newblock \emph{Affective domain}.

\bibitem[{Li et~al.(2024)Li, Wang, Li, Guo, Zhang, and Feng}]{li2024evaluating}
Xiaoyuan Li, Wenjie Wang, Moxin Li, Junrong Guo, Yang Zhang, and Fuli Feng. 2024.
\newblock Evaluating mathematical reasoning of large language models: A focus on error identification and correction.
\newblock In \emph{Findings of the Association for Computational Linguistics ACL 2024}, pages 11316--11360.

\bibitem[{Lin et~al.(2020)Lin, Lee, Khanna, and Ren}]{lin2020birds}
Bill~Yuchen Lin, Seyeon Lee, Rahul Khanna, and Xiang Ren. 2020.
\newblock \href {https://doi.org/10.18653/v1/2020.emnlp-main.557} {{B}irds have four legs?! {N}umer{S}ense: {P}robing {N}umerical {C}ommonsense {K}nowledge of {P}re-{T}rained {L}anguage {M}odels}.
\newblock In \emph{Proceedings of the 2020 Conference on Empirical Methods in Natural Language Processing (EMNLP)}, pages 6862--6868, Online. Association for Computational Linguistics.

\bibitem[{Liu and Singh(2004)}]{liu2004conceptnet}
Hugo Liu and Push Singh. 2004.
\newblock Conceptnet—a practical commonsense reasoning tool-kit.
\newblock \emph{BT technology journal}, 22(4):211--226.

\bibitem[{Ma et~al.(2025)Ma, Du, Wang, Zhang, ZhoufutuWen, Qu, Yang, Liu, minghao liu, Yue, Huang, and Zhang}]{ma2024kor}
Kaijing Ma, Xeron Du, Yunran Wang, Haoran Zhang, ZhoufutuWen, Xingwei Qu, Jian Yang, Jiaheng Liu, minghao liu, Xiang Yue, Wenhao Huang, and Ge~Zhang. 2025.
\newblock {KOR}-bench: Benchmarking language models on knowledge-orthogonal reasoning tasks.
\newblock In \emph{The Thirteenth International Conference on Learning Representations}.

\bibitem[{Mihaylov et~al.(2018{\natexlab{a}})Mihaylov, Clark, Khot, and Sabharwal}]{mihaylov2018can}
Todor Mihaylov, Peter Clark, Tushar Khot, and Ashish Sabharwal. 2018{\natexlab{a}}.
\newblock Can a suit of armor conduct electricity? a new dataset for open book question answering.
\newblock In \emph{Proceedings of the 2018 Conference on Empirical Methods in Natural Language Processing}, pages 2381--2391.

\bibitem[{Mihaylov et~al.(2018{\natexlab{b}})Mihaylov, Clark, Khot, and Sabharwal}]{OpenBookQA2018}
Todor Mihaylov, Peter Clark, Tushar Khot, and Ashish Sabharwal. 2018{\natexlab{b}}.
\newblock Can a suit of armor conduct electricity? a new dataset for open book question answering.
\newblock In \emph{EMNLP}.

\bibitem[{OpenAI(2023)}]{DBLP:journals/corr/abs-2303-08774}
OpenAI. 2023.
\newblock \href {https://doi.org/10.48550/ARXIV.2303.08774} {{GPT-4} technical report}.
\newblock \emph{CoRR}, abs/2303.08774.

\bibitem[{OpenAI et~al.(2024)OpenAI, Adler, Agarwal, Ahmad, Akkaya, Aleman, Almeida, Altenschmidt, Altman, Anadkat et~al.}]{openai2024gpt}
Josh~Achiam OpenAI, Steven Adler, Sandhini Agarwal, Lama Ahmad, Ilge Akkaya, Florencia~Leoni Aleman, Diogo Almeida, Janko Altenschmidt, Sam Altman, Shyamal Anadkat, et~al. 2024.
\newblock Gpt-4 technical report, 2024.
\newblock \emph{URL https://arxiv. org/abs/2303.08774}.

\bibitem[{Qin et~al.(2019)Qin, Bosselut, Holtzman, Bhagavatula, Clark, and Choi}]{DBLP:conf/emnlp/QinBHBCC19}
Lianhui Qin, Antoine Bosselut, Ari Holtzman, Chandra Bhagavatula, Elizabeth Clark, and Yejin Choi. 2019.
\newblock \href {https://doi.org/10.18653/V1/D19-1509} {Counterfactual story reasoning and generation}.
\newblock In \emph{Proceedings of the 2019 Conference on Empirical Methods in Natural Language Processing and the 9th International Joint Conference on Natural Language Processing, {EMNLP-IJCNLP} 2019, Hong Kong, China, November 3-7, 2019}, pages 5042--5052. Association for Computational Linguistics.

\bibitem[{Qin et~al.(2021)Qin, Gupta, Upadhyay, He, Choi, and Faruqui}]{qin2021timedial}
Lianhui Qin, Aditya Gupta, Shyam Upadhyay, Luheng He, Yejin Choi, and Manaal Faruqui. 2021.
\newblock Timedial: Temporal commonsense reasoning in dialog.
\newblock In \emph{Proceedings of the 59th Annual Meeting of the Association for Computational Linguistics and the 11th International Joint Conference on Natural Language Processing (Volume 1: Long Papers)}, pages 7066--7076.

\bibitem[{Rivi{\`e}re et~al.(2024)Rivi{\`e}re, Pathak, Sessa, Hardin, Bhupatiraju, Hussenot, Mesnard, Shahriari, Ram{\'e}, Ferret et~al.}]{riviere2024gemma}
Morgane Rivi{\`e}re, Shreya Pathak, Pier~Giuseppe Sessa, Cassidy Hardin, Surya Bhupatiraju, L{\'e}onard Hussenot, Thomas Mesnard, Bobak Shahriari, Alexandre Ram{\'e}, Johan Ferret, et~al. 2024.
\newblock Gemma 2: Improving open language models at a practical size.
\newblock \emph{CoRR}.

\bibitem[{Ross et~al.(2022)Ross, Peters, and Marasovi{\'c}}]{ross2022does}
Alexis Ross, Matthew~E Peters, and Ana Marasovi{\'c}. 2022.
\newblock Does self-rationalization improve robustness to spurious correlations?
\newblock In \emph{Proceedings of the 2022 Conference on Empirical Methods in Natural Language Processing}, pages 7403--7416.

\bibitem[{Rudinger et~al.(2020)Rudinger, Shwartz, Hwang, Bhagavatula, Forbes, Bras, Smith, and Choi}]{DBLP:conf/emnlp/RudingerSHBFBSC20}
Rachel Rudinger, Vered Shwartz, Jena~D. Hwang, Chandra Bhagavatula, Maxwell Forbes, Ronan~Le Bras, Noah~A. Smith, and Yejin Choi. 2020.
\newblock \href {https://doi.org/10.18653/V1/2020.FINDINGS-EMNLP.418} {Thinking like a skeptic: Defeasible inference in natural language}.
\newblock In \emph{Findings of the Association for Computational Linguistics: {EMNLP} 2020, Online Event, 16-20 November 2020}, volume {EMNLP} 2020 of \emph{Findings of {ACL}}, pages 4661--4675. Association for Computational Linguistics.

\bibitem[{Sap et~al.(2019)Sap, Rashkin, Chen, Le~Bras, and Choi}]{sap2019socialiqa}
Maarten Sap, Hannah Rashkin, Derek Chen, Ronan Le~Bras, and Yejin Choi. 2019.
\newblock \href {https://doi.org/10.18653/v1/D19-1454} {Social {IQ}a: Commonsense reasoning about social interactions}.
\newblock In \emph{Proceedings of the 2019 Conference on Empirical Methods in Natural Language Processing and the 9th International Joint Conference on Natural Language Processing (EMNLP-IJCNLP)}, pages 4463--4473, Hong Kong, China. Association for Computational Linguistics.

\bibitem[{Shi et~al.(2024)Shi, You, Huang, Li, and Xiong}]{shi2024corecode}
Dan Shi, Chaobin You, Jiantao Huang, Taihao Li, and Deyi Xiong. 2024.
\newblock Corecode: A common sense annotated dialogue dataset with benchmark tasks for chinese large language models.
\newblock In \emph{Proceedings of the AAAI Conference on Artificial Intelligence}, volume~38, pages 18952--18960.

\bibitem[{Shi et~al.(2022)Shi, Suzgun, Freitag, Wang, Srivats, Vosoughi, Chung, Tay, Ruder, Zhou et~al.}]{shi2022language}
Freda Shi, Mirac Suzgun, Markus Freitag, Xuezhi Wang, Suraj Srivats, Soroush Vosoughi, Hyung~Won Chung, Yi~Tay, Sebastian Ruder, Denny Zhou, et~al. 2022.
\newblock Language models are multilingual chain-of-thought reasoners.
\newblock \emph{arXiv preprint arXiv:2210.03057}.

\bibitem[{Sinha et~al.(2019)Sinha, Sodhani, Dong, Pineau, and Hamilton}]{DBLP:conf/emnlp/SinhaSDPH19}
Koustuv Sinha, Shagun Sodhani, Jin Dong, Joelle Pineau, and William~L. Hamilton. 2019.
\newblock \href {https://doi.org/10.18653/V1/D19-1458} {{CLUTRR:} {A} diagnostic benchmark for inductive reasoning from text}.
\newblock In \emph{Proceedings of the 2019 Conference on Empirical Methods in Natural Language Processing and the 9th International Joint Conference on Natural Language Processing, {EMNLP-IJCNLP} 2019, Hong Kong, China, November 3-7, 2019}, pages 4505--4514. Association for Computational Linguistics.

\bibitem[{Storks and Chai(2021)}]{storks2021beyond}
Shane Storks and Joyce Chai. 2021.
\newblock Beyond the tip of the iceberg: Assessing coherence of text classifiers.
\newblock In \emph{Findings of the Association for Computational Linguistics: EMNLP 2021}, pages 3169--3177.

\bibitem[{Sun et~al.(2024)Sun, Huang, Wu, Gu, Li, Zhang, Yan, and He}]{sun2024benchmarking}
Jiaxing Sun, Weiquan Huang, Jiang Wu, Chenya Gu, Wei Li, Songyang Zhang, Hang Yan, and Conghui He. 2024.
\newblock \href {https://doi.org/10.18653/v1/2024.acl-long.604} {Benchmarking {C}hinese commonsense reasoning of {LLM}s: From {C}hinese-specifics to reasoning-memorization correlations}.
\newblock In \emph{Proceedings of the 62nd Annual Meeting of the Association for Computational Linguistics (Volume 1: Long Papers)}, pages 11205--11228, Bangkok, Thailand. Association for Computational Linguistics.

\bibitem[{Talmor et~al.(2019)Talmor, Herzig, Lourie, and Berant}]{talmor2019commonsenseqa}
Alon Talmor, Jonathan Herzig, Nicholas Lourie, and Jonathan Berant. 2019.
\newblock Commonsenseqa: A question answering challenge targeting commonsense knowledge.
\newblock In \emph{Proceedings of the 2019 Conference of the North American Chapter of the Association for Computational Linguistics: Human Language Technologies, Volume 1 (Long and Short Papers)}, pages 4149--4158.

\bibitem[{Team et~al.(2024)Team, Georgiev, Lei, Burnell, Bai, Gulati, Tanzer, Vincent, Pan, Wang et~al.}]{team2024gemini}
Gemini Team, Petko Georgiev, Ving~Ian Lei, Ryan Burnell, Libin Bai, Anmol Gulati, Garrett Tanzer, Damien Vincent, Zhufeng Pan, Shibo Wang, et~al. 2024.
\newblock Gemini 1.5: Unlocking multimodal understanding across millions of tokens of context.
\newblock \emph{arXiv preprint arXiv:2403.05530}.

\bibitem[{Team(2023)}]{2023internlm}
InternLM Team. 2023.
\newblock Internlm: A multilingual language model with progressively enhanced capabilities.
\newblock \url{https://github.com/InternLM/InternLM-techreport}.

\bibitem[{Wallace et~al.(2019)Wallace, Feng, Kandpal, Gardner, and Singh}]{wallace2019universal}
Eric Wallace, Shi Feng, Nikhil Kandpal, Matt Gardner, and Sameer Singh. 2019.
\newblock Universal adversarial triggers for attacking and analyzing nlp.
\newblock In \emph{Proceedings of the 2019 Conference on Empirical Methods in Natural Language Processing and the 9th International Joint Conference on Natural Language Processing (EMNLP-IJCNLP)}, pages 2153--2162.

\bibitem[{Wu et~al.(2024)Wu, Qiu, Ross, Aky{\"u}rek, Chen, Wang, Kim, Andreas, and Kim}]{wu-etal-2024-reasoning}
Zhaofeng Wu, Linlu Qiu, Alexis Ross, Ekin Aky{\"u}rek, Boyuan Chen, Bailin Wang, Najoung Kim, Jacob Andreas, and Yoon Kim. 2024.
\newblock \href {https://doi.org/10.18653/v1/2024.naacl-long.102} {Reasoning or reciting? exploring the capabilities and limitations of language models through counterfactual tasks}.
\newblock In \emph{Proceedings of the 2024 Conference of the North American Chapter of the Association for Computational Linguistics: Human Language Technologies (Volume 1: Long Papers)}, pages 1819--1862, Mexico City, Mexico. Association for Computational Linguistics.

\bibitem[{Yang et~al.(2023)Yang, Xiao, Wang, Zhang, Bian, Yin, Lv, Pan, Wang, Yan, Yang, Deng, Wang, Liu, Ai, Dong, Zhao, Xu, Sun, Zhang, Liu, Ji, Xie, Dai, Fang, Su, Song, Liu, Ru, Ma, Wang, Liu, Lin, Nie, Guo, Sun, Zhang, Li, Li, Cheng, Chen, Zeng, Wang, Chen, Men, Yu, Pan, Shen, Wang, Li, Jiang, Gao, Zhang, Zhou, and Wu}]{DBLP:journals/corr/abs-2309-10305}
Aiyuan Yang, Bin Xiao, Bingning Wang, Borong Zhang, Ce~Bian, Chao Yin, Chenxu Lv, Da~Pan, Dian Wang, Dong Yan, Fan Yang, Fei Deng, Feng Wang, Feng Liu, Guangwei Ai, Guosheng Dong, Haizhou Zhao, Hang Xu, Haoze Sun, Hongda Zhang, Hui Liu, Jiaming Ji, Jian Xie, Juntao Dai, Kun Fang, Lei Su, Liang Song, Lifeng Liu, Liyun Ru, Luyao Ma, Mang Wang, Mickel Liu, MingAn Lin, Nuolan Nie, Peidong Guo, Ruiyang Sun, Tao Zhang, Tianpeng Li, Tianyu Li, Wei Cheng, Weipeng Chen, Xiangrong Zeng, Xiaochuan Wang, Xiaoxi Chen, Xin Men, Xin Yu, Xuehai Pan, Yanjun Shen, Yiding Wang, Yiyu Li, Youxin Jiang, Yuchen Gao, Yupeng Zhang, Zenan Zhou, and Zhiying Wu. 2023.
\newblock \href {https://doi.org/10.48550/ARXIV.2309.10305} {Baichuan 2: Open large-scale language models}.
\newblock \emph{CoRR}, abs/2309.10305.

\bibitem[{Yang et~al.(2024)Yang, Yang, Hui, Zheng, Yu, Zhou, Li, Li, Liu, Huang et~al.}]{yang2024qwen2}
An~Yang, Baosong Yang, Binyuan Hui, Bo~Zheng, Bowen Yu, Chang Zhou, Chengpeng Li, Chengyuan Li, Dayiheng Liu, Fei Huang, et~al. 2024.
\newblock Qwen2 technical report.
\newblock \emph{arXiv preprint arXiv:2407.10671}.

\bibitem[{Yang et~al.(2022)Yang, Wang, Wu, Yang, and Zhang}]{DBLP:journals/corr/abs-2204-07408}
Linyi Yang, Zhen Wang, Yuxiang Wu, Jie Yang, and Yue Zhang. 2022.
\newblock \href {https://doi.org/10.48550/ARXIV.2204.07408} {Towards fine-grained causal reasoning and {QA}}.
\newblock \emph{CoRR}, abs/2204.07408.

\bibitem[{Young et~al.(2024)Young, Chen, Li, Huang, Zhang, Zhang, Li, Zhu, Chen, Chang, Yu, Liu, Liu, Yue, Yang, Yang, Yu, Xie, Huang, Hu, Ren, Niu, Nie, Xu, Liu, Wang, Cai, Gu, Liu, and Dai}]{DBLP:journals/corr/abs-2403-04652}
Alex Young, Bei Chen, Chao Li, Chengen Huang, Ge~Zhang, Guanwei Zhang, Heng Li, Jiangcheng Zhu, Jianqun Chen, Jing Chang, Kaidong Yu, Peng Liu, Qiang Liu, Shawn Yue, Senbin Yang, Shiming Yang, Tao Yu, Wen Xie, Wenhao Huang, Xiaohui Hu, Xiaoyi Ren, Xinyao Niu, Pengcheng Nie, Yuchi Xu, Yudong Liu, Yue Wang, Yuxuan Cai, Zhenyu Gu, Zhiyuan Liu, and Zonghong Dai. 2024.
\newblock \href {https://doi.org/10.48550/ARXIV.2403.04652} {Yi: Open foundation models by 01.ai}.
\newblock \emph{CoRR}, abs/2403.04652.

\bibitem[{Yuan et~al.(2023)Yuan, Chen, Fu, Ge, Shah, Jankowski, Xiao, and Yang}]{yuan2023distilling}
Siyu Yuan, Jiangjie Chen, Ziquan Fu, Xuyang Ge, Soham Shah, Charles Jankowski, Yanghua Xiao, and Deqing Yang. 2023.
\newblock Distilling script knowledge from large language models for constrained language planning.
\newblock In \emph{Proceedings of the 61st Annual Meeting of the Association for Computational Linguistics (Volume 1: Long Papers)}, pages 4303--4325.

\bibitem[{Zellers et~al.(2018)Zellers, Bisk, Schwartz, and Choi}]{zellers2018swag}
Rowan Zellers, Yonatan Bisk, Roy Schwartz, and Yejin Choi. 2018.
\newblock Swag: A large-scale adversarial dataset for grounded commonsense inference.
\newblock In \emph{Proceedings of the 2018 Conference on Empirical Methods in Natural Language Processing}, pages 93--104.

\bibitem[{Zellers et~al.(2019)Zellers, Holtzman, Bisk, Farhadi, and Choi}]{zellers2019hellaswag}
Rowan Zellers, Ari Holtzman, Yonatan Bisk, Ali Farhadi, and Yejin Choi. 2019.
\newblock Hellaswag: Can a machine really finish your sentence?
\newblock In \emph{Proceedings of the 57th Annual Meeting of the Association for Computational Linguistics}, pages 4791--4800.

\bibitem[{Zhou et~al.(2019)Zhou, Khashabi, Ning, and Roth}]{zhou2019going}
Ben Zhou, Daniel Khashabi, Qiang Ning, and Dan Roth. 2019.
\newblock “going on a vacation” takes longer than “going for a walk”: A study of temporal commonsense understanding.
\newblock In \emph{Proceedings of the 2019 Conference on Empirical Methods in Natural Language Processing and the 9th International Joint Conference on Natural Language Processing (EMNLP-IJCNLP)}, pages 3363--3369.

\bibitem[{Zhou et~al.(2021)Zhou, Khanna, Lee, Lin, Ho, Pujara, and Ren}]{zhou2021rica}
Pei Zhou, Rahul Khanna, Seyeon Lee, Bill~Yuchen Lin, Daniel Ho, Jay Pujara, and Xiang Ren. 2021.
\newblock Rica: Evaluating robust inference capabilities based on commonsense axioms.
\newblock In \emph{Proceedings of the 2021 Conference on Empirical Methods in Natural Language Processing}, pages 7560--7579.

\end{thebibliography}
